\documentclass[11pt,letterpaper,logo]{mystyle}

\usepackage{algorithmic}
\usepackage{algorithm}
\usepackage{array}
\usepackage{stfloats}
\usepackage{verbatim}
\usepackage[numbers]{natbib}
\usepackage{nicefrac}       
\usepackage{multirow}
\usepackage{tablefootnote}
\usepackage{makecell}
\usepackage{xparse}
\usepackage{fontawesome5}
\usepackage{bxcoloremoji}
\usepackage{twemojis}
\usepackage{bbding}

\usepackage{forest}
\usepackage{tikz}

\definecolor{deepgreen}{HTML}{057311}
\definecolor{AgentIndigo}{HTML}{3F51B5}
\definecolor{AgentIndigoLight}{HTML}{E8EAF6}
\definecolor{AgentAmber}{HTML}{FF8F00}
\definecolor{AgentAmberLight}{HTML}{FFF3E0}

\tcbset{
  agentscope/.style={
    colback=AgentIndigoLight,
    colframe=AgentIndigo,
    colbacktitle=AgentIndigo!20!white,
    coltitle=black,
    boxrule=0.9pt, arc=2mm,
    left=3mm, right=3mm, top=2mm, bottom=2mm,
    fonttitle=\bfseries,
    title=Survey Scope
  },
  agentcontrib/.style={
    colback=AgentAmberLight,
    colframe=AgentAmber,
    colbacktitle=AgentAmber!20!white,
    coltitle=black,
    boxrule=0.9pt, arc=2mm,
    left=3mm, right=3mm, top=2mm, bottom=2mm,
    fonttitle=\bfseries,
    title=Contributions
  },
  }

\runningtitle{Code as Agent Harness}

\title{%
\textbf{Code as Agent Harness}\\
\vspace{0.3em}
{\fontsize{11.5}{13.5}\selectfont\scshape\color{LARGBlue!85}
$\lozenge$~Toward Executable, Verifiable, and Stateful Agent Systems~$\lozenge$}
\vspace{-0.8em}
}

\author{
   \normalfont
   Xuying Ning$^{\textcolor{Maroon}{1}\dag}$ \quad
   Katherine Tieu$^{\textcolor{Maroon}{1}\dag}$ \quad
   Dongqi Fu$^{\textcolor{Maroon}{2}\dag}$ \quad
   Tianxin Wei$^{\textcolor{Maroon}{1}\dag}$ \quad
   Zihao Li$^{\textcolor{Maroon}{1}\dag}$ \quad
   Yuanchen Bei$^{\textcolor{Maroon}{1}\dag}$ \quad \\
   Jiaru Zou$^{\textcolor{Maroon}{3}}$ \quad 
   Mengting Ai$^{\textcolor{Maroon}{1}}$ \quad
   Zhining Liu$^{\textcolor{Maroon}{1}}$ \quad
   Ting-Wei Li$^{\textcolor{Maroon}{1}}$ \quad
   Lingjie Chen$^{\textcolor{Maroon}{1}}$ \quad
   Yanjun Zhao$^{\textcolor{Maroon}{1}}$ \quad
   Ke Yang$^{\textcolor{Maroon}{1}}$ \quad \\
   Bingxuan Li$^{\textcolor{Maroon}{1}}$ \quad 
   Cheng Qian$^{\textcolor{Maroon}{1}}$ \quad
   Gaotang Li$^{\textcolor{Maroon}1}$ \quad 
   Xiao Lin$^{\textcolor{Maroon}{1}}$ \quad 
   Zhichen Zeng$^{\textcolor{Maroon}{1}}$ \quad
   Ruizhong Qiu$^{\textcolor{Maroon}{1}}$ \quad 
   Sirui Chen$^{\textcolor{Maroon}{1}}$ \quad \\
   Yifan Sun$^{\textcolor{Maroon}{1}}$ \quad 
   Xiyuan Yang$^{\textcolor{Maroon}{1}}$ \quad 
   Ruida Wang$^{\textcolor{Maroon}{1}}$ \quad 
   Rui Pan$^{\textcolor{Maroon}{1}}$ \quad 
   Chenyuan Yang$^{\textcolor{Maroon}{1}}$ \quad
   Dylan Zhang$^{\textcolor{Maroon}{1}}$ \quad 
   Liri Fang$^{\textcolor{Maroon}{1}}$ \quad \\
   Zikun Cui$^{\textcolor{Maroon}{2}}$ \quad
   Yang Cao$^{\textcolor{Maroon}{2}}$ \quad
   Pan Chen$^{\textcolor{Maroon}{2}}$ \quad
   Dorothy Sun$^{\textcolor{Maroon}{2}}$ \quad
   Ren Chen$^{\textcolor{Maroon}{2}}$ \quad\\
   Mahesh Srinivasan$^{\textcolor{Maroon}{2}}$ \quad
   Nipun Mathur$^{\textcolor{Maroon}{2}}$ \quad 
   Yinglong Xia$^{\textcolor{Maroon}{2}}$ \quad
   Hong Li$^{\textcolor{Maroon}{2}}$ \quad 
   Hong Yan$^{\textcolor{Maroon}{2}}$ \quad\\
Pan Lu$^{\textcolor{Maroon}{3}}$ \quad 
Lingming Zhang $^{\textcolor{Maroon}{1}}$ \quad
Tong Zhang$^{\textcolor{Maroon}{1}}$ \quad 
Hanghang Tong$^{\textcolor{Maroon}{1}}$$^{\coloremojicode{2709}}$ \quad
   Jingrui He$^{\textcolor{Maroon}{1}}$$^{\coloremojicode{2709}}$ \quad
   \\
   \vspace{12pt}
   \small
   \raisebox{1.0ex}{\includegraphics[height=1.1ex]{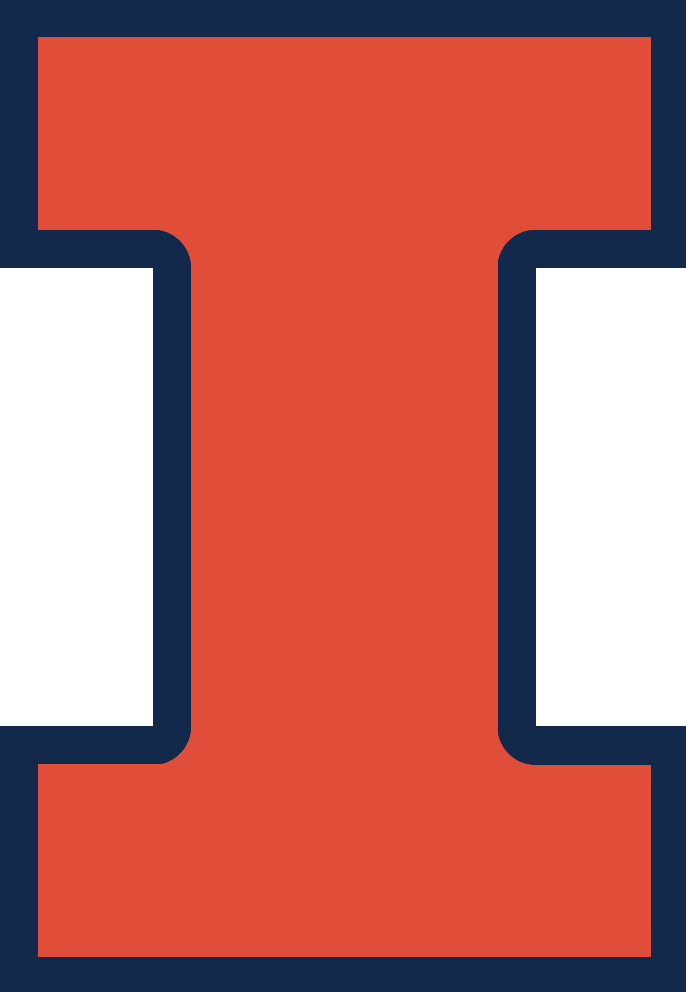}}$^{\textcolor{Maroon}{1}}$University of Illinois Urbana-Champaign \quad
   \raisebox{0.9ex}{\includegraphics[height=1.1ex]{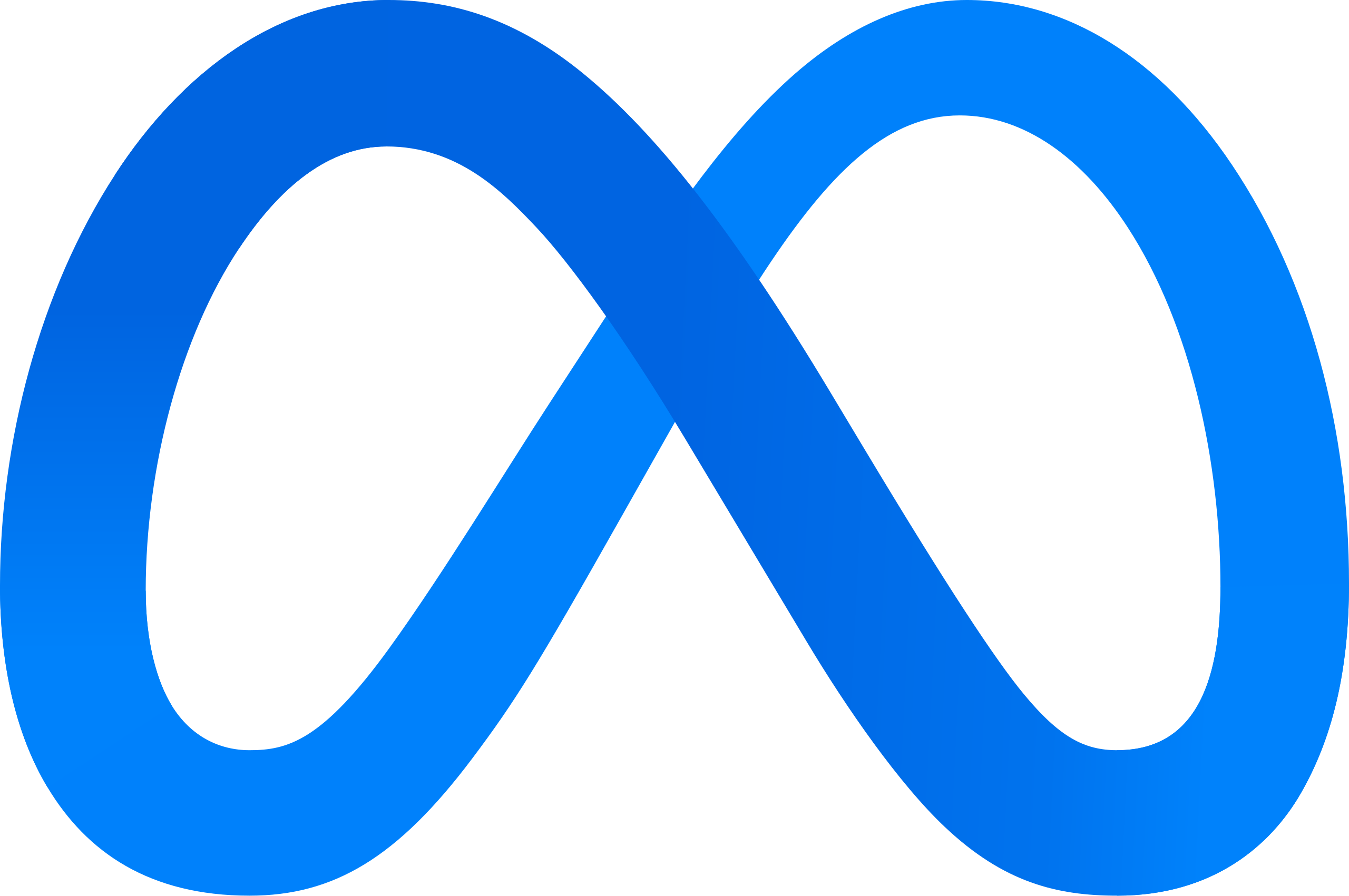}}$^{\textcolor{Maroon}{2}}$Meta \quad
   \raisebox{0.9ex}{\includegraphics[height=1.9ex] {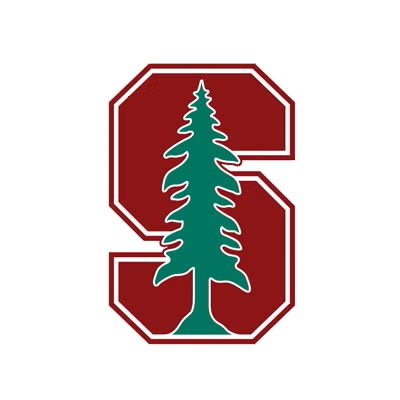}}$^{\textcolor{Maroon}{3}}$Stanford University
   \\
   $^{\dag}$ \textit{Core Contributor}, \quad
   $^{\coloremojicode{2709}}$ \textit{Corresponding Author}
}

\date{\vspace{-3ex}}

\begin{document}

\begin{abstract}
\textbf{\large Abstract:} Recent large language models (LLMs) have demonstrated strong
capabilities in understanding and generating code, from
competitive programming to repository-level software engineering.
In emerging agentic systems, code is no longer only a target output.
It increasingly serves as an operational substrate for agent
reasoning, acting, environment modeling, and execution-based verification.
We frame this shift through the lens of \emph{agent harnesses} and
introduce \emph{code as agent harness}: a unified view that centers code as the basis for agent
infrastructure.
To systematically study this perspective, we organize the survey around three connected layers. First, we study the \emph{harness interface}, where code connects agents
to reasoning, action, and environment modeling. Second, we examine \emph{harness mechanisms}: planning, memory, and tool use for long-horizon execution, together with feedback-driven control and optimization that make harness reliable and adaptive. Third, we discuss \emph{scaling the harness} from single-agent systems to multi-agent settings, where shared code artifacts support multi-agent coordination, review, and verification.
Across these layers, we summarize representative methods and practical
applications of \emph{code as agent harness}, spanning coding
assistants, GUI/OS automation, embodied agents, scientific
discovery, personalization and recommendation, DevOps, and enterprise
workflows.
We further outline open challenges for harness engineering,
including evaluation beyond final task success, verification under incomplete
feedback, regression-free harness improvement, consistent shared state across
multiple agents, human oversight for safety-critical actions, and extensions
to multimodal environments.
By centering code as the harness of agentic AI, this survey
provides a unified roadmap toward executable, verifiable, and
stateful AI agent systems.

\vspace{5mm}
\textcolor{Maroon}{\faBullseye}~\textbf{Keywords}: Agent Harness, Coding Agent, Harness Engineering, Agentic AI \\
\faGithub~\textbf{Github}: \url{https://github.com/YennNing/Awesome-Code-as-Agent-Harness-Papers}
\end{abstract}

\maketitle

\vspace{-3mm}
\section{Introduction}
\label{sec:intro}
Recent large language models (LLMs) have demonstrated strong capabilities in
understanding and generating code~\cite{chen2021evaluating,austin2021program,nijkamp2022codegen},
achieving strong performance in tasks ranging from competitive
programming~\cite{li2022competition} to repository-level software
engineering~\cite{jimenez2023swe}.
Building on these capabilities, the role of code in agentic systems is
expanding beyond a target artifact to be generated.
Programs are increasingly used as the medium through which LLM
agents reason, act, and model their environments.
Program-aided reasoning methods externalize intermediate
computation into executable code~\cite{chen2022program,gao2023pal,li2023chain};
robotic and embodied agents use generated programs as executable
policies for interacting with physical or simulated
worlds~\cite{ahn2022can,liang2023code};
and software-engineering or interactive environments use
codebases, execution traces, tests, and runtime feedback as
structured representations of environment state and dynamics, in
which agents plan, act, and revise their behavior~\cite{yang2023intercode,jimenez2023swe,liu2023agentbench}.
Taken together, these developments suggest a broader view:
code is not only an artifact generated by LLMs, but also an
executable, inspectable, and stateful medium through which agents
reason, act, observe feedback, and verify progress. We refer to
this view as \emph{code as agent harness}.

\begin{figure}[t]
    \centering
    \includegraphics[width=1.0\linewidth]{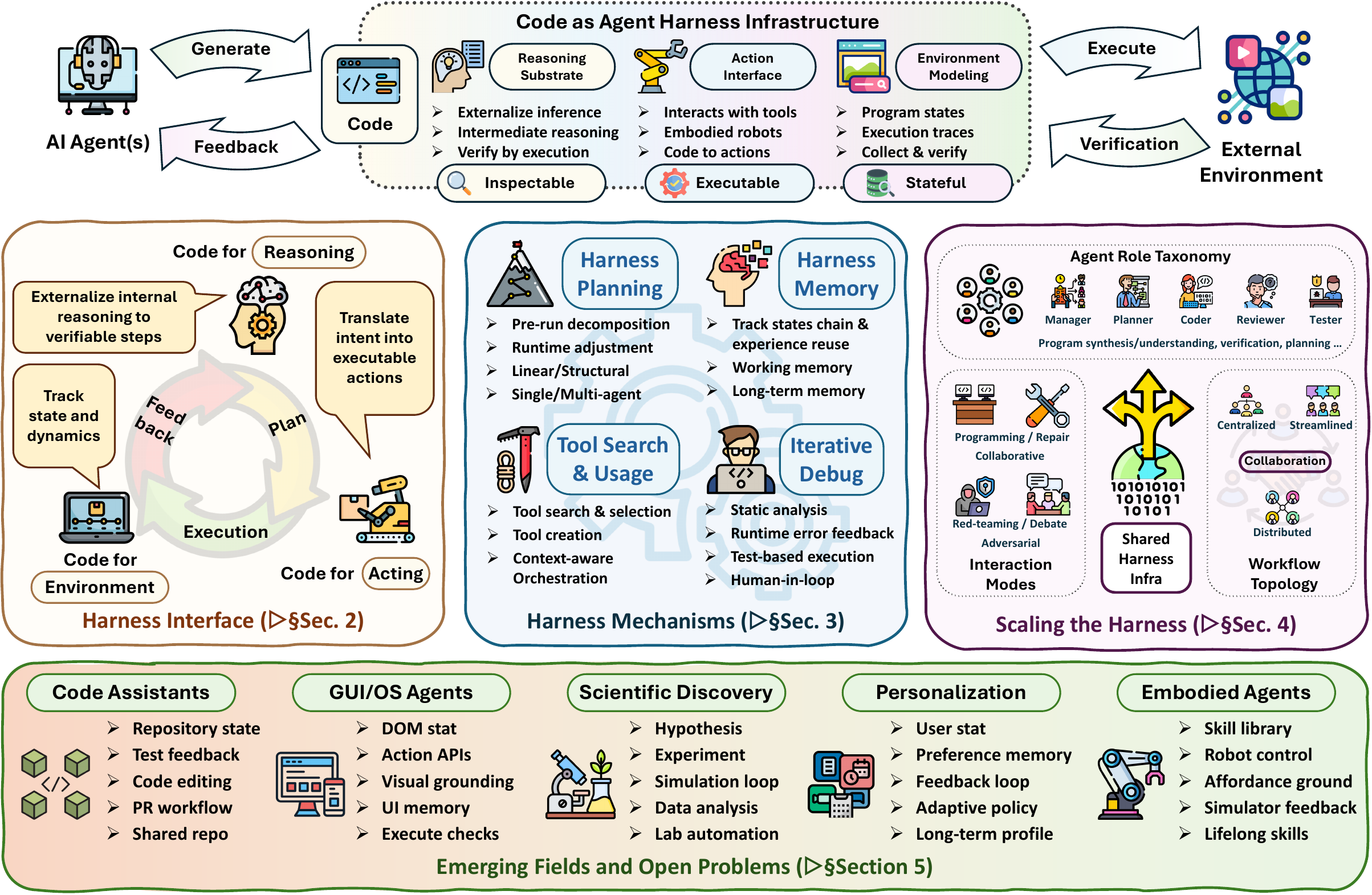}
    \caption{Taxonomy of code as agent harness.}
    \label{fig:taxonomy}
    \vspace{-5mm}
\end{figure}

Recent discussions on \emph{agent harnesses}~\cite{lee2026metaharness,lou2026autoharness,anthropic2025longrunning,lopopolo2026harnessengineering}
provide a useful system-level lens for understanding this shift.
An agent harness refers to the software layer that surrounds an LLM
with tools, APIs, sandboxes, memory, validators, permission
boundaries, execution loops, and feedback channels, thereby turning
a stateless model into a functional agent capable of long-running
task execution~\cite{zhang2025agentic,agrawal2025gepa,zhang2023toolcoder,wang2025teaching,lavon2025execution,cheng2026llm,dai2025feedbackeval}.
In this view, the bottleneck of autonomy is not only the reasoning
ability of the base model, but also the reliability of the system
that connects model outputs to long-horizon actions and persistent states.

To clarify the role of code in this broader harness view, we
distinguish three coupled elements of long-running agentic systems:
\emph{model-internal capabilities}, \emph{system-provided harness
infrastructure}, and \emph{agent-initiated code artifacts}.
\emph{Model-internal capabilities} refer to the model's reasoning,
perception, planning, simulation, and evaluation abilities.
\emph{System-provided harness infrastructure} refers to the
predefined tools, APIs, sandboxes, memory systems, validators,
permission boundaries, telemetry, and workflows that connect model
outputs to external actions and feedback, and forms the main focus
of harness engineering~\cite{openai2026harnessengineering,langchainanatomyharness2026}.
In contrast, \emph{agent-initiated code artifacts}, which remain
relatively underexplored, are interactive code objects that agents
create, execute, observe, revise, persist, and share within the task
execution loop. Through execution feedback, these artifacts help
agents reason, act, verify progress, store state, and coordinate
with other agents. Examples include regression tests, temporary
tools, DSL programs, executable workflows, reusable skills, and
intermediate program states. Representative systems such as Claude
Code~\cite{claudecode2025}, Codex~\cite{codex2025}, LangChain
~\cite{langchaindeepagentsharness2026}, and enterprise agent
platforms show how these elements jointly enable adaptation in long-running
agent systems.

With this distinction in mind, we revisit the role of code
in agentic systems. Existing surveys typically either treat code as the end product of
LLMs. In contrast, we focus on \emph{agent-initiated code artifacts} and
how model capabilities construct and evolve them through
interaction with harness infrastructure, with code serving as the
organizing center for the interface, agent capabilities,
and multi-agent coordination.
Across diverse agentic systems, code is used not only to produce
solutions, but also to execute reasoning, ground actions, maintain
state, and expose feedback. We term this view \emph{code as agent
harness}: code as the executable and inspectable medium through
which agents reason, act, and adapt. This shifts the scope from producing correct programs to
understanding how code supports reliable closed-loop agentic
behavior.

To systematically characterize \emph{code as agent harness}, we
organize the survey into three connected layers, as shown in
Figure~\ref{fig:taxonomy}.
This organization follows how code becomes an operational medium
inside the agent loop: it first enters as a harness interface
for reasoning, acting, and environment representation; it then
supports harness mechanisms that manage planning, memory, tool
use, execution, and repair over time; and it finally becomes a
shared artifact through which multiple agents coordinate over
repositories, tests, traces, workflows, and execution states.

First, \emph{\textbf{Harness Interface: Code for Reasoning, Acting, and
Environment Modeling}} (\S\ref{sec:foundations}) studies how code
forms the basic interface between a model and its task
environment.
At this layer, code is the medium that converts model outputs into
executable and inspectable structures.
We review \emph{code for reasoning}, where programs externalize
intermediate computation and allow interpreters, symbolic solvers,
execution traces, or process rewards to check and refine reasoning
~\cite{gao2023pal,chen2022program,li2023chain,ye2023satlm,ni2024next,li2025codeprm}.
We then review \emph{code for acting}, where generated programs
serve as policies, tool calls, behavior trees, or reusable skills
for embodied, GUI, and software environments
~\cite{ahn2022can,liang2023code,wang2023voyager,mu2024robocodex,zhang2025codebt,lin2026ui}.
Finally, we examine \emph{code for environment modeling}, where
program states, repositories, traces, simulators, and tests
represent state, dynamics, and feedback signals for agent
interaction
~\cite{tang2024worldcoder,copet2025cwm,zheng2026code2world,jimenez2023swe,liu2023agentbench,gandhi2026endless}.
This layer establishes the core harness interface: code is how the
agent makes reasoning executable, action programmable, and
environment state inspectable.

Building on this interface, \emph{\textbf{Harness Mechanisms:
Planning, Memory, Tool Use, Control, and Optimization}}~(\S\ref{sec:modules})
studies how code-harnessed agents remain reliable beyond a single
generation step.
Once code is placed inside the agent loop, the harness must decide
what to execute next, preserve useful state, expose the right
tools, and convert failures into corrective actions.
We therefore review planning methods that organize long-horizon
software tasks through decomposition, structural grounding,
trajectory search, or workflow orchestration
~\cite{jiang2024selfplanning,gur2023webagent,bairi2024codeplan,li2025codetree,islam2024mapcoder};
memory methods that maintain working state, retrieve repository
evidence, store reusable experience, and support shared
interaction histories
~\cite{gaurav2025codemem,zhang2024autocoderover,zhang2023repocoder,wang2026memgovern};
tool-use methods that connect agents to APIs, repositories,
execution environments, and verification tools
~\cite{zhang2023toolcoder,liu2024toolnet};
and feedback-driven control and harness optimization methods that use static analysis,
runtime errors, tests, and human feedback to revise code through
repeated execution
~\cite{huang2023agentcoder,ukai2024adacoder,Nunez2024AutoSafeCoder,li2026agentharness}.
This layer turns the interface in \S\ref{sec:foundations} into an
operational harness: planning controls the execution trajectory,
memory preserves state, tools expand the action space, and
feedback-driven adaptation closes the loop between failure and revision.

Finally, \emph{\textbf{Scaling the Harness: Multi-Agent Orchestration over Code}} (\S\ref{sec:mas}) extends the harness from
a single agent to collaborative ecosystems.
When multiple agents operate over code, the harness must not only
support individual reasoning and execution, but also coordinate
roles, share intermediate artifacts, maintain common state, and
verify collective progress.
We review multi-agent code-centric systems through agent roles
such as manager, planner, coder, reviewer, and tester;
collaboration modes such as programming, repair, debate,
red-teaming, and adversarial interaction; and workflow topologies
ranging from centralized coordination to distributed or streaming
collaboration
~\cite{wu2024autogen,Hong2023MetaGPT,Dong2024SelfCollaboration}.
This layer shows how code becomes a shared harness for
orchestrated autonomy: repositories, tests, traces, and structured
artifacts provide the common workspace through which agents
coordinate, inspect, and improve each other's behavior.

\begin{tcolorbox}[
  agentscope,
  float,
  floatplacement=t
]
This survey studies \emph{code as agent harness}: code-centered agent systems where reasoning, action, state, feedback,
and verification are organized around executable, inspectable, and
stateful programs.
We organize the literature up to 2026 into three connected layers:
\begin{itemize}
    \item \textbf{Harness Interface}: code enters the agent loop
    as a reasoning substrate, an action interface, and an
    environment representation.

    \item \textbf{Harness Mechanisms}: planning, memory, tool use,
    control, and harness optimization sustain code-centric agents over
    long-horizon execution and revision.

    \item \textbf{Scaling the Harness}: shared code artifacts,
    execution states, repositories, and structured workflows
    support coordination, review, and collective verification in
    multi-agent systems.
\end{itemize}

\end{tcolorbox}

Beyond the taxonomy, we examine how agent-initiated code interaction appears across five application domains. In coding assistance, agents author patches, tests, and issue-resolution workflows over live repositories~\cite{jimenez2023swe,yang2024swe,wang2024openhands}. In GUI and OS automation, agents synthesize and execute interface commands grounded in DOM trees, accessibility APIs, and executable evaluators~\cite{deng2023mind2webgeneralistagentweb,zhou2024webarenarealisticwebenvironment}. In scientific discovery, agents dynamically compose and execute hypothesis-testing pipelines spanning simulations, lab protocols, and data analysis~\cite{bran2023chemcrowaugmentinglargelanguagemodels,boiko2023autonomous,lu2024aiscientistfullyautomated,huang2025biomni}. In personalization and embodied control, agents author and revise executable policies, simulators, and skill libraries in response to environment feedback~\cite{ahn2022can,liang2023code,wang2023voyager}. We further outline open challenges for harness engineering, including evaluation beyond final task success, verification under incomplete feedback, regression-free harness improvement, consistent shared state across multiple agents, human oversight, and extensions to multimodal environments. This survey provides a roadmap for studying code not only as something agents generate, but as the runtime medium through which they execute, adapt, and coordinate reliable behavior.

\begin{tcolorbox}[agentcontrib]
\begin{itemize}
    \item \textbf{Conceptual framing}: We formalize
    \emph{code as agent harness}, reframing code from a generated
    artifact into the operational substrate of executable,
    verifiable, and stateful AI agent systems.

     \item \textbf{Taxonomy and synthesis}: We organize
   code as agent harness into three connected layers: harness
    interfaces, harness mechanisms, and scaling harness, and synthesize representative methods.

    \item \textbf{Applications and future agenda}:We connect the taxonomy
to real-world applications and outline key challenges in evaluation,
verification, safety, and coordination.
\end{itemize}
\end{tcolorbox}

\clearpage
\tableofcontents
\clearpage

\section{Harness Interface: Code for Reasoning, Acting, and
Environment Modeling}
\label{sec:foundations}

A harness turns a stateless language model into a functional 
agent by grounding its outputs in external execution, persistent 
state, and verifiable feedback. The most fundamental design 
question for any harness is therefore: \emph{what medium 
connects the model to its task environment?}

We argue that code is the answer. Unlike natural language, code 
is \emph{executable}, meaning model outputs become operations 
with formally verifiable outcomes; \emph{inspectable}, meaning 
intermediate computation is exposed as structured traces that 
the harness can read, store, and act upon; and \emph{stateful}, 
meaning the evolving program represents task progress in a 
persistent, modifiable form across steps. Crucially, these are 
not merely properties of code as a notation; they are properties 
that make code functional as a harness interface. Executability 
means the harness can verify what the model intended. 
Inspectability means failures can be diagnosed and fed back. 
Statefulness means the agent's interaction history is not lost 
between steps.

\paragraph{Scope boundary.}
We use \emph{code} broadly, but not metaphorically. In this
survey, code refers to executable or machine-checkable artifacts,
including programs, scripts, formal specifications, proof scripts,
API schemas, tool definitions, tests, repositories, simulators,
configuration files, and code-adjacent execution artifacts such as
traces and logs when they are produced by or consumed by executable
systems. By contrast, raw perception, physical state, human intent,
and model-internal latent reasoning are not themselves code.
They may be sensed, estimated, serialized, verified, or acted upon
through code, but they should not be conflated with the code
interface. This boundary is important because code as a harness
interface does not replace perception, embodiment, human goals, or
model reasoning; rather, it makes selected aspects of them
executable, inspectable, and stateful within the agent loop.

We organize this interface around three roles that code assumes in
agentic systems. \emph{Code for reasoning} externalizes internal
logic into verifiable computation, allowing external interpreters,
symbolic solvers, execution traces, or process rewards to check
and refine reasoning (\S\ref{subsec:reasoning}). \emph{Code for
acting} translates high-level intent into executable operations
grounded in embodied, GUI, software, or tool-use environments
(\S\ref{subsec:acting}). \emph{Code for environment modeling}
represents world state, transition dynamics, and feedback signals
through program states, repositories, simulators, tests, and logs
that agents can execute, edit, and query
(\S\ref{subsec:environment}). Overall, these roles define the
harness interface: code makes reasoning executable, action
programmable, and environment state inspectable.

\subsection{Code for Reasoning}
\label{subsec:reasoning}

A central role of the agent harness is to transform model
reasoning from transient text generation into executable and verifiable computation. Early prompting techniques such as pure chain-of-thought (CoT)~\cite{wei2022chain} perform reasoning and computation
entirely in \textit{natural language}, forcing the model to both decompose problems and execute intermediate operations within a single latent textual process.
While language models are often effective at proposing reasoning steps, they remain unreliable at faithfully carrying out symbolic, logical, or arithmetic computation~\cite{gao2023pal}. More importantly, purely textual reasoning provides the agent harness
with little ability to verify intermediate states, inspect execution behavior, or persist computational progress across steps.

\textit{Code-for-reasoning} thus introduces code as the execution interface
between the model and the harness, moving beyond purely text-based reasoning. The model generates executable programs that external runtimes, interpreters, symbolic solvers, or verification modules can execute and evaluate.
This separates high-level reasoning from low-level computation:
the model proposes procedures, while the harness executes them,
observes runtime behavior, stores intermediate states, and feeds
execution results into future reasoning.

Recent work further broadens this interface from program execution as an external calculator to execution artifacts as reusable reasoning signals. Inputs and outputs, execution traces, variable states, control-flow structures, and function-level tests can all serve as intermediate states that the harness verifies, scores, and feeds back into subsequent reasoning. Existing work can therefore be organized into three paradigms: program-delegated reasoning, formal verification and symbolic reasoning, and iterative code-grounded reasoning.
We detail each of them in the following subsections.

\begin{figure}[t!]
    \centering
    \includegraphics[width=\linewidth]{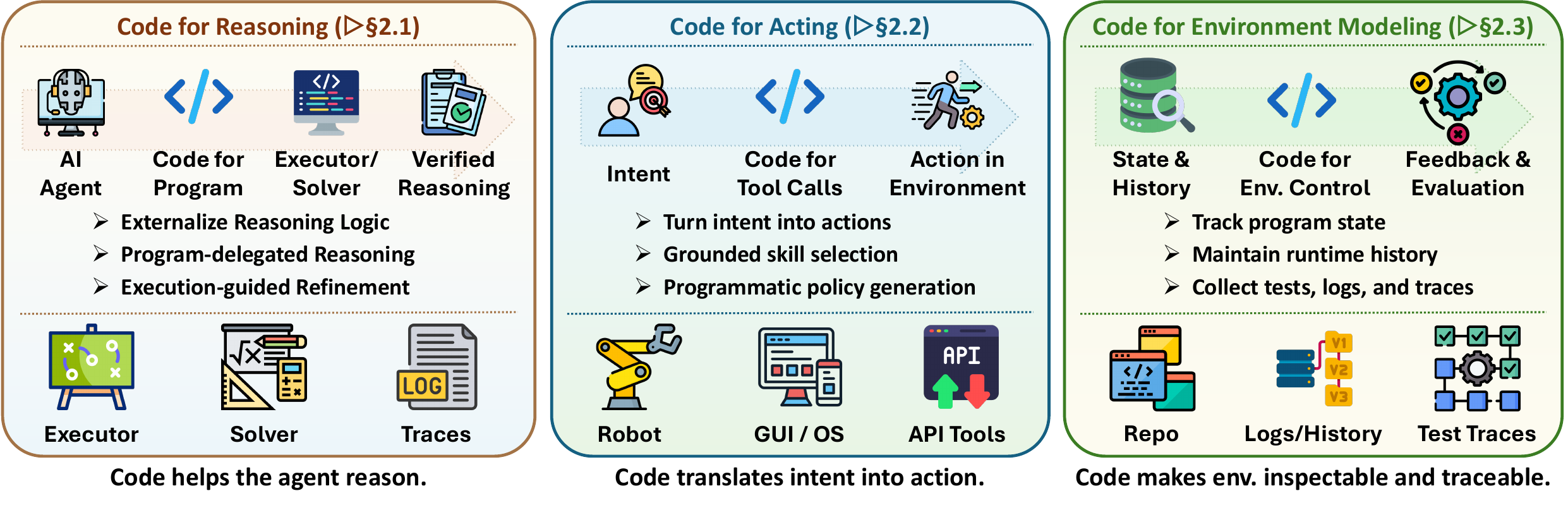}
    \caption{Overview of code as the harness interface, connecting agents to reasoning, action, and environment modeling through executable programs, tool calls, state tracking, and feedback traces.}
    \label{fig:sec2}
\end{figure}

\begin{figure}[t!]
    \centering
    \includegraphics[width=0.85\linewidth]{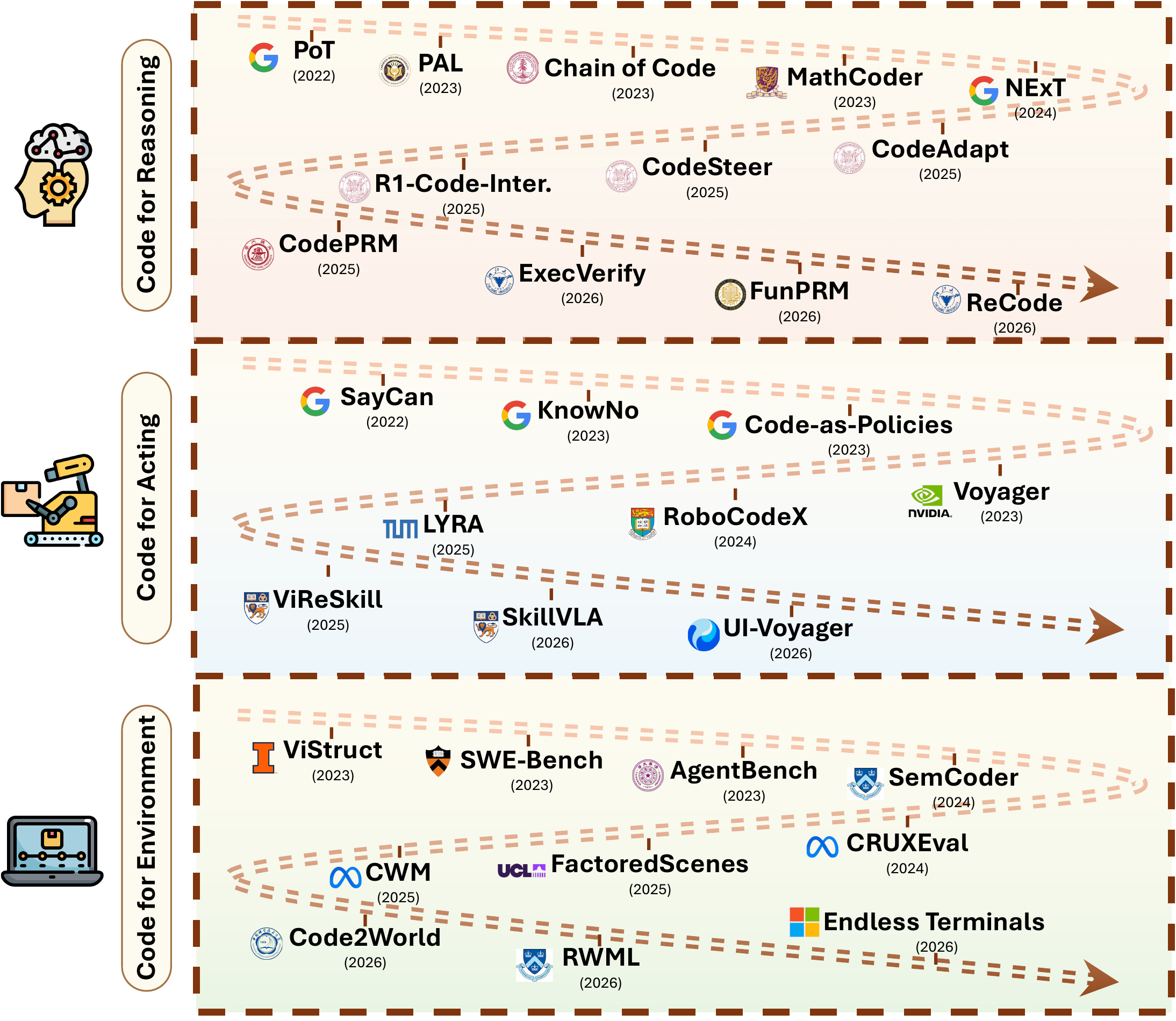}
    \caption{Roadmap of the harness interface, organized by code's role in reasoning, acting, and environment modeling, with representative works ordered chronologically within each role.}
    \label{fig:roadmap_sec2}
\end{figure}

\subsubsection{Program-Delegated Reasoning}
Program-delegated reasoning uses executable programs as the
primary interface between problem decomposition and computation.
Instead of relying solely on natural language reasoning, the
model generates code that external interpreters execute to
produce formally grounded outputs.
Early works~\cite{nye2021show,gao2023pal} demonstrate that
delegating computation to programs substantially improves
reliability by moving intermediate reasoning into structured,
verifiable execution traces.
Program-of-Thoughts (PoT) prompting~\cite{chen2022program}
further systematizes this paradigm by explicitly decomposing
reasoning into executable programs, followed by extensions such
as POET~\cite{pi2022reasoning} and
MathCoder~\cite{wang2023mathcoder}, which improve execution
fidelity and domain specialization.
Subsequent work investigates the conditions under which program
delegation is effective, including the role of execution
correctness, task structure, and runtime interaction.
For example, Chain of Code (CoC)~\cite{li2023chain} and
CIRS~\cite{bi2024program} analyze how executable reasoning
changes failure modes relative to pure language-based reasoning.
Later directions extend this interface beyond isolated task
execution. Cross-lingual reasoning
frameworks~\cite{payoungkhamdee2025towards} demonstrate that
program-based reasoning can generalize across linguistic
environments through shared executable structure, while
method-based reasoning~\cite{su2025method} introduces reusable
programmatic procedures that persist across tasks.
More recent systems such as
CodeAdapt~\cite{zhang2025code} further suggest that tightly
coupling language models with executable reasoning interfaces can surpass specialized reasoning-oriented models. Additionally, CodeI/O~\cite{pmlr-v267-li25t} transforms contextually grounded programs into code input-output prediction tasks, exposing reasoning primitives such as logic-flow planning, state-space search, decision-tree traversal, and modular decomposition while preserving procedural rigor through executable verification.

\subsubsection{Formal Verification and Symbolic Reasoning Interfaces}

Hybrid neural-symbolic methods combine flexible language-based inference with structured symbolic computation, using code and symbolic artifacts as persistent intermediate representations rather than treating programs as mere generated text. Early formulations such as Graph-of-Thoughts~\cite{besta2024graph} generalize chain-of-thought reasoning into graph-structured trajectories, enabling intermediate states to branch, merge, and be reused. Building on this direction, self-verifying reflection~\cite{yu2025self}, MA-LoT~\cite{wang2025ma}, and Socratic self-refine~\cite{shi2025ssr} introduce iterative verification loops in which symbolic consistency checks guide the refinement of generated solution paths.

Recent work further tightens the coupling between neural generation and symbolic execution through code-based interfaces. CodeSteer~\cite{chen2025codesteer} and Code-as-Symbolic-Planner~\cite{chen2025code} explicitly coordinate free-form language reasoning with executable symbolic operations, treating programs as structured substrates that the harness can inspect, transform, and execute across multiple stages. VisualCoder~\cite{chi-etal-2025-visualcoder} extends this idea by making program behavior visible through control-flow representations. By aligning generated reasoning with visual control-flow graphs and execution paths, it turns dynamic program behavior into an inspectable artifact for program-behavior prediction. Together, these methods broaden the neural-symbolic interface from textual code to multimodal execution artifacts that a harness can reference, validate, and reuse.

A complementary line of work uses machine-verifiable formal languages as the reasoning interface itself. Proof assistants such as Lean~\cite{moura2021lean}, Isabelle~\cite{nipkow2002isabelle}, and Coq~\cite{barras1999coq} provide formal proof languages based on rigorous logical foundations, enabling each derivation step to be checked by a verifier. Early LLM-based theorem-proving systems, including ReProver~\cite{yang2023leandojo}, DeepSeek-Prover~\cite{xin2025deepseek}, and TheoremLlama~\cite{wang2024theoremllama}, establish practical recipes for combining language models with proof-assistant feedback in mathematical reasoning. More recent systems, such as DeepSeek-Prover-V2~\cite{ren2025deepseek2}, Kimina-Prover~\cite{wang2025kimina}, MA-LoT~\cite{wang2025ma}, and Goedel-Prover-V2~\cite{lin2025goedel2}, improve this process through deliberative proof search, self-correction, and repeated proof generation and verification.
Formal verification interfaces are also expanding beyond theorem proving in mathematics. HybridReasoning~\cite{wang2025let} applies formal provers to support natural-language reasoning; Lean4Physics~\cite{li2025lean4physics} and PhysLib~\cite{physlib} extend Lean-based verification to physics; and VERINA~\cite{ye2025verina} and Goedel-Code-Prover~\cite{li2026goedel} adapt formal methods to code verification. Lean4Agent~\cite{wang2026lean4agent} further extends this trajectory to agentic systems by using Lean4 to model and verify agent workflows and trajectories. From the harness perspective, these systems show how formal languages can serve not only as reasoning tools, but also as executable contracts that constrain, certify, and audit agent behavior.

\subsubsection{Iterative Code-Grounded Reasoning}
Iterative code-grounded reasoning focuses on closed-loop
interaction between generation, execution, and feedback.
In these systems, reasoning is not a single-pass process, but an
iterative computational trajectory grounded in executable state
transitions.
Early work such as NExT~\cite{ni2024next} trains models to
anticipate execution behavior by reasoning over program traces,
thereby grounding intermediate reasoning in runtime semantics.
Related efforts~\cite{armengol2025cannot} similarly emphasize
that executable traces provide a richer supervision signal than
final textual outputs alone.
Building on this foundation, subsequent approaches introduce
explicit generate--execute--verify--refine loops.
Methods such as CodePRM~\cite{li2025codeprm} and
ORPS~\cite{yu2024reasoning} use execution outcomes to evaluate
and refine intermediate reasoning trajectories, enabling the
harness to guide reasoning through runtime feedback rather than
pure next-token prediction.
Along the same direction, systems such as
CYCLE~\cite{ding2024cycle} and
Self-Edit~\cite{zhang2023self} iteratively revise generated
solutions using execution-aware correction signals.
Reinforcement learning further strengthens this paradigm by
treating execution feedback as an optimization signal over
reasoning trajectories.
Methods such as CodeRL~\cite{le2022coderl},
CodeRL+~\cite{jiang2025coderl+}, and
RLTF~\cite{liu2023rltf} optimize functional correctness through
unit-test-based rewards, while approaches such as
StepCoder~\cite{dou2024stepcoder} incorporate fine-grained
compiler and runtime feedback during optimization.
RLEF~\cite{gehring2024rlef} formalizes this interaction as
policy optimization grounded in multi-step execution feedback,
allowing reasoning policies to adapt through iterative runtime
interaction.
More recent approaches move toward fully interactive reasoning
environments.
For example, EG-CFG~\cite{lavon2025execution} injects execution
signals directly during generation to support step-level
correction, while systems such as
R1-Code-Interpreter~\cite{chen2025r1} interleave reasoning and
multiple rounds of code execution within persistent interactive
sessions.

\begin{table}[t]
\centering
\caption{
Representative systems where code serves as a reasoning substrate.
}
\renewcommand{\arraystretch}{1.15}
\setlength{\tabcolsep}{4pt}
\footnotesize
\begin{tabularx}{\textwidth}{p{2.8cm}p{2.2cm}p{3.1cm}X}
\toprule
\textbf{Method} & \textbf{Mechanism} & \textbf{Reasoning Paradigm} & \textbf{Key Innovation} \\
\midrule
PoT~\cite{chen2022program} & Delegated & Hybrid comments & Merges code with natural language CoT \\
PAL~\cite{gao2023pal} & Delegated & Program-aided & Decouples logic from computation \\
CodeAdapt~\cite{zhang2025code} & Delegated & Generalizable logic & Code-enabled LLMs outperforming reasoning models \\
CodeI/O~\cite{pmlr-v267-li25t} & Delegated & I/O prediction & Converts code into verifiable input-output reasoning tasks \\
SATLM~\cite{ye2023satlm} & Formal & SAT/SMT solving & Uses symbolic solvers as machine-checkable reasoning backends \\
ReProver~\cite{yang2023leandojo} & Formal & Lean proof search & Combines LLM generation with proof-assistant feedback \\
Dpsk-Prover~\cite{xin2025deepseek} & Formal & Lean theorem proving & Trains LLMs for formal mathematical proof generation \\
Dpsk-Prover-V2~\cite{ren2025deepseek2} & Formal & Deliberative proving & Lean proof search through decomposition and self-correction \\
Goedel-Code-Prover~\cite{li2026goedel} & Formal & Lean code proof & Searches hierarchical Lean proofs for code verification \\
Lean4Agent~\cite{wang2026lean4agent} & Formal & Agent verification & Models and verifies agent workflows and trajectories in Lean4 \\
Chain of Code~\cite{li2023chain} & Hybrid & LMulator & Simulates non-executable semantic code \\
SATLM~\cite{ye2023satlm} & Hybrid & Formal Logic & Uses SAT/SMT solvers as reasoning backend \\
CodeSteer~\cite{chen2025codesteer} & Hybrid & Symbolic control & Explicitly transitions between symbolic code and neural text \\
VisualCoder~\cite{chi-etal-2025-visualcoder} & Hybrid & CFG-grounded & Aligns code reasoning with visual control-flow artifacts. \\
NExT~\cite{ni2024next} & Iterative & Trace-grounded & Anticipates execution behavior via program traces \\
MathCoder~\cite{wang2023mathcoder} & Iterative & Feedback-driven SFT & Interleaves code, output, and reflection \\
CodePRM~\cite{li2025codeprm} & Iterative & Process rewards & Learns reward functions over reasoning-execution trajectories \\
RLEF~\cite{gehring2024rlef} & Iterative & Multi-step RL & Optimizes policy directly using execution feedback \\
EG-CFG~\cite{lavon2025execution} & Iterative & Execution-guided & Integrates execution signals directly during generation \\
R1-Code-Int.~\cite{chen2025r1} & Iterative & Fully interactive & Autonomously interleaves reasoning and multiple executions \\
ExecVerify~\cite{tang2026execverifywhiteboxrlverifiable} & Iterative & Stepwise RL & Uses statement- and variable-level execution rewards. \\
FunPRM~\cite{zhang2026funprmfunctionasstepprocessreward} & Iterative & Function-step PRM & Treats functions as verifiable process-reward units. \\
ReCode~\cite{fan2026recodereinforcingcodegeneration} & Iterative & Process RL & Reinforces code generation with reasoning-process rewards \\
\bottomrule
\end{tabularx}
\label{tab:code_reasoning_systems}
\end{table}

\subsection{Code for Acting}
\label{subsec:acting}

Beyond reasoning, the agent must also connect the model to external environments where decisions produce real executable effects. At this stage, code no longer serves primarily as a medium for
computation, but as an action interface that converts model outputs into grounded operations such as tool invocations, robot-control policies, GUI actions, or software commands.
Through this interface, the harness translates high-level intent into executable behaviors that can interact with embodied, digital, and interactive environments.
The central challenge is therefore grounding: the harness must map abstract language outputs into executable behaviors that respect the constraints of the target environment, including embodiment limits, interface APIs, environment dynamics, and safety requirements. Unlike code-for-reasoning, where interpreters can often directly verify correctness, action execution occurs in partially observed and dynamically evolving environments, where failures may emerge through invalid state transitions, delayed feedback, or silent execution errors. For example, a robot may attempt to grasp an object outside its reachable workspace without producing an explicit runtime exception.

Importantly, executable action code is an interface to these
components, not a replacement for them. In embodied settings,
perception modules provide observations, affordance or feasibility
models estimate which actions are possible, motion planners and
controllers connect symbolic commands to sensors and actuators,
and safety layers constrain dangerous or invalid behavior. In GUI
and software settings, the analogous components include screen
parsers, DOM or accessibility trees, backend APIs, user-intent
models, permission systems, and programmatic validators. Code sits
between the model and these components: it serializes observations,
calls grounding and planning modules, invokes executable actions,
and exposes validation results back to the harness.

\textit{Code-for-acting} therefore introduces structured
executable programs as the control interface between the model
and the environment, allowing the harness to execute, monitor,
validate, reuse, and refine actions through interaction feedback.
This interface can be realized in different forms: a predefined
skill library, a generated control policy, a persistent skill
memory, a GUI/API tool protocol, or an explicit action-validation
harness. AutoHarness~\cite{lou2026autoharnessimprovingllmagents}
makes the last form explicit by automatically synthesizing a code
harness that mediates between the LLM and the environment,
filtering invalid actions before execution. This highlights the
core harness view of code-for-acting: code is not only the action
to be executed, but also the executable boundary that connects
model intent to perception, grounding, affordance estimates,
controllers, APIs, actuators, and safety constraints.

\begin{table}[t]
\centering
\caption{Representative systems where code serves as an action interface.}
\renewcommand{\arraystretch}{1.15}
\setlength{\tabcolsep}{4pt}
\footnotesize
\begin{tabularx}{\textwidth}{p{2.7cm}p{1.7cm}p{3.0cm}X}
\toprule
\textbf{Method} & \textbf{Mechanism} & \textbf{Action Paradigm} & \textbf{Key Innovation} \\
\midrule
AutoHarness~\cite{lou2026autoharnessimprovingllmagents} & Harness Gen. & Action validation & Synthesizes code harnesses that mediate model actions and filter invalid environment interactions \\
SayCan~\cite{ahn2022can} & Skill Selec. & Affordance-based & Links LLM plans to physical feasibility \\
KnowNo~\cite{ren2023robots} & Skill Selec. & Conformal prediction & Calibrates planner uncertainty for ambiguous instructions \\
SkillVLA~\cite{zhai2026skillvla} & Skill Selec. & Bimanual grounding & Extends grounding to combinatorial skill reuse \\
BOSS~\cite{zhang2023bootstrap} & Skill Selec. & Skill bootstrapping & Synthesizes new executable skill chains via guided practice \\
LLM-Guided Traj.~\cite{ha2023scaling} & Skill Selec. & Trajectory generation & Generates diverse manipulation trajectories and executable success conditions \\
LRLL~\cite{tziafas2024lifelong} & Skill Selec. & Lifelong grounding & Evolving skill interface via memory and self-exploration \\
CaP~\cite{liang2023code} & Policy Gen. & Hierarchical Python & Generates reactive robot control policies \\
RoboCodeX~\cite{mu2024robocodex} & Policy Gen. & Multimodal tree & Synthesizes tree-structured code across navigation \\
Code-BT~\cite{zhang2025codebt} & Policy Gen. & Behavior-tree & Imposes rule constraints via code-to-behavior-tree planning \\
ALRM~\cite{santos2026alrm} & Policy Gen. & Closed-loop control & Integrates programmatic generation with ReAct execution \\
CP-Agent~\cite{szeider2025cp} & Policy Gen. & Constraint solving & Uses persistent execution loops for formal constraint-model repair \\
Robot-Code Sim.~\cite{wang2025llm} & Policy Gen. & Static simulation & Uses LLMs as static simulators for robot code evaluation \\
GenSwarm~\cite{ji2026genswarm} & Policy Gen. & Multi-robot control & Coordinates policy generation and deployment across robotic agents \\
NormCode~\cite{guan2025normcode} & Policy Gen. & Governed interface & Enforces auditability and data isolation through semi-formal code \\
RACAS~\cite{ashley2026racas} & Policy Gen. & Cooperative control & Robot-agnostic architecture for closed-loop cooperative agents \\
Voyager~\cite{wang2023voyager} & Lifelong & Skill Library & Autonomous curriculum for open-ended tasks \\
LYRA~\cite{meng2025growing} & Lifelong & Human-in-loop & Encodes human corrections into reusable structured skills \\
ViReSkill~\cite{kagaya2025vireskill} & Lifelong & Vision-grounded & Replanning on failure using a skill-memory cache \\
UI-Voyager~\cite{lin2026ui} & Lifelong & Self-evolving & Rejection fine-tuning and self-distillation for mobile GUI agents \\
SkillsCrafter~\cite{wang2026lifelong} & Lifelong & Continual skills & Mitigates forgetting as executable manipulation skills accumulate \\
\bottomrule
\end{tabularx}
\label{tab:code_acting_systems}
\end{table}

\subsubsection{Grounded Skill Selection}
Grounded skill selection studies how the agent maps high-level language intent into executable behaviors through reusable skill interfaces. Rather than generating low-level actions directly, these systems treat the environment as a collection of executable capabilities
that the agent harness can invoke, compose, and refine under environmental constraints. SayCan~\cite{ahn2022can} establishes the core paradigm by coupling language planning with grounded skill execution, allowing the agent to select actions based not only on semantic relevance but also embodiment feasibility.
Subsequent work extends this execution interface in several directions. KnowNo~\cite{ren2023robots} introduces uncertainty-aware control through conformal prediction, enabling the harness to detect ambiguous states and trigger clarification before unsafe
execution. BOSS~\cite{zhang2023bootstrap} addresses the rigidity of fixed skill libraries by using language-guided practice to synthesize
new executable skill chains, allowing the harness to expand its
action space over time.
Similarly, \cite{ha2023scaling} tackles the data bottleneck of
grounded interaction by using LLM-guided generation to construct
diverse manipulation trajectories and executable success
conditions for automatic retry and relabeling.
Beyond static execution, LRLL~\cite{tziafas2024lifelong}
introduces memory and self-guided exploration to maintain a
persistent and evolving skill interface across tasks.
Finally, SkillVLA~\cite{zhai2026skillvla} extends this paradigm
to combinatorial bimanual interaction, emphasizing that grounded
action interfaces must support structured skill reuse and
recomposition under increasingly complex embodiment settings.

\subsubsection{Programmatic Policy Generation}
Programmatic policy generation treats code itself as the control
interface between the model and the environment.
Instead of selecting from predefined skills, the harness directly
materializes executable policies as programs that specify control
logic, perception-conditioned branching, feedback loops, and API
interaction.
CaP~\cite{liang2023code} crystallizes this paradigm by framing
LLM-generated Python programs as executable robot policies.
Building on this idea, RoboCodeX~\cite{mu2024robocodex}
introduces multimodal and tree-structured code generation to
support more complex manipulation and navigation behaviors.
Subsequent work focuses on scaling the interaction substrate.
RoboPro~\cite{xie2025robotic} synthesizes executable policy code
from large-scale in-the-wild videos, while
Code-BT~\cite{zhang2025codebt} compiles generated programs into
behavior-tree controllers that support constrained execution and
iterative runtime feedback.
Beyond robotics, CP-Agent~\cite{szeider2025cp} demonstrates that
persistent execution loops can support formal constraint-solving
agents through iterative execution and repair.
To reduce dependence on expensive physical environments,
\cite{wang2025llm} configures language models as static execution
simulators for robot code evaluation.
GenSwarm~\cite{ji2026genswarm} further extends programmatic
control to multi-agent robotic systems, where the harness must
coordinate policy generation, constraint analysis, and deployment
across multiple embodied agents.
At the systems level, NormCode~\cite{guan2025normcode}
emphasizes governance and auditability by introducing a
semi-formal programming interface with enforced data isolation,
allowing execution traces and control logic to remain
inspectable and constrained.
Finally, ALRM~\cite{santos2026alrm} and
RACAS~\cite{ashley2026racas} consolidate these ideas into
persistent closed-loop control architectures that integrate code
generation, execution, monitoring, and iterative interaction
within unified agent harnesses.

\subsubsection{Lifelong Code-Based Agents}
Lifelong code-based agents study how executable interaction
interfaces can persist, evolve, and accumulate capabilities over
long-horizon interaction.
In these systems, code is not only an execution mechanism, but
also a persistent memory substrate through which the harness
stores reusable behaviors, interaction traces, and environment
knowledge.
Voyager~\cite{wang2023voyager} establishes this paradigm through
an automatic curriculum and continually expanding executable
skill library for open-ended interaction in Minecraft.
Extending this idea to embodied environments,
LRLL~\cite{tziafas2024lifelong} introduces persistent memory,
self-guided task exploration, and skill abstraction to overcome
the limitations of fixed policy libraries without requiring
gradient updates.
A central challenge in lifelong harnesses is that interaction
feedback and corrections are often transient and difficult to
reuse.
LYRA~\cite{meng2025growing} addresses this issue by converting
human corrections into reusable executable skills and
retrieval-augmented memory structures.
Similarly, ViReSkill~\cite{kagaya2025vireskill} combines
vision-grounded replanning with skill-memory caching to maintain
stable interaction under environmental failures and output
variability.
Recent work further focuses on continual adaptation and
self-evolution under persistent deployment.
SkillsCrafter~\cite{wang2026lifelong} introduces continual
language-conditioned manipulation structures to mitigate
catastrophic forgetting as executable capabilities accumulate,
while UI-Voyager~\cite{lin2026ui} generalizes the self-evolving
interaction paradigm to GUI agents through failure-driven
adaptation and self-distillation.
Together, these systems move beyond one-shot execution toward
persistent agent harnesses that continuously expand, refine, and
reuse executable interaction interfaces over time.

\subsection{Code for Environment}
\label{subsec:environment}

The agent must also maintain an explicit representation
of the environment with which the agent interacts.
Without such a representation, the environment is exposed to the agent only indirectly through textual observations, API returns, or sparse feedback signals.
As a result, environment state often remains implicit, transient, and difficult to verify, making it challenging to
track state transitions, evaluate interaction outcomes, or reuse
past interaction history across long-horizon tasks.
This limitation becomes particularly severe in complex software,
robotic, and multi-step interactive environments, where
successful interaction depends on maintaining consistent world
state and grounded feedback over time.

\textit{Code-for-environment} addresses this limitation by
introducing executable programs as the environment interface
itself.
Instead of treating the environment as an opaque external process,
these systems materialize environment structure and dynamics
through computational artifacts such as simulators,
repositories, tests, execution traces, logs, and state-transition
programs.
This allows the agent to explicitly store, inspect, execute,
and modify environment state throughout interaction.
Representing environments through executable code provides two
major advantages.
First, executable environments expose verifiable state
transitions, allowing the agent to evaluate interaction
outcomes through execution rather than ambiguous natural-language
judgment.
Second, code-based environments are persistent and
modifiable that agents can query, simulate, edit,
and refine during interaction.
Rather than interacting with an opaque world solely through language, agent harness can ground reasoning and action in explicit computational state and runtime dynamics. 
Existing work in this direction can be organized into four
paradigms: structured world representations, execution-trace world
modeling, code-grounded evaluation environments, and verifiable
environment construction.

\begin{table}[t]
\centering
\caption{Representative systems where code serves as an environment representation.}
\renewcommand{\arraystretch}{1.15}
\setlength{\tabcolsep}{4pt}
\footnotesize
\begin{tabularx}{\textwidth}{p{2.8cm}p{2.0cm}p{3.1cm}X}
\toprule
\textbf{Method} & \textbf{Mechanism} & \textbf{Environment Paradigm} & \textbf{Key Innovation} \\
\midrule
ViStruct~\cite{chen2023vistruct} & Structured & Class/object hierarchy & Encodes visual scenes as data structures \\
FactoredScenes~\cite{hsu2025programs} & Structured & Room programs & Composes object/relation functions for 3D layout generation \\
PoE-World~\cite{piriyakulkij2025poe} & Structured & Programmatic experts & Scales symbolic world models beyond simple grid-worlds \\
Code2World~\cite{zheng2026code2world} & Structured & Render-aware RL & Re-frames GUI state prediction as renderable HTML generation \\
SemCoder~\cite{ding2024semcoder} & Trace-based & Semantic alignment & Pairs code with detailed execution traces \\
WorldCoder~\cite{tang2024worldcoder} & Trace-based & Model-based RL & Synthesizes transition and reward models \\
CWM~\cite{copet2025cwm} & Trace-based & Open-weights trace & Trains large LLMs natively on program execution traces \\
RWML~\cite{yu2026reinforcement} & Trace-based & Self-supervised RL & Aligns simulated next states with realized environment states \\
AWM~\cite{wang2026agent} & Trace-based & World-modeling & Aligns multiple executable world models across tasks \\
WorldMind~\cite{ren2026aligning} & Trace-based & Model fusion & Coordinates executable world models from knowledge sources \\
SWE-bench~\cite{jimenez2023swe} & Evaluation & Repo-level testing & Uses unit tests as objective world states \\
AgentBench~\cite{liu2023agentbench} & Evaluation & Multi-env interaction & Benchmarks across OS, databases, and games \\
CRUXEval~\cite{gu2024cruxeval} & Evaluation & Execution tasks & Benchmarks functional input and output prediction \\
End Terms.~\cite{gandhi2026endless} & Evaluation & Procedural RL envs & Automates generation of terminal-use evaluation tasks \\
InterCode~\cite{yang2023intercode} & Evaluation & Interactive execution & Frames coding tasks as actions with sandbox feedback \\
LiveCodeBench~\cite{jain2024livecodebench} & Evaluation & Live coding eval & Continuously updates execution-based evaluation pipelines \\
CRUXEval-X~\cite{xu2025cruxeval} & Evaluation & Multilingual execution & Extends input-output execution evaluation across languages \\
CoRe~\cite{xie2025core} & Evaluation & Runtime reasoning & Evaluates code reasoning through execution-centered tasks \\
CodeGlance~\cite{wang2026codeglance} & Evaluation & Multimodal code eval & Evaluates code understanding under visual and structural settings \\
SWE-smith~\cite{yang2025swesmithscalingdatasoftware} & Construction & Synthetic SWE envs & Generates repository-level tasks and execution environments \\
EnvScaler~\cite{song2026envscalerscalingtoolinteractiveenvironments} & Construction & Tool-interactive envs & Synthesizes tool-use environments with programmatic validators \\
\bottomrule
\end{tabularx}
\label{tab:code_environment_systems}
\end{table}

\subsubsection{Structured World Representations}
Structured world representations model environments through
explicit programmatic structures that the agent can execute,
inspect, and manipulate.
Rather than representing the environment solely through latent
embeddings or textual descriptions, these approaches encode world
state, object relations, spatial layouts, and interaction
dynamics as structured computational artifacts.
For example, ViStruct~\cite{chen2023vistruct} uses
programming-language structure as an explicit interface for
visual structural knowledge extraction, enabling multi-granular
visual events to be represented through consistent executable
structures.
FactoredScenes~\cite{hsu2025programs} similarly models indoor
environments as compositional ``room programs,'' where reusable
object and relation functions define physically consistent scene
layouts.
Extending this idea to scalable symbolic world modeling,
PoE-World~\cite{piriyakulkij2025poe} introduces a compositional
framework that combines many small programmatic experts to
represent increasingly complex environment dynamics.
More recent systems broaden structured environment interfaces to
high-fidelity interactive worlds.
Code2World~\cite{zheng2026code2world} reframes GUI state
prediction as renderable HTML generation, allowing environment
transitions to be represented and evaluated through executable
rendering code.
Code2Worlds~\cite{zhang2026code2worlds} further extends this
paradigm to 4D simulated environments through language-to-simulation
program generation, where physics-aware execution loops
reduce semantic-physical inconsistencies during environment
construction and interaction.

\subsubsection{Execution-Trace World Modeling}
Execution-trace world modeling studies how the agent can learn
environment dynamics directly from executable interaction traces.
Instead of treating execution merely as a final evaluation step,
these approaches model runtime transitions themselves as the
primary representation of environment behavior.
SemCoder~\cite{ding2024semcoder} bridges static programs and
runtime semantics by training language models to reason about
functional behavior, statement-level execution effects, and
input-output transitions.
Building on this perspective, Code World
Model~(CWM)~\cite{copet2025cwm} learns predictive world models
directly from program traces, enabling the agent to anticipate
future environment states through executable dynamics.
WorldCoder~\cite{tang2024worldcoder} further introduces a
model-based interaction framework in which the agent explicitly
writes and updates executable world models represented as Python
programs.
Rather than storing environment knowledge implicitly in model
parameters alone, the agent maintains editable computational
representations that can be executed, revised, and reused during
planning and interaction.
Subsequent work extends this paradigm toward continual and
interactive world-model adaptation.
RWML~\cite{yu2026reinforcement} combines execution traces with
reinforcement learning to refine environment dynamics through
runtime interaction, while
AWM~\cite{wang2026agent} and
WorldMind~\cite{ren2026aligning} study how multiple executable
world models can be aligned, fused, and coordinated across tasks
and knowledge sources.

\subsubsection{Code-Grounded Evaluation Environments}
Code-grounded evaluation environments use executable systems as
the interface for measuring agent behavior and interaction
quality.
Unlike static benchmarks based solely on textual outputs, these
environments expose explicit runtime state transitions, execution
feedback, and verifiable interaction outcomes that the agent
can directly observe and evaluate.
InterCode~\cite{yang2023intercode} establishes this paradigm by
reframing coding tasks as interactive execution environments,
where code acts as actions, execution feedback serves as
observations, and sandboxed runtimes provide grounded
interaction.
CRUXEval~\cite{gu2024cruxeval} further evaluates program
understanding through executable input-output prediction tasks,
while LiveCodeBench~\cite{jain2024livecodebench} introduces
continuously updated evaluation pipelines that assess execution,
self-repair, and runtime reasoning capabilities under evolving
problem distributions.
SWE-bench~\cite{jimenez2023swe} extends executable evaluation to
real-world software repositories, where agents must modify
large-scale codebases and are evaluated through repository-level
unit-test execution rather than textual correctness alone.
More broadly, AgentBench~\cite{liu2023agentbench} demonstrates
that executable interaction environments can evaluate reasoning
and decision-making across diverse embodied and digital tasks.
Subsequent benchmarks such as
CRUXEval-X~\cite{xu2025cruxeval},
CoRe~\cite{xie2025core},
GeoGramBench~\cite{luo2025geogrambench},
CodeGlance~\cite{wang2026codeglance}, and
Endless Terminals~\cite{gandhi2026endless} further expand this
paradigm toward multilingual, multimodal, and continuously
interactive evaluation settings, where runtime interaction rather
than static answer matching becomes the primary evaluation
interface.

\subsubsection{Verifiable Environment Construction}
A newer direction treats executable environments not only as
benchmarks to evaluate agents, but as harness artifacts that can
be synthesized, scaled, and validated programmatically. This is
especially important for long-horizon agents, where the harness
must provide not only a task prompt, but also a runnable state,
transition dynamics, feedback channels, and verification oracles.
SWE-smith~\cite{yang2025swesmithscalingdatasoftware} scales
software-engineering agent data by constructing repository-level
tasks and execution environments from existing codebases, turning
software repositories into reproducible program worlds for agent
training and evaluation. EnvScaler~\cite{song2026envscalerscalingtoolinteractiveenvironments}
extends this idea beyond software engineering by programmatically
synthesizing tool-interactive environments together with scenarios
and rule-based trajectory validators. From the harness perspective,
these methods make the environment interface itself an object of
construction: code specifies not only what the agent edits or
executes, but also the state transitions, tool affordances, and
verifiers that determine whether an interaction has succeeded.

\section{Harness Mechanisms: Planning, Memory, Tool Use, Control, and Optimization}
\label{sec:modules}

Harness mechanisms form the central systems layer that makes code-harnessed agents reliable beyond a
single generation step. Once code enters the agent loop, software generation is no longer only a problem of producing correct programs from a prompt. It becomes an interaction among the model, mutable task state, and human-designed harness infrastructure. The model provides judgment: it decomposes goals, selects actions, interprets feedback, and decides when to revise. Mutable state records repository evidence, working context, execution traces, validation results, memories, and intermediate beliefs about the task. The harness infrastructure exposes tools and execution substrates, persists and compacts state, constrains actions through policies and permission tiers, routes feedback, and verifies whether each state transition is acceptable. From this perspective, harness mechanisms are not isolated add-on modules, but coordinated control surfaces that turn model decisions into bounded, observable, and revisable changes in an executable environment.
In its basic form, code allows the agent to call existing executable interfaces. 
Further, the agent can dynamically author task-specific executable interfaces. These agent-authored artifacts make the harness more adaptive because they allow the execution environment to be reshaped around the current task. However, dynamically authored code does not replace the broader human-designed harness infrastructure. Reliability still depends on model-side judgment together with human-designed policies, sandbox boundaries, permission tiers, verification oracles, audit logs, and human-review gates. Code therefore serves as an executable medium inside the harness, while the harness remains the larger policy-governed system that decides what code may be executed, trusted, persisted, reused, or promoted into future workflows.

In this section, we review five interacting categories of harness mechanisms for code agents. Planning (\S~\ref{sec:planning}) organizes long-horizon task execution by externalizing goals into decompositions, structural constraints, search trajectories, or workflow-level orchestration. Memory and context engineering (\S~\ref{sec:memory}) manage mutable state across long interactions by preserving working context,
retrieving repository evidence, storing reusable experience, supporting shared histories, and offloading state beyond the active context window. Tool usage (\S~\ref{sec:tool}) connects the agent to governed executable interfaces, including APIs, repositories, terminals, sandboxes, verification tools, and workflow orchestrators. Harness control through the Plan-Execute-Verify loop (\S~\ref{sec:debug}) reframes feedback-guided debugging as a broader control process: plans form contracts over intended changes, execution applies them inside sandboxed and permissioned environments, and verification uses deterministic sensors and human-review gates to decide whether the state should be accepted, revised, escalated, or rolled back. Finally, agentic harness engineering (\S~\ref{sec:ahe}) studies how the harness itself can be measured and improved through deep telemetry, evolution agents, replay-based evaluation, and governed harness mutation.

\begin{figure}[t]
    \centering
    \includegraphics[width=\linewidth]{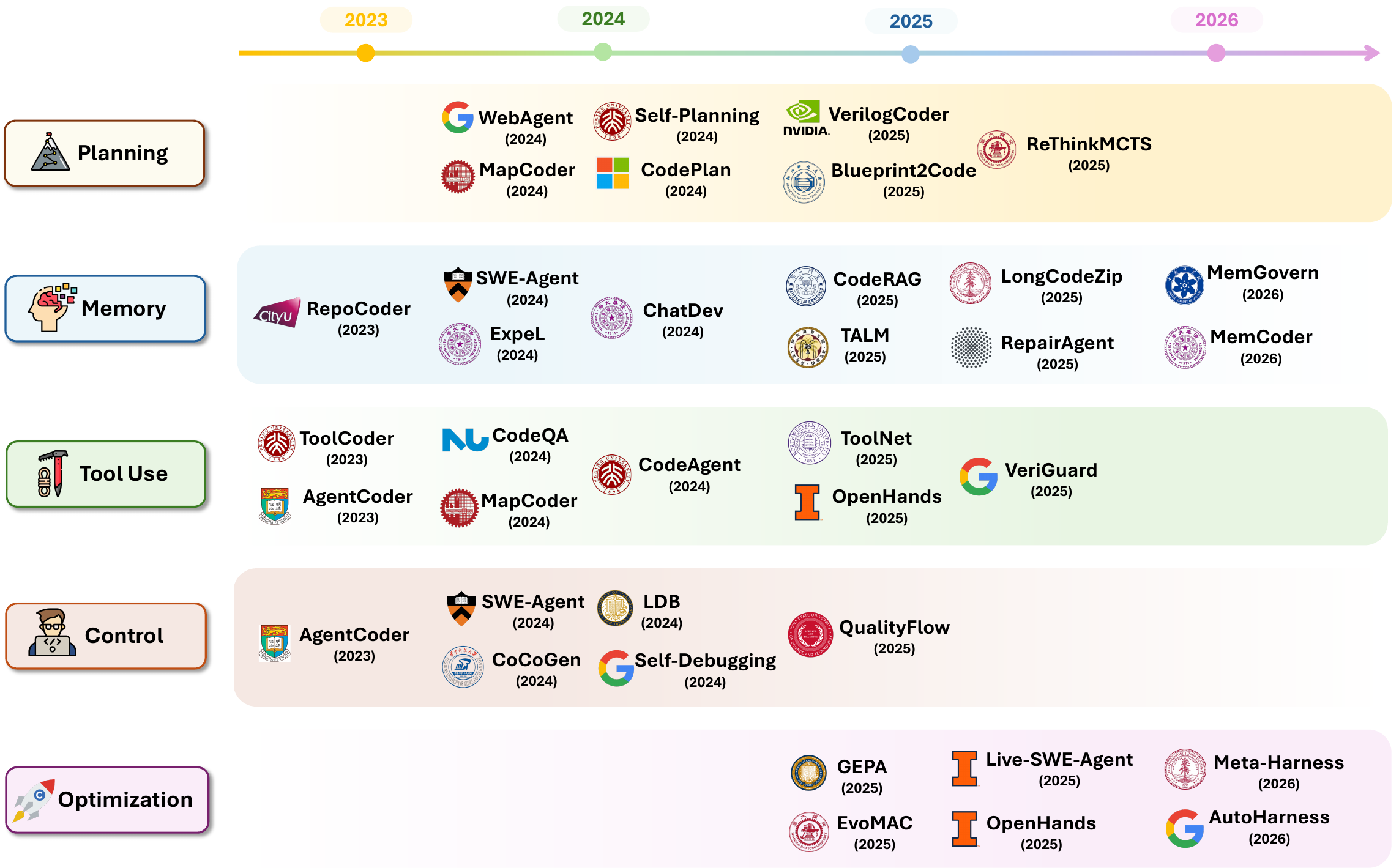}
    \caption{A roadmap overview of agent harness mechanisms.}
    \label{fig:roadmap_sec3}
\end{figure}

\subsection{Planning for Agent Harness}\label{sec:planning}

Planning plays a central role in agentic harness because real-world software engineering tasks rarely admit a direct one-shot mapping from natural language intent to correct implementation. From the harness perspective, planning is not merely an internal reasoning capability of the LLM, but a form of \emph{harness control}: it structures how the agent externalizes intent into executable steps, schedules interactions with code artifacts and tools, and regulates the trajectory of reasoning, execution, and revision over time. Beyond generating code tokens, an effective agent harness must organize long-horizon problem solving into a coherent course of action, deciding what intermediate goals to pursue, in what order to execute them, what artifacts to inspect or modify, and how to revise the trajectory when execution feedback reveals errors, missing dependencies, or violated constraints. This need becomes especially pronounced in repository-level editing, web interaction, competitive programming, and hardware design, where the agent must operate over large action spaces, sparse feedback, and deeply interdependent subproblems. In such settings, a fundamental challenge arises \textbf{between the complexity of the target task and the limited reliability of unconstrained agent execution}: without an explicit planning mechanism as harness control, the agent may commit too early to brittle solution paths, overlook latent dependencies, or fail to coordinate reasoning, retrieval, execution, and revision into a stable workflow.

Early planning-oriented systems mainly treated planning as a linear decomposition step, where the model first produced a natural-language solution outline and then translated it into code. As code agents were applied to more complex environments, however, planning gradually evolved from a simple pre-generation scaffold into a richer harness-level control mechanism. It can be grounded in repository structure or external knowledge to constrain the agent's action space, expanded through explicit search over multiple candidate trajectories to improve robustness, or distributed across specialized agent roles and feedback loops to coordinate execution at the system level. Based on the \textbf{primary locus where harness control is realized}, we categorize existing planning methods in code agents into four types: \textit{linear decomposition planning}, \textit{structure-grounded planning}, \textit{search-based planning}, and \textit{orchestration-based planning}.

\begin{figure}[t]
    \centering
    \includegraphics[width=0.8\linewidth]{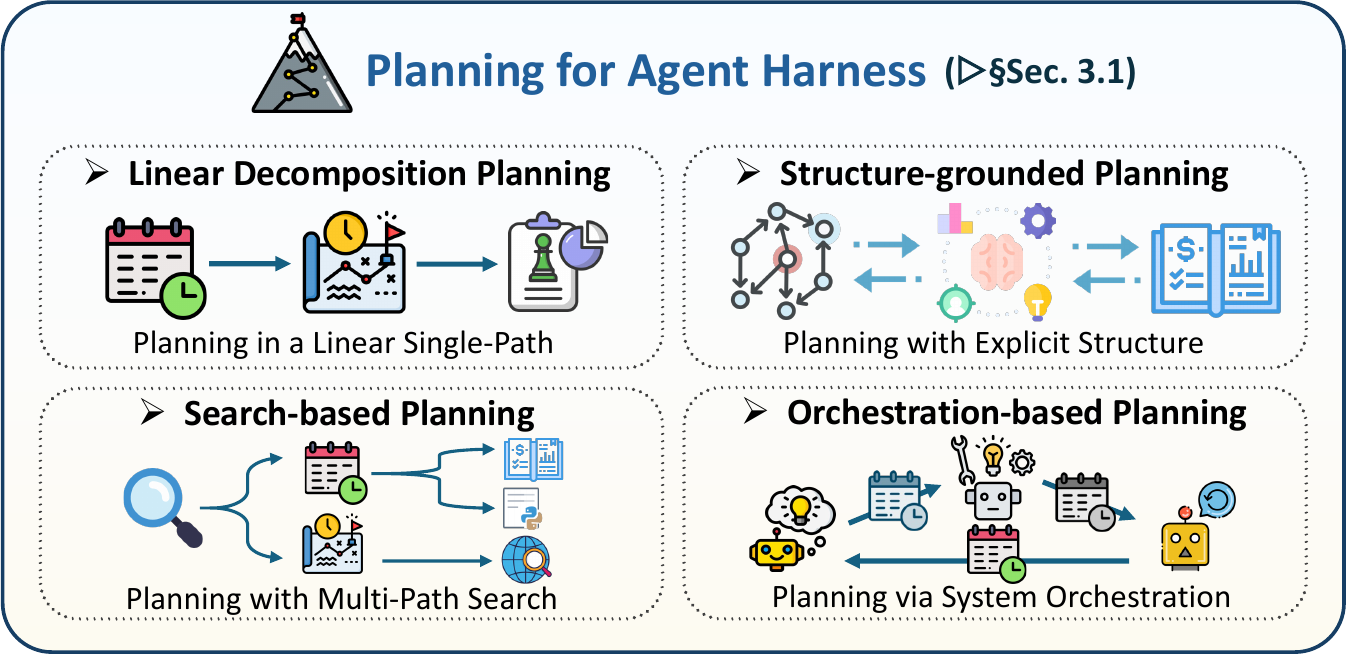}
    \caption{Overview of planning mechanisms for agent harnesses.}
    \label{fig:modules-planning}
\end{figure}

\subsubsection{Linear Decomposition Planning} 
In this planning paradigm, the agent first produces a single explicit, executable sequence of steps, and then carries out generation by following this decomposition~\cite{huang2024knowledge,jiang2024selfplanning,gur2023webagent,linearplan1,zhang2025linearplan2}. A lightweight precursor of this pattern is ReAct~\cite{yao2023reactsynergizingreasoningacting}, where the agent interleaves thoughts, actions, and observations in a serial trajectory. In this framework, each reasoning step externalizes the current subgoal and constrains the next action, turning the trajectory itself into a stepwise harness for control. 
This pattern is most directly instantiated in Self-Planning~\cite{jiang2024selfplanning}: the model first decomposes the intent into concise, high-level numbered steps, and then generates code step by step under the guidance of this plan. Plan-And-Act~\cite{erdogan2025plan} further makes this harness explicit by separating a planner, which produces structured high-level plans: the planner repeatedly refreshes the linear scaffold as new observations arrive, allowing the planning strategy to preserve task-level control while adapting to environmental feedback. WebAgent~\cite{gur2023webagent} extends this idea to web automation: it decomposes a user instruction into successive sub-instructions, summarizes task-relevant HTML conditioned on the current subgoal, and then synthesizes executable Python actions from that linear sub-instruction sequence. KareCoder~\cite{huang2024knowledge} follows a similar template in a knowledge-augmented setting, where the model first constructs a knowledge-aware, step-by-step prompt from an external knowledge library and then uses this prompt to generate code, making planning a structured intermediate layer between problem understanding and implementation. 
Recent industrial practice shows that this linear scaffold can be lifted from an ephemeral prompt artifact to a persistent harness object. In long-horizon coding workflows, files such as \texttt{PLAN.md}, \texttt{Implement.md}, and status logs record milestones, acceptance criteria, validation commands, and recovery rules, allowing the agent to reload, update, verify, and document progress across context resets or multi-session execution~\cite{openai2025execplans,openai2026codexlonghorizon}. In this view, planning is no longer merely an internal reasoning trace, but a filesystem-backed control object: it can be reviewed by humans, versioned with Git, consumed by subagents, and used as the source of truth for implementation. The main limitation remains that these methods typically commit to a single decomposition trajectory: when the initial plan is incomplete or misaligned, the harness can improve persistence and auditability, but it still provides limited exploration beyond the chosen path.

\subsubsection{Structure-grounded Planning}
In this line of work, the agent does not derive its action sequence solely from a free-form natural language prompt, but instead grounds planning in an explicit structured representation of the task environment, such as dependency graphs, repository graphs, circuit graphs, or knowledge graphs. These structures act as natural harness scaffolds: they expose relevant entities, encode dependency relations, and guide the order in which subtasks should be generated, revised, or verified. For example, CodePlan~\cite{bairi2024codeplan} constructs a plan graph over edit obligations and derives new steps through dependency analysis and change-impact propagation. Meanwhile, repository understanding methods ~\cite{luo2025rpg,chen2025locagent,tao2025cgm,luo2025rpg} convert codebases into heterogeneous graphs or text-rich code graphs, then use graph-integrated reasoning to localize relevant entities and condition downstream generation on structural dependencies rather than flat text context. GraphCodeAgent~\cite{li2025graphcodeagent} extends this idea with a dual-graph harness, where a Requirement Graph captures relations among natural-language requirements and a Structural-Semantic Code Graph captures repository dependencies. 
The same principle also appears in recent agent-native repository practices. Files such as architecture notes, API specifications, and testing guides turn project knowledge into persistent, inspectable, and version-controlled artifacts that the agent can consult before acting~\cite{agentsmd2025,openai2026agentsmd,anthropic2025claudememory}. This broadens structure-grounded planning beyond graph construction: the relevant structure determines explicit rules, build commands, directory boundaries, coding conventions, and design constraints, thereby promoting a coherent and stable harness control over the programs. 
Specialized domains follow the same pattern~\cite{wang2026domagent,ho2025verilogcoder}. VerilogCoder~\cite{ho2025verilogcoder} grounds subtask planning in a Task and Circuit Relation Graph so that each subtask is enriched with signals, transitions, and examples, while DomAgent~\cite{wang2026domagent} uses knowledge graphs to combine top-down structured knowledge with bottom-up examples for domain-specific code generation. Overall, these works show that structure-grounded planning improves coherence, dependency awareness, and long-horizon consistency by turning project or domain knowledge into explicit and inspectable harness objects that guide the agent's behavior over time.

\subsubsection{Search-based Planning}
Search-Based Planning allocates inference-time compute to systematically explore, evaluate, and select among multiple candidate solution paths. Rather than committing the agent to a single plan, the key idea is to expand the decision space and use feedback to control which alternatives should be pursued, revised, or discarded. A first group of methods~\cite{wang2024planning,li2025rethinkmcts} instantiates this harness in the thought space. Instead of directly writing code, they first branch over high-level observations, strategies, or reasoning traces, with the goal of increasing conceptual diversity before implementation. In this view, better planning comes from covering a broader idea space and using feedback to refine reasoning itself, rather than merely repairing final code. A second group~\cite{li2025codetree,ni2024treeofcode,dainese2024codegenerating,aggarwal2025dars} performs search in the trajectory space of coding actions: these methods model coding as a branching process over strategy choice, implementation, debugging, and revision, and rely on execution signals or learned critics to decide which nodes to expand. Therefore, long-horizon coding quality improves when the agent can backtrack from suboptimal decisions and compare partial trajectories. Another line of these works, such as ReLoc~\cite{lyu2025reloc} and SFS~\cite{light2025sfs}, treats planning as search in code space. Here the methods iteratively explore neighboring programs through mutation, revision, or local optimization, guided by validation feedback or fine-grained scoring signals.
Beyond the above methods, recent systems increasingly treat candidate plans, patches, logs, tests, and execution traces as persistent artifacts rather than transient generations. SWE-Search~\cite{sweSearch2024} combines Monte Carlo Tree Search with software-engineering agents to explore alternative repair trajectories, while CodeTree~\cite{li2025codetree} organizes strategy exploration, solution generation, and refinement within a unified tree. More broadly, Meta-Harness~\cite{lee2026metaharness} pushes this idea to the harness level itself: it searches over harness code by giving an agent access to prior source code, scores, and execution traces through a filesystem. These developments suggest that search-based planning is not only a model-side sampling strategy, but also a harness-level state management problem: the runtime must preserve candidates, expose evidence, run validators, and decide which branch deserves further computation.

\begin{table}[t]
\centering
\caption{Representative planning modules for code agents.}
\label{tab:planning_modules}
\renewcommand{\arraystretch}{1.12}
\setlength{\tabcolsep}{3pt}
\footnotesize
\begin{tabularx}{\textwidth}{@{}llllX@{}}
\toprule
\textbf{Method} & \textbf{Category} & \textbf{Core Mechanism} & \textbf{Interface} & \textbf{Feedback} \\
\midrule
Self-Planning~\cite{jiang2024selfplanning}
& Linear decomposition
& Stepwise decomposition
& Shared prompt
& None \\
WebAgent~\cite{gur2023webagent}
& Linear decomposition
& Sub-instruction sequencing
& APIs
& Runtime exception \\
CodePlan~\cite{bairi2024codeplan}
& Structure-grounded
& Plan graph
& Repo graph
& Critique \\
VerilogCoder~\cite{ho2025verilogcoder}
& Structure-grounded
& Task-circuit relation graph
& Repo graph
& Test pass/fail \\
Tree-of-Code~\cite{ni2024treeofcode}
& Search-based
& Trajectory tree search
& Execution env
& Test pass/fail \\
ReThinkMCTS~\cite{li2025rethinkmcts}
& Search-based
& MCTS over reasoning paths
& Execution env
& Critique, tests \\
MapCoder~\cite{islam2024mapcoder}
& Orchestration-based
& Role orchestration
& APIs
& Critique, tests \\
Blueprint2Code~\cite{mao2025blueprint2code}
& Orchestration-based
& Blueprint-to-code
& Repo interface
& Critique \\
\bottomrule
\end{tabularx}
\end{table}

\subsubsection{Orchestration-based Planning}

Orchestration-Based Planning refers to a planning paradigm in which the core planning function is realized through a harness design for system-level coordination. In this paradigm, the harness governs how agents or modules specialize roles, execute stages, route feedback, and trigger verification loops, thereby determining what actions should be taken next in long-horizon code generation workflows.
A first common pattern~\cite{huang2023agentcoder,ukai2024adacoder,Nunez2024AutoSafeCoder} is feedback-centered orchestration, where the system distributes coding, testing, analysis, and repair across different modules, so that progress is driven by repeated execution-grounded feedback and adaptive escalation. In this group, planning is not an up-front artifact, but an emergent property of how failures are detected, interpreted, and routed back into subsequent actions. A second pattern~\cite{islam2024mapcoder,Pan2025CodeCoR,mao2025blueprint2code} is staged workflow orchestration, which casts code generation as a structured software-process pipeline, such as comprehension, retrieval or preview, planning or blueprinting, coding, debugging, and repair. The main advantage of this group lies in decomposing complex generation into interpretable stages with explicit handoff rules, and the actual planning power comes from cross-stage control, candidate pruning, and iterative refinement. A third pattern~\cite{khan2025macog,doualgoforge,zhang2026sgagent,lu2025requirements} is controller-centric orchestration, where planning is embedded in the transformation of intermediate artifacts and in the routing substrate itself. Here, systems organize decision-making through mechanisms such as formal-specification pipelines, suggestion stages between localization and repair, typed intermediate representations, shared blackboards, or specialized planner–coder coordination, so that the next plan is determined by the scaffold’s control logic rather than by a single textual prompt.

Recent harness systems make this orchestration view especially explicit. Anthropic's long-running harnesses separate planning, generation, and evaluation into distinct roles, using structured artifacts and independent evaluation to maintain progress across long sessions~\cite{anthropic2025longrunning,anthropic2026longrunningapps}. Cursor's large-scale autonomous coding experiments similarly highlight planner--worker coordination as a way to scale from focused single-agent tasks to many parallel agents working on a shared project~\cite{cursor2026scalingagents}. The most general formulation appears in Natural-Language Agent Harnesses, where high-level harness logic (such as roles, stages, contracts, adapters, state conventions, and failure taxonomies) is written as editable natural language and executed by an Intelligent Harness Runtime~\cite{pan2026nlah}. The IHR interprets these high-level natural-language instructions at runtime and converts them into constrained execution steps under explicit contracts, budgets, tool interfaces, and environment state. This reframes orchestration-based planning as a runtime interpretation problem: the plan is not merely a document, but an executable harness specification that mediates between model outputs, filesystem state, tools, validators, and multi-agent delegation.

\textbf{\textit{Discussion:}} 
Planning for code generation can be understood as a core form of \emph{agentic harness}: a control layer that organizes how an LLM agent decomposes tasks, grounds decisions in program structure, explores alternatives at inference time, and coordinates multi-stage software engineering workflows. From this perspective, planning is a set of harness mechanisms centered on one essential question: how to decide what the agent should do next, and how to keep that decision process constrained, inspectable, and coherent across long-horizon coding tasks.
Notably, planning in code generation cannot be cleanly separated from the evaluation problem. Many current conclusions about the benefits of planning depend heavily on the surrounding execution conditions, including execution environments, feedback quality, tool access, trajectory budgets, and whether the benchmark truly stresses long-range dependency management rather than localized patch generation. If execution signals are weak, revision budgets are unrealistic, or benchmarks fail to expose multi-step coordination errors, then reported planning gains may not reflect genuine improvements in agent-level problem solving. Therefore, planning is not only a method design problem, but also a harness problem between the agent and the environment.
Looking forward, the central challenge is not merely to build larger planners or longer reasoning traces, but to design more reliable agentic harnesses for planning: adaptive commitment mechanisms that decide when to follow, revise, or abandon a plan; structurally meaningful planning states that expose dependencies and progress; efficient exploration-and-revision strategies that use feedback without excessive computation; and rigorous long-horizon evaluation paradigms that can faithfully measure planning quality beyond final-pass accuracy.

\subsection{Memory and Context Engineering for Agent Harness}\label{sec:memory}

\begin{figure}[t]
    \centering
    \includegraphics[width=0.85\linewidth]{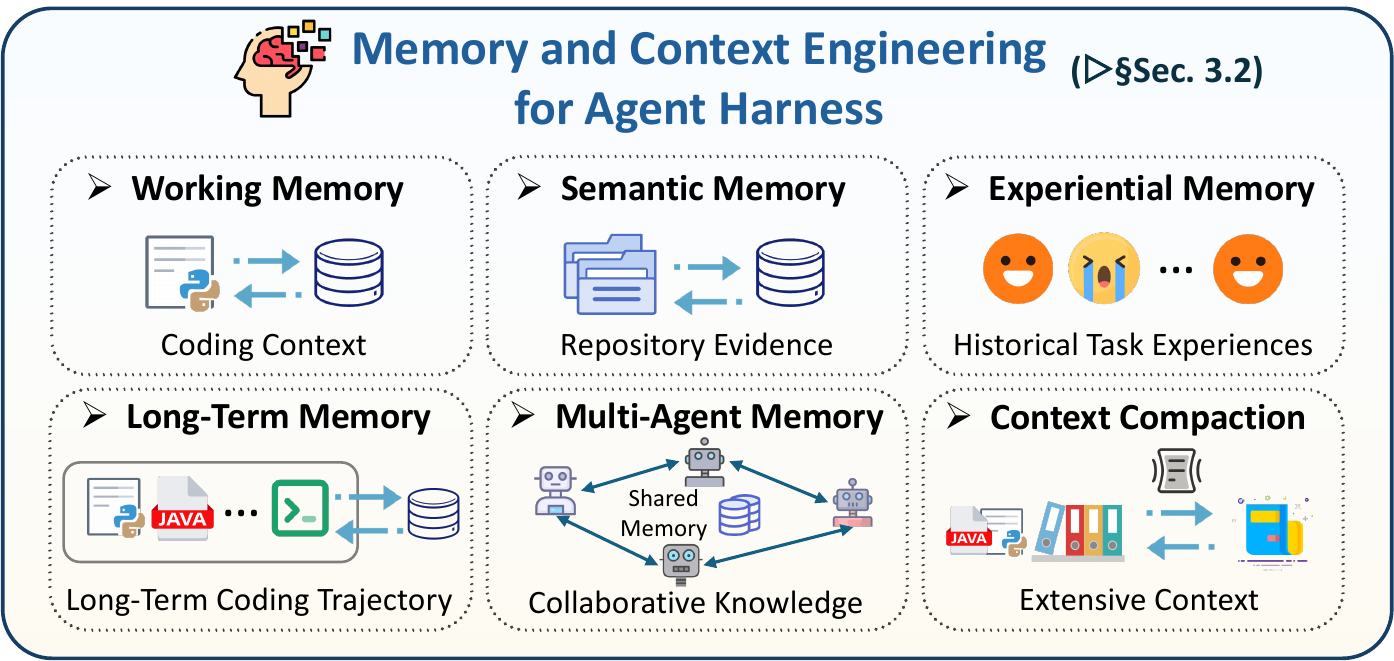}
    \caption{Overview of memory and context engineering mechanisms for agent harnesses.}
    \label{fig:modules-memory}
\end{figure}

Memory has become a core infrastructure for code agents, largely because real-world software engineering tasks are inherently long-horizon and state-intensive~\cite{dong2025survey,huang2026rethinking}. Unlike single-turn code completion, practical coding scenarios require an agent to sustain a sequence of interdependent steps across many rounds of interaction, such as requirement understanding, code localization, evidence retrieval, multi-file editing, test execution, bug fixing, and regression verification~\cite{xia2025demystifying,zhang2025survey}. This introduces a fundamental tension \textbf{between the limited context window of the model and the continuously expanding intermediate state of the task}. 
From a harness perspective, memory is not simply a larger context window or a vector database. It is a state-management layer that decides which information should remain in the active model context, which information should be compacted into summaries, and which information should be offloaded to durable external storage~\cite{zhou2026externalization}.
Without an effective memory mechanism and context management, an agent can easily lose critical clues during long-range reasoning, repeat searches and analyses that were already completed, or break local consistency established in earlier steps during later modifications~\cite{zhang2025ragsurvey,huang2026rethinking}.

Early systems largely relied on prompts to preserve historical information, treating memory as little more than conversation history or an unstructured scratchpad. However, with the emergence of repository-level repair and other long-horizon coding tasks, it has become increasingly clear that simply accumulating natural language history cannot reliably support complex software engineering loops~\cite{jiang2026survey}. As a result, memory is now increasingly externalized as a system component that is retrievable, governable, and traceable. In this subsection, we categorize memory in code agents according to their \textbf{primary functional role} in the software engineering loop. Under this view, existing approaches can be broadly organized into five types: \textit{working memory, semantic memory, experiential memory, long-term memory, and multi-agent memory}. In addition, we discuss context compaction and state offloading as cross-cutting context-engineering mechanisms that determine how large execution artifacts move between the active model context and durable task state. Representative works are illustrated in Table~\ref{tab:memory_modules}.

\subsubsection{Working Memory}
Working memory supports state maintenance along the current coding-task trajectory~\cite{huang2025language}. Its central concern is not how much history to retain, but which pieces of information are most useful for the next action under a limited context budget. In code agents, working memory often appears as structured prompt regions, state summaries, failed-test records, file lists, or critical stack information. Its purpose is to mitigate context explosion, reduce repeated localization, and preserve the local consistency of an ongoing repair or editing trajectory~\cite{yang2024swe,xia2025live,bouzenia2025repairagent,gaurav2025codemem}. 
From a harness perspective, working memory is the active control surface between the model and the code environment: it determines what the agent observes before choosing the next tool call, edit, or verification step. Representative systems such as SWE-agent~\cite{yang2024swe} and RepairAgent~\cite{bouzenia2025repairagent} show that, even with the same underlying model, repository-level repair performance can vary substantially depending on how interaction state and execution feedback are organized. CodeMem~\cite{gaurav2025codemem} similarly treats context as a managed resource, using budgeted memory slots to stabilize multi-step edits.

\subsubsection{Semantic Memory}
Semantic memory provides task-relevant external evidence for the current coding process~\cite{wu2025human,huang2026rethinking}. In code-agent settings, such evidence is usually repository-specific and program-structured, including class definitions, function implementations, call relations, configuration files, documentation, issue descriptions, dependency metadata, and historical implementation patterns. Semantic memory therefore transforms the external codebase into a queryable evidence space that the harness can retrieve from and inject into the active context~\cite{zhang2024autocoderover,zhang2024codeagent,biswal2026agentsm,zhang2025coderag,phan2025repohyper}. 
Representative works such as AutoCodeRover~\cite{zhang2024autocoderover} and RepoCoder~\cite{zhang2023repocoder} show that repository-level coding tasks benefit not simply from retrieving more content, but from retrieving evidence aligned with program structure. Mechanisms such as AST-based structured chunking, iterative query rewriting, and retrieval strategies conditioned on current localization clues can substantially improve the utility of retrieved context for downstream generation. In this sense, semantic memory turns the codebase into a structured evidence layer for the current decision process.

\begin{table}[t!]
\centering
\caption{Representative memory and context management mechanisms for code-agent harnesses.}
\renewcommand{\arraystretch}{1.00}
\setlength{\tabcolsep}{3.5pt}
\scriptsize
\begin{tabularx}{\textwidth}{p{2.35cm}p{2.5cm}p{3.2cm}p{2.8cm}X}
\toprule
\textbf{Method} & \textbf{Role} & \textbf{Managed State} & \textbf{Harness Operation} & \textbf{Primary Use} \\
\midrule
SWE-agent~\cite{yang2024swe} 
& Working Memory 
& Repair trajectory; runtime state
& Structured state tracking 
& Grounds repo repair in files, commands, and tests \\

CodeMem~\cite{gaurav2025codemem} 
& Working Memory 
& Context slots; edit state
& Budgeted slot management
& Stabilizes multi-step edits under context limits \\

RepairAgent~\cite{bouzenia2025repairagent}
& Working Memory 
& Bug evidence; tool outputs
& Dynamic prompt-state updates
& Carries evidence across autonomous cycles \\
\midrule

AutoCodeRover~\cite{zhang2024autocoderover} 
& Semantic Memory 
& Repo structure; code evidence
& Structure-aware retrieval 
& Grounds localization and patching in repo structure \\

RepoCoder~\cite{zhang2023repocoder} 
& Semantic Memory 
& Retrieved repo context; snippets
& Iterative repo retrieval 
& Expands evidence for context-aware generation \\

CodeRAG~\cite{zhang2025coderag} 
& Semantic Memory 
& Repo knowledge; code paths
& Querying; multi-path retrieval; reranking
& Selects repo knowledge for long-context completion \\
\midrule

MemGovern~\cite{wang2026memgovern} 
& Experiential Memory 
& Trajectories; reflections; critiques
& Governed experience replay
& Reuses quality experience while filtering noise \\

ExpeL~\cite{zhao2024expel} 
& Experiential Memory 
& Reflection traces; learned lessons
& Reflection replay
& Reuses reflections as task-solving strategies \\
\midrule

MemCoder~\cite{deng2026your}
& Long-term Memory 
& Commits; root causes; validated fixes
& Structured memory; self-internalization
& Learns repo-specific intent-to-code mappings \\

TALM~\cite{shen2025talm}
& Long-term Memory 
& Task histories; reasoning traces; validated code
& Vector retrieval; consolidation
& Reuses past episodes for tree-structured generation \\
\midrule

MIRIX~\cite{wang2025mirix} 
& Multi-agent Memory 
& Cross-agent state; interaction history
& Cross-agent memory routing 
& Routes shared memory across specialized roles \\

ChatDev~\cite{qian2024chatdev}
& Multi-agent Memory 
& Dialogue history; software artifacts
& Phase-level context passing
& Maintains context across role-based phases \\
\midrule

LongCodeZip~\cite{shi2025longcodezip} 
& Context Compaction
& Long code context; repo snippets
& Coarse-to-fine compression
& Compresses code while preserving reasoning cues \\

SWE-Pruner~\cite{wang2026swe} 
& Context Compaction
& Interaction context; surrounding code
& Task-aware pruning
& Removes irrelevant context before agent decisions \\

SWEZZE~\cite{jia2026compressing} 
& Context Compaction
& Issue context; fix ingredients
& Lightweight learned compression
& Distills compact, fix-relevant evidence \\
\bottomrule
\end{tabularx}
\label{tab:memory_modules}
\end{table}

\subsubsection{Experiential Memory}
As code agents move from single-task completion toward continual repair and cross-project generalization, increasing attention has been paid to experiential or episodic memory~\cite{dong2025towards,huet2025episodic}. Unlike working memory, which maintains the current trajectory, or semantic memory, which retrieves repository evidence, experiential memory captures reusable experience accumulated across tasks, such as repair trajectories, failure cases, debugging records, and higher-level strategy patterns~\cite{zhao2024expel,wei2025evo,liang2026generalizable}. Its main value lies in enabling cross-task transfer. Through mechanisms such as experience cards, reflection buffers, and record-and-replay pipelines, a system can convert past successful or failed debugging processes into reusable units for future problem solving~\cite{wei2025evo,wang2026memgovern,chu2024leveraging}. 
Works such as MemGovern~\cite{wang2026memgovern} further suggest that the quality of stored experience matters more than its scale. Ungoverned historical records can introduce semantic noise, error propagation, and false retrievals, whereas curated and quality-controlled experiential memory is more likely to become a useful asset for repository-level repair.

\subsubsection{Long-Term Memory}
When coding trajectories become longer, working memory and semantic memory alone are insufficient, because the system must also cope with memory growth, compression-induced evidence distortion, and long-term drift. This makes long-term retrieval planning and memory control an increasingly important research direction~\cite{maharana2024evaluating,wang2026memex,bei2026mem,zhao2026papermind,ning2026mcsearch}. The focus therefore shifts from memory capacity to memory governance. Representative systems such as MemGPT~\cite{packer2023memgpt} and MemoryOS~\cite{kang2025memory} move the discussion from what to store toward when to write, when to compress, when to retrieve, and how to avoid contamination.
Recent code-centric studies further ground this line of work in software engineering workflows. MemCoder~\cite{deng2026your} leverages structured historical commits and human-validated solutions as persistent memory, enabling repository-specific experience accumulation over time. TALM~\cite{shen2025talm} incorporates long-term memory into multi-agent code generation, retrieving prior problem--solution traces and consolidating overlapping memories to control redundancy. These works suggest that, for code agents, long-term memory should not simply accumulate more history, but preserve validated and reusable experience in a compact and controllable form. Otherwise, memory may shift from a resource for long-horizon software engineering into a burden that amplifies noise, staleness, and error.

\subsubsection{Multi-Agent Memory}
Multi-agent memory extends state management from an individual agent to a shared harness. From a systems perspective, memory in code generation has a strong collaborative dimension~\cite{li2025swe,chen2023gamegpt}. In multi-agent frameworks, memory is not only a container for individual state, but also a medium for information sharing, intention passing, and consistency maintenance across specialized roles~\cite{zhang2025gmemory}. Representative works such as AgentCoder~\cite{huang2023agentcoder}, MapCoder~\cite{islam2024mapcoder}, MIRIX~\cite{wang2025mirix}, ChatDev~\cite{qian2024chatdev}, and G-Memory~\cite{zhang2025gmemory} illustrate how memory supports multi-agent planning, testing, reviewing, and trajectory coordination.
In this setting, the central challenge is no longer only retrieving relevant content, but controlling the granularity of sharing, preventing information flooding, and supporting bidirectional access between high-level decisions and fine-grained execution traces~\cite{chen2023gamegpt}. Accordingly, memory in multi-agent code generation increasingly resembles a shared blackboard or collaborative state graph rather than a purely individual storage unit~\cite{Ishibashi2024SelfOrganized,yuan2025graphs}.

\subsubsection{Context Compaction and State Offloading}
Context compaction and state offloading are cross-cutting context-engineering mechanisms for memory in code-agent harnesses~\cite{liu2026dive}. Their goal is not to define another memory category, but to control the boundary between active model context and durable task state. Long-horizon software engineering workflows continuously generate high-volume artifacts, such as build logs, execution traces, repository diffs, test outputs, and intermediate plans. Directly placing these artifacts into the prompt can quickly overload the context window, amplify noise, and obscure decision-relevant evidence. A harness must therefore decide which observations should remain in the active context, which should be compacted into concise summaries, and which should be offloaded to external storage with retrievable handles~\cite{zhou2026externalization}.
Context compaction compresses long interaction histories and massive tool outputs into structured, provenance-preserving summaries. For example, a failing-test report can be reduced to the failing test name, key stack frames, suspected files, and links to the full log~\cite{jia2026compressing,sun2025scaling,shi2025longcodezip,wang2026swe}. State offloading complements this process by preserving full-fidelity artifacts outside the active window, such as in files, databases, trace stores, or protocol-style resource interfaces such as MCP-style servers. The agent then receives compact summaries and resource identifiers rather than raw logs or traces. By separating decision-relevant context from durable evidence, context compaction and state offloading make memory more scalable, auditable, and compatible with execution-time verification.

\textbf{\textit{Discussion:}} 
Memory in code-as-agent-harness systems can be understood as a unified state-management layer that connects context management, repository evidence retrieval, experiential transfer, long-term control, and multi-agent synchronization. Rather than being a single data structure, an enlarged context window, or simply a vector database, memory coordinates where task-relevant state should reside and how it should be reused throughout long-horizon software engineering workflows. Working memory keeps the next action grounded; semantic memory exposes repository evidence; experiential memory supports cross-task transfer; long-term memory preserves validated knowledge; and multi-agent memory synchronizes shared state across roles. Context compaction and state offloading further extend this layer by separating decision-relevant active context from durable full-fidelity artifacts, making memory more scalable, auditable, and compatible with execution-time verification. Importantly, memory research in code agents cannot be separated from \textit{evaluation reliability}. Many conclusions about memory gains depend on the quality of evaluation pipelines~\cite{jimenez2024swebench,feng2026longcli}: if tests are insufficient, log parsing is flawed, or benchmarks suffer from memorization and contamination, then reported improvements may not reflect robust long-horizon behavior. Looking forward, the key challenge is not merely to enlarge memory capacity, but to build higher-quality write gates, structurally aligned retrieval keys, provenance-preserving compaction mechanisms, reliable state offloading protocols, and rigorous evaluation paradigms that measure whether memory truly helps agents remain grounded, consistent, and verifiable over extended trajectories.

\subsection{Tool Use for Agent Harness}\label{sec:tool}

\begin{figure}[t]
    \centering
    \includegraphics[width=0.8\linewidth]{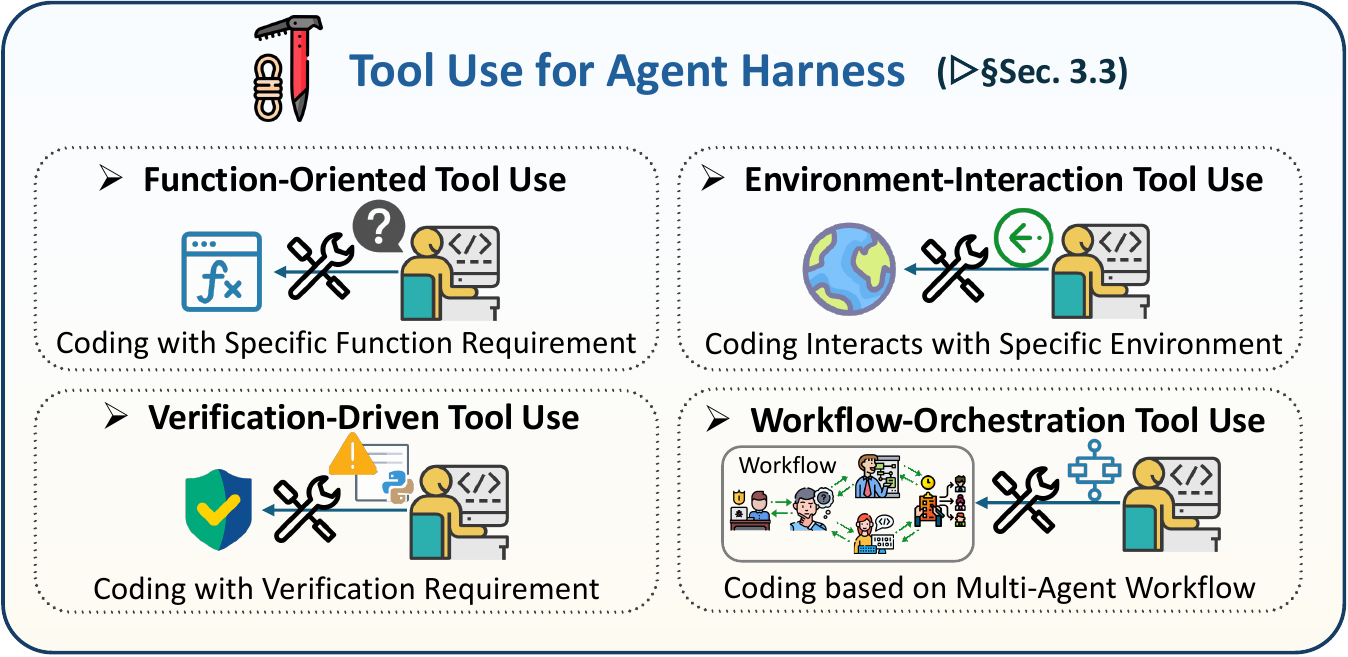}
    \caption{Overview of tool-using mechanisms for agent harnesses.}
    \label{fig:modules-tool}
\end{figure}

Tool usage is the action and observation layer of the code-agent harness. Once code is placed inside the agent loop, the model must not only generate text, but also search repositories, edit code, execute tests, call APIs, query documentation, and verify intermediate results~\cite{watanabe2025use,sapkota2025vibe}. Tools therefore expand the agent's action space while also exposing external feedback signals that make the harness executable and inspectable. From the perspective of code as agent harness, tool use is not merely an auxiliary capability for code generation. It is a governed interface between model intent and external systems. A reliable harness must decide which tools are available, how their schemas are exposed, what permissions each tool receives, where execution happens, how results are sanitized or compacted, and when risky actions require human approval. Recent agent SDKs and software-agent platforms make this shift explicit by packaging tools, sessions, guardrails, handoffs, workspaces, and execution environments into reusable harness components~\cite{wang2024openhands,meng2026agent,xi2025agentgym}. In parallel, sandboxed execution environments, including containerized or microVM-based workspaces, isolate agent actions from the host system and make code execution more reproducible and auditable~\cite{cheng2026llm,wang2024executable,wang2025ui}.
This harness-level view also highlights the importance of \textbf{tool lifecycle control}. Before a tool is executed, the harness may apply permission checks, policy rules, argument validation, or human-in-the-loop gates. After execution, the harness may sanitize outputs, summarize large logs, offload traces to durable storage, update memory, or trigger verification tools. Lifecycle hooks make these control points explicit. They turn tool use from a raw model-selected action into a monitored transition in the agent's execution loop.

Existing work on tool usage for code agents can therefore be organized according to the primary harness function that tools serve: (1) \textit{function-oriented tool use}, (2) \textit{environment-interaction tool use}, (3) \textit{verification-driven tool use}, and (4) \textit{workflow-orchestration tool use}. 
Function-oriented tools ground the agent in APIs, libraries, and external documentation. Environment-interaction tools allow the agent to act inside repositories, terminals, IDEs, browsers, and sandboxes. Verification-driven tools provide deterministic feedback through tests, linters, type checkers, static analyzers, and runtime errors. Workflow-orchestration tools coordinate multiple tools, roles, memory updates, and lifecycle policies into a reliable long-horizon execution process.
Representative works are illustrated in Table~\ref{tab:tool_modules}.

\begin{table}[t]
\centering
\caption{Representative tool-use mechanisms for code-agent harnesses.}
\label{tab:tool_modules}
\renewcommand{\arraystretch}{1.12}
\setlength{\tabcolsep}{3pt}
\scriptsize
\begin{tabularx}{\textwidth}{@{}
  >{\raggedright\arraybackslash}p{2.6cm}
  >{\raggedright\arraybackslash}p{3.1cm}
  >{\raggedright\arraybackslash}p{3.0cm}
  >{\raggedright\arraybackslash}p{3.8cm}
  >{\raggedright\arraybackslash}X@{}}
\toprule
\textbf{Method} & \textbf{Role} & \textbf{Tool Boundary} & \textbf{Harness Operation} & \textbf{Primary Use} \\
\midrule
ToolCoder~\cite{zhang2023toolcoder}
& Function-oriented
& API search tools
& API selection via trigger prediction
& Grounds generation in retrieved APIs \\
CodeQA~\cite{ahmed2024codeqa}
& Function-oriented
& API/doc query tools
& Tool-augmented API QA
& Retrieves API evidence for coding \\
RAG-for-Code~\cite{zhao2025rag}
& Function-oriented
& Repo, docs, API
& Retrieval-augmented context
& Knowledge for long-tail libraries \\
\midrule
CodeAgent~\cite{zhang2024codeagent}
& Environment-interaction
& Repo files, tests
& Repo navigation, editing, validation
& Repo-level coding via environment interaction \\
SWE-agent~\cite{yang2024swe}
& Environment-interaction
& Shell, editor, repo, tests
& Agent--computer interface loop
& Resolves GitHub issues via shell commands \\
\midrule
AgentCoder~\cite{huang2023agentcoder}
& Verification-driven
& Test generation
& Programmer--tester--executor loop
& Refines code via generated tests \\
VeriGuard~\cite{miculicich2025veriguard}
& Verification-driven
& Execution, tests, verifier
& Verifier-guided tool loop
& Gates and repairs code via verification \\
\midrule
ToolNet~\cite{liu2024toolnet}
& Workflow-orchestration
& APIs, tools, execution
& Learned multi-tool policy routing
& Routes tool invocations across workflows \\
MapCoder~\cite{islam2024mapcoder}
& Workflow-orchestration
& Coding agents
& Multi-agent tool-supported workflow
& Coordinates planning, generation, debugging \\
OpenHands~\cite{wang2024openhands}
& Workflow-orchestration
& Workspace, terminal, browser, files, runtime
& Unified software-agent workspace
& Long-horizon tasks via reusable interfaces \\
\bottomrule
\end{tabularx}
\end{table}

\subsubsection{Function-Oriented Tool Use}
This line of work uses tools primarily to fill gaps in the model's programming knowledge, especially APIs, libraries, documentation, and external coding utilities~\cite{zhang2023toolcoder,ahmed2024codeqa,zhao2025rag,li2025survey,yuan2025easytool, zou2025autotool}. ToolCoder~\cite{zhang2023toolcoder}, for example, starts from a clear bottleneck: code models often hallucinate APIs, choose inappropriate functions, or fail on public and private libraries with sparse training coverage. To address this problem, it integrates API search tools into the code generation process and trains models to decide when to query the tool and how to select APIs from retrieved results. The key contribution is therefore not better syntax generation alone, but better knowledge acquisition and API grounding. More broadly, retrieval-oriented methods reduce dependence on parametric memory and make code generation more adaptable to long-tail APIs, private libraries, and continuously evolving software ecosystems~\cite{zhao2025rag,zhou2023devil}. They are most effective when the main bottleneck is that the model lacks reliable knowledge of which function, API, or library construct should be used. Accordingly, the core design challenges lie in query formulation, result selection, evidence compression, and robust injection of retrieved knowledge into downstream generation. These agentic methods are particularly suitable for API-oriented generation, library migration, and private SDK usage, but retrieval alone is often insufficient when tasks require cross-file understanding and reasoning, runtime debugging, or repository-wide dependency analysis.

\subsubsection{Environment-Interaction Tool Use}
Unlike function-oriented tools, environment-interaction approaches treat tools as the interface through which an agent acts inside the software engineering environment~\cite{li2026environment,chen2026grounded,song2026envscaler,gao2026teaching}. Their central problem is no longer only to obtain missing functions, but to operate effectively over repositories, development artifacts, and execution environments. CodeAgent~\cite{zhang2024codeagent} shows that real-world repository-level code generation is not simply about completing a single function from a prompt. Instead, the model must locate relevant files, understand dependencies, inspect documentation, implement modifications, and validate outcomes through testing. To support this process, CodeAgent integrates programming tools and agent strategies for information retrieval, code-symbol navigation, code implementation, and test interaction over real repositories. SWE-agent~\cite{yang2024swe} pushes this idea further by formalizing the agent-computer interface, where shell commands, file editing, and test execution become the primary interaction channel. RepairAgent~\cite{bouzenia2025repairagent} similarly equips the agent with repair-specific tools for reading code, searching repair ingredients, applying patches, and running tests. Together, these methods define the core trajectory of environment-interaction tool use, which is especially relevant for repository-level generation, issue resolution, and open-ended software engineering tasks.

\subsubsection{Verification-Driven Tool Use}
A third line of work uses tools primarily for post-generation verification and iterative improvement. Verification-driven tool use treats external tools as deterministic sensors for the harness. Compared with function-oriented and environment-interaction tools, these approaches do not necessarily emphasize external retrieval or broad repository navigation. Instead, they use tests, execution results, compiler errors, runtime traces, type checkers, static analyzers, and verifier feedback as the main signals for improving code quality~\cite{miculicich2025veriguard,liu2026agents4plc,liu2026llm,jin2025reveal}. AgentCoder~\cite{huang2023agentcoder}, for example, uses a programmer agent, a test designer agent, and a test executor agent to form a closed loop of code generation, test construction, execution, and refinement. In this paradigm, the central role of tools is verification rather than retrieval. From the code-as-agent-harness view, verification tools make agent progress inspectable: test failures, stack traces, coverage gaps, type errors, and static-analysis warnings become structured observations that update working memory and guide the next action. The key design issue is how to route these observations back into the loop~\cite{miculicich2025veriguard}. Since raw logs may be too long or noisy for the active context, the harness should parse, summarize, and offload verification traces while preserving full-fidelity artifacts for audit and replay.

\subsubsection{Workflow-Orchestration Tool Use}
Workflow-orchestration tool use focuses on how multiple tools, roles, and control policies are organized into a coherent agent workflow~\cite{xiong2025self,shi2025flowxpert,lumer2025tool, su2025toolorchestra}. In long-horizon software tasks, the agent may need to retrieve evidence, localize bugs, modify files, run tests, inspect failures, update memory, ask for approval, and repeat this cycle several times. The challenge is not simply adding more tools, but deciding when each tool should be invoked, with what permissions, under which context, and how its result should update the harness state~\cite{liu2024toolnet}. Recent agent SDKs and software-agent platforms make this orchestration layer explicit by packaging typed tool schemas, session state, workspaces, guardrails, handoffs, tracing, and human-review mechanisms into reusable harness components. Lifecycle hooks further refine this boundary: pre-use hooks can validate arguments, enforce permission policies, or block risky commands, while post-use hooks can sanitize outputs, compact logs, update memory, or trigger follow-up verification. Representative systems such as MapCoder~\cite{islam2024mapcoder} exemplify workflow orchestration by assigning agents to example recall, planning, code generation, and debugging, thereby decomposing a difficult coding problem into coordinated subproblems. CodeAgent~\cite{zhang2024codeagent} also studies how tool calls should be scheduled and structured in repository-level workflows. This class is particularly important for long-horizon code agents, where realistic software tasks require demand decomposition, context selection, candidate exploration, execution-based verification, and final repair under explicit control policies~\cite{liu2024toolnet,liu2024controlllm}.

\textbf{\textit{Discussion}}: Tool usage in code agents has evolved from isolated API retrieval to a full harness mechanism for action, observation, verification, and governance. Function-oriented tools ground implementation choices in external knowledge; environment-interaction tools allow agents to act over repositories and execution environments; verification-driven tools provide deterministic feedback; and workflow-orchestration tools coordinate these capabilities through SDKs, sandboxes, guardrails, and lifecycle hooks. The core challenge is no longer whether a model can call a tool, but whether the harness can make tool use safe, auditable, and useful for long-horizon execution. Future code-agent harnesses should support typed tool schemas, permission-aware invocation, sandboxed execution, lifecycle hooks, result sanitization, context compaction, state offloading, and reproducible traces. These mechanisms ensure that tools expand the agent's action space without sacrificing reliability, safety, or verifiability.

\subsection{Harness Control through the Plan, Execute, and Verify Loop}
\label{sec:debug}

\begin{figure}[t]
    \centering
    \includegraphics[width=0.8\linewidth]{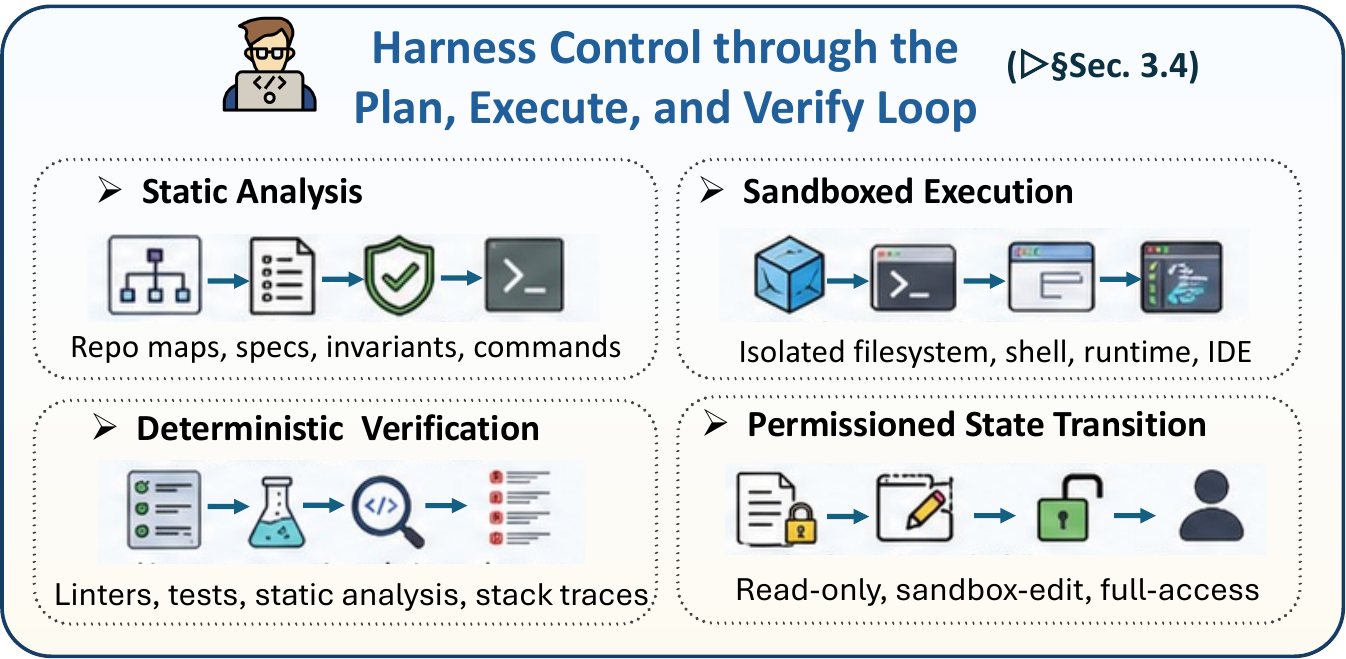}
    \caption{Overview of harness control through PEV loop.}
    \label{fig:modules-verification}
\end{figure}

Code-as-harness systems require a control loop that turns model intentions into bounded, observable, and revisable state transitions. This subsection frames that loop as \emph{Plan--Execute--Verify} (PEV): the harness first externalizes an intended change and its validation criteria, then executes the change inside a sandboxed and permissioned environment, and finally verifies the resulting state through deterministic sensors and human-review gates. This framing unifies planning, execution, debugging, verification, and escalation as parts of a single harness-level control process.
\subsubsection{From Debugging to Harness-Level Control}
The preceding subsections describe planning as trajectory control, memory as state management, and tool use as a governed action interface. Feedback-guided debugging connects these mechanisms into a closed loop: plans specify intended changes, memory preserves relevant evidence, tools execute and inspect actions, and validation signals determine whether the agent should continue, revise, or stop. As code-centric agents move from single-turn generation to repository-level software work, debugging is therefore better understood as control over executable program state rather than as a post hoc correction stage. Generated programs can fail through syntax errors, runtime exceptions, incorrect outputs, incomplete edge-case handling, unsafe operations, or violations of project-specific conventions, making one-pass generation insufficient~\cite{chen2023teaching}. Recent systems revise code through feedback from compilers, runtimes, tests, static analyzers, humans, and auxiliary agents~\cite{shinn2023reflexion, zhong2024debug, bi2024iterative, dai2025feedbackeval}. From the harness perspective, this process can be reframed as a \emph{Plan--Execute--Verify} (PEV) loop: the agent externalizes an intended trajectory, executes bounded actions inside a controlled environment, and verifies the resulting state before the next transition. The growing engineering ecosystem around agent harnesses reinforces this view: recent curated resources distinguish orchestration, working state, execution substrates, evaluation harnesses, observability, and governance as separable harness layers rather than incidental implementation details~\cite{picrew2026awesomeagentharness,openaiharnessengineering2026,opencodexloop2026,langchainanatomyharness2026}.

In this view, the harness acts as a \emph{cybernetic governor}: a control layer that observes the effects of agent actions and regulates subsequent state transitions. Rather than merely forwarding error messages to the model, it observes the repository and execution environment through deterministic sensors such as linters, parsers, compilers, type checkers, unit tests, integration tests, static analyzers, fuzzers, runtime monitors, and CI pipelines. These sensors turn a coding trajectory into inspectable signals, including pass/fail outcomes, diagnostics, failing traces, coverage gaps, security warnings, resource limits, and policy violations. The harness can then decide whether to continue execution, revise a patch, request more context, route the task to another module, reduce permissions, or escalate to a human reviewer. Table~\ref{tab:pev_modules} summarizes this control surface; the remainder of this subsection follows the loop from contract formation, through sandboxed state transition, to deterministic verification and evidence-grounded repair.

\begin{table}[t]
\centering
\caption{Representative methods and systems for PEV-loop harness control.}
\label{tab:pev_modules}
\renewcommand{\arraystretch}{1.12}
\setlength{\tabcolsep}{3pt}
\footnotesize
\resizebox{\textwidth}{!}{%
\begin{tabular}{@{}llll@{}}
\toprule
\textbf{Method} & \textbf{PEV Role} & \textbf{Core Mechanism} & \textbf{Signals and Gates} \\
\midrule
CodePlan~\cite{bairi2024codeplan} & Plan, structural & Dependency plan graph & Repo links, critiques \\
MapCoder~\cite{islam2024mapcoder} & Plan, orchestration & Map-code-test stages & Handoffs, tests, failures \\
Open\-Hands~\cite{wang2025openhands} & Full PEV harness & Stateful edit-exec workspace & Diffs, logs, tests, approvals \\
SWE-agent~\cite{yang2024swe} & Execute, CLI & Replayable shell interface & Commands, patches, tests \\
Daytona~\cite{daytona2026} & Execute, cloud sandbox & Isolated dev workspace & Files, limits, snapshots \\
E2B~\cite{e2b2026} & Execute, code-browser sandbox & Cloud code-browser sandbox & Stdout, limits, UI state \\
Self-Debugging~\cite{chen2023teaching} & Verify, self-debug & Explanation-guided repair & Errors, tests \\
Reflexion~\cite{shinn2023reflexion} & Verify, reflection memory & Verbal feedback memory & Outcomes, critiques \\
Debug Like a Human~\cite{zhong2024debug} & Verify, stepwise debug & Runtime-step checks & Traces, variables, asserts \\
Iterative Refinement~\cite{bi2024iterative} & Plan--Verify feedback & Project-context repair & Compiler diagnostics \\
Quality\-Flow~\cite{Hu2025QualityFlow} & Verify, quality gate & Quality feedback routing & Tests, success, stopping \\
AgentCoder~\cite{huang2023agentcoder} & Verify, multi-agent repair & Coder-tester-executor loop & Tests, failures, critique \\
Auto\-SafeCoder~\cite{Nunez2024AutoSafeCoder} & Verify, safety sensors & Static checks, fuzzing & Alerts, traces, tests \\
VeriGuard~\cite{miculicich2025veriguard} & Verify, verified gen. & Verifier guard layer & Proofs, tests, alerts \\
LiteLLM~\cite{litellm2026} & Permission gateway & Proxy policy routing & Approvals, denials, cost logs \\
\bottomrule
\end{tabular}%
}
\end{table}

\subsubsection{Planning as Contract Formation}
The planning phase turns a user request into an explicit contract over the next state transition. A robust plan does more than decompose the request into implementation steps; it also identifies relevant files, expected invariants, validation commands, rollback points, and risky operations. This makes planning a harness artifact rather than an unobserved reasoning trace. In repository-level tasks, such artifacts constrain the subsequent action space by specifying which components may be read, which files may be edited, and which verification criteria must be satisfied before completion~\cite{jiang2024selfplanning, bairi2024codeplan, islam2024mapcoder}. Repository-local instructions and tool protocols strengthen this contract layer: AGENTS.md-style guidance, MCP server registries, typed tool schemas, adapters, and protocol gateways make the available actions inspectable before execution rather than discovered opportunistically during execution~\cite{agentsmd2026,mcpservers2026,modelcontextprotocol2026,langchainmcpadapters2026,RayASO,hou2025model,li2025glue,contextforge2026}. The PEV framing also clarifies why planning and debugging should not be separated: failed verification updates the plan, while the plan determines which verification evidence is meaningful.

\subsubsection{Sandboxed Execution and Permissioned State Transition}
The execution phase realizes the plan as a bounded and observable state transition. The sandboxed environment is the operational substrate of the loop: it provides an isolated filesystem, dependency state, shell, language runtime, browser or IDE interface, and resource boundary in which agent-generated actions can be run without directly compromising the host system~\cite{vijayvargiya2025openagentsafety, cheng2026llm}. Contemporary execution-substrate work is best read as functional clusters rather than as an undifferentiated catalog. Coding sandboxes expose filesystems, Git operations, shells, package managers, and code-execution backends~\cite{daytona2026,e2b2026,alibabaopensandbox2026,judge02026,swerex2026,wang2025openhands}; computer-use substrates add browser, desktop, LSP, or IDE state~\cite{trycua2026,browserharness2026,e2bdesktop2026,agentinfrasandbox2026,agentscoperuntime2026}; and durable runtimes emphasize microVM or WASM isolation, snapshots, warm pools, resumable sessions, benchmark environments, and always-on operating contexts~\cite{tensorlake2026,arrakis2025,capsule2026,kubernetesagentsandbox2026,sandboxedsh2026,terminalbenchenv2026,stakpakagent2026}. Sandboxes also improve reproducibility because the harness can replay the same patch, command, seed, dependency lockfile, or test configuration under comparable conditions. Without this stable substrate, verification signals become difficult to interpret, and failures may reflect environment drift rather than program defects~\cite{wang2025openhands, feng2026longcli,anthropicinfranoise2026}.

Execution must also be permissioned. A multi-tier model separates low-risk observation from high-risk action: a read-only tier supports repository browsing, retrieval, static inspection, and log analysis; a sandbox-edit tier supports local patching, test execution, and temporary dependency installation inside an isolated workspace; and a full-access tier covers network access, credentials, deployment commands, package publishing, destructive filesystem operations, or Git history mutation. Actions in the final tier should be guarded by mandatory human-in-the-loop (HITL) gates because their consequences can extend beyond the sandbox. Recent software-agent systems and harness engineering work increasingly expose these control points through explicit tools, sessions, policies, approval prompts, and audit logs~\cite{sergeyuk2026human, wang2025openhands, lin2026agentic, zhou2026externalization,anthropicclaudecodeautomode2026,anthropicsandboxing2026}. Gateway and policy layers then provide the production counterpart: systems for model routing, tool registration, proxy-level logging, centralized guardrails, security automation, and falsifiable approval evidence keep governance outside the prompt alone~\cite{litellm2026,kong2026,portkey2026,contextforge2026,agentgateway2026,openairealtimeagents2026,openaicsagentsdemo2026,tracecat2026,archestra2026,haft2026}.

\subsubsection{Verification through Deterministic Sensors}

The verification phase closes and, when necessary, reopens the loop by comparing the new state against explicit constraints. Compilation and static-analysis feedback provide low-cost sensors before full execution, including parser diagnostics, type errors, lint warnings, and security alerts~\cite{bi2024iterative, adnan2025debugging, blyth2025static}. Runtime signals expose failures that only arise along concrete execution paths, such as exceptions, assertion breaks, invalid API usage, resource exhaustion, and timeouts~\cite{sun2024llm, huang2025mldebugging, zhong2024debug}. Test-based feedback then evaluates whether the observed behavior satisfies the intended specification, using unit tests, integration tests, regression tests, fuzzing, or benchmark-specific evaluators~\cite{chen2023teaching, fakhoury2024llm, gu2024testart, shi2025from}. Evaluation harnesses broaden this idea from a single test command to repeatable task distributions: they encode evaluator logic, simulation hooks, red-team cases, or RL-style environments that can compare harness variants under controlled conditions~\cite{promptfoo2026,deepeval2026,ragas2026,lmevaluationharness2026,langwatch2026,evalscope2026,harbor2026,tau2bench2026,nemogym2026,agentevaluation2026,inspectevals2026}. Compared with natural-language critique, these sensors are deterministic or at least reproducible enough to serve as control signals. Human or agentic critiques remain useful when failure evidence is sparse, but in a governed PEV loop they should interpret sensor outputs rather than replace them~\cite{shinn2023reflexion, ross2023programmer, wu2024autogen}.

Verification also supplies the evidence for repair, reflection, and termination, so these activities are treated as consequences of the Verify phase rather than as an independent stage. When a check fails, the same sensor evidence can determine whether the harness should ask the model to diagnose the failure, retrieve missing context, regenerate a localized patch, route the task to a testing or security agent, or abandon the current branch. Self-reflection mechanisms help transform raw diagnostics into actionable hypotheses, such as whether the failure comes from incorrect control flow, missing edge cases, misunderstood APIs, or inadequate tests~\cite{Wu2025IterPrefFP, Pan2025CodeCoR}. However, reflection is reliable only when it remains grounded in executable evidence. Systems such as AgentCoder, AutoSafeCoder, and QualityFlow illustrate this principle by combining agentic critique with independent execution, static analysis, fuzzing, or test-quality gates~\cite{huang2023agentcoder, Nunez2024AutoSafeCoder, Hu2025QualityFlow}. Termination should likewise be governed by verification rather than by model confidence: a loop can stop when required checks pass, when additional attempts no longer improve the state, when the risk tier changes, or when human review is required.

\textbf{\textit{Discussion:}} Recasting iterative debugging as the PEV loop emphasizes that reliability comes from governed state transitions, not simply from better repair prompts. Planning externalizes intended changes and risk assumptions; execution applies them inside sandboxed and permissioned environments; verification uses deterministic sensors to decide whether the state is acceptable; and HITL gates preserve accountability when the action space crosses a safety boundary. This framing unifies static analysis, runtime errors, tests, critique, self-reflection, and human review as components of a cybernetic harness that regulates the agent's trajectory over executable program state.

\subsection{Agentic Harness Engineering for Adaptive Harness Optimization}
\label{sec:ahe}

Agentic Harness Engineering (AHE) names a harness-level design problem: how to measure and revise the software substrate that turns a language model into a coding agent. Whereas prompt engineering changes instructions and context engineering changes what evidence is presented to the model, AHE treats the operating environment itself as the object of analysis, including tool schemas, planning artifacts, memory policies, retrieval strategies, sandbox configuration, verification sensors, permission tiers, routing rules, multi-agent workflows, and human-review gates~\cite{lin2026agentic, zhou2026externalization}. This perspective is useful because many observed failures in code agents arise from missing repository context, brittle tool interfaces, weak validators, excessive token cost, poor retry policies, or mismatched permission boundaries rather than from model generation.

Existing work can be read as three complementary strands. AutoHarness studies automatic synthesis of code harnesses~\cite{lou2026autoharness}; Meta-Harness formulates harness design as an optimization problem over model-facing infrastructure~\cite{lee2026metaharness}; and observability-driven AHE emphasizes telemetry-rich diagnosis of where the agent loop fails and which harness component should change~\cite{lin2026agentic}. Related work on reflective prompt evolution, self-evolving workflows, and live software-engineering agents supports the same systems view: changing the scaffold around the model can change agent behavior without retraining the base model~\cite{agrawal2025gepa, Liu2025SEW, xia2025live}. Engineering guides from OpenAI, Anthropic, and LangChain converge on the same practical lesson: reliable agents require explicit harness loops, tool contracts, trace replay, evaluation suites, context budgets, and controlled execution boundaries~\cite{openaiharnessengineering2026,opencodexloop2026,anthropicmanagedagents2026,anthropicmcpexecution2026,langchaindeepagentsharness2026}.

\begin{figure}[t]
    \centering
    \includegraphics[width=0.85\linewidth]{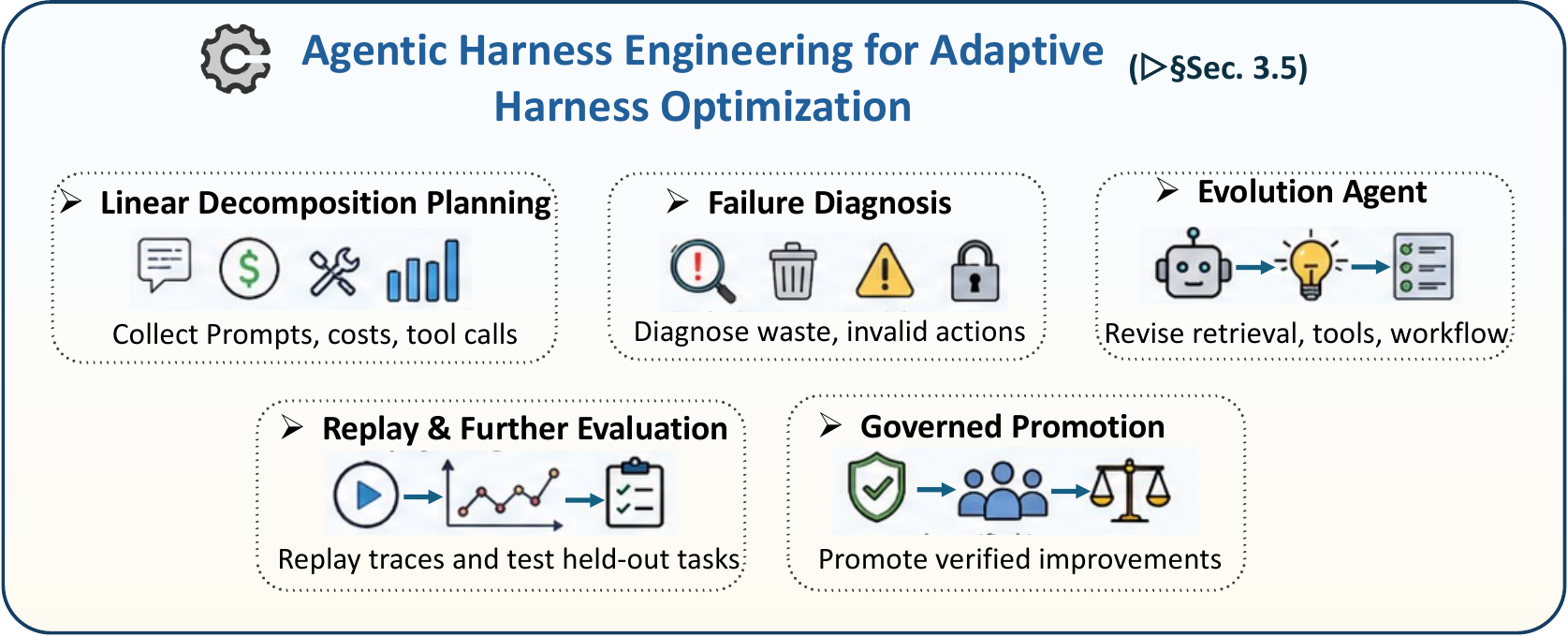}
    \caption{Overview of harness engineering for adaptive harness optimization.}
    \label{fig:modules-harness-engineering}
\end{figure}

\subsubsection{Deep Telemetry as the Optimization Substrate}
The central substrate of AHE is \emph{deep telemetry}: structured traces that connect model decisions, harness actions, environment states, and outcomes. A shallow log may record only the final answer or pass/fail result. Deep telemetry records the decision process in greater detail: prompts and retrieved context, token usage and cost, model/tool latency, tool arguments, permission requests, edited files, sandbox snapshots, command outputs, test results, stack traces, lint warnings, branch decisions, rejected alternatives, human interventions, and final task outcome. In code-centric settings, these traces are especially valuable because program execution already exposes state transitions through logs, tests, diffs, and runtime behavior~\cite{ding2024semcoder, armengol2025cannot, copet2025cwm}. In production systems, this role is increasingly served by observability and reliability stacks that record traces, metrics, prompts, model traffic, eval results, and cost signals~\cite{langfuse2026,mlflow2026,opik2026,ragaaicatalyst2026,tensorzero2026,arizephoenix2026,openllmetry2026,helicone2026,agentops2026,latitude2026,laminar2026,openinference2026,futureagi2026}. Evaluation, observability, and governance systems therefore provide complementary telemetry channels: evaluators expose task-level regressions, tracing stacks expose trajectory-level causes, and policy gateways expose boundary violations that an Evolution Agent can turn into harness revisions.

Telemetry turns harness revision from anecdotal debugging into comparative diagnosis. Token-cost traces reveal when retrieval or reflection stages consume budget without improving verification outcomes. Decision-tree traces show where the agent repeatedly chooses unproductive tools, edits irrelevant files, or loops between failed strategies. Failure traces cluster recurring patterns such as missing dependencies, weak tests, hallucinated APIs, flaky sandboxes, over-permissive tool calls, or premature termination. Because these signals are linked to concrete artifacts, they can be replayed and compared across harness versions, making it possible to evaluate whether a change improves reliability rather than merely changing surface behavior~\cite{jimenez2024swebench, feng2026longcli}.

\subsubsection{The Evolution Agent}
An \emph{Evolution Agent} is a meta-level agent that uses deep telemetry to propose, evaluate, and promote revisions to harness components. Unlike a task agent, which edits the target repository, the Evolution Agent edits the operating conditions under which later task agents work. Its input is a corpus of trajectories; its output may be a revised prompt template, a retrieval policy, a more precise tool schema, an added validator, a changed permission rule, a workflow-topology adjustment, or a new regression test. This role is closely related to self-evolving multi-agent systems in which specialized agents inspect execution logs, attribute failures to workflow components, and update collaboration structures~\cite{Hu2025EvoMAC,zou2025latentmas}. In the harness setting, the same idea is generalized from multi-agent topology to the control surface of the agent runtime.

A typical Evolution-Agent loop contains five stages. First, it \emph{observes} trajectories by collecting telemetry from PEV executions. Second, it \emph{diagnoses} failure modes by attributing cost, latency, invalid actions, test failures, or permission denials to specific harness components. Third, it \emph{proposes} candidate revisions, such as rewriting tool descriptions, changing context packing rules, adding a linter, modifying retry limits, or inserting a HITL gate before risky commands. Fourth, it \emph{evaluates} the revised harness on held-out tasks or replayed traces using deterministic sensors and regression tests. Finally, it \emph{promotes} only changes that improve reliability, cost, or safety without regressing previously solved cases. This keeps AHE within the same engineering discipline as the PEV loop: proposed changes must be executed, verified, and made auditable before adoption.

\begin{table}[t]
\centering
\caption{Representative methods for Agentic Harness Engineering with telemetry-driven revision targets.}
\label{tab:ahe_telemetry}
\renewcommand{\arraystretch}{1.12}
\setlength{\tabcolsep}{3pt}
\footnotesize
\resizebox{\textwidth}{!}{%
\begin{tabular}{@{}llll@{}}
\toprule
\textbf{Method} & \textbf{Category} & \textbf{Telemetry} & \textbf{Revision Target} \\
\midrule
AutoHarness~\cite{lou2026autoharness} & Harness synthesis & Failures, fixtures, assertions & Harness code and tests \\
Meta-Harness~\cite{lee2026metaharness} & Harness search & Code, scores, traces & Prompts, tools, scripts \\
AHE~\cite{lin2026agentic} & Telemetry-driven optimization & Cost, decisions, latency, failures & Context, tools, validators \\
GEPA~\cite{agrawal2025gepa} & Reflective prompt evolution & Scores, feedback, critiques & Prompts and instructions \\
EvoMAC~\cite{Hu2025EvoMAC} & Workflow topology evolution & Handoffs, idle roles, loops & Agent roles and graph \\
SEW~\cite{Liu2025SEW} & Self-evolving workflow & Workflow scores, failures & Stage order and roles \\
Live-SWE~\cite{xia2025live} & Online agent evolution & Live issue trajectories & Policies, tools, memory \\
GroundedTTA~\cite{chen2026grounded} & Test-time adaptation & State-action evidence & Adaptation rules \\
RLEF~\cite{gehring2024rlef} & Execution-feedback learning & Execution rewards, failures & Feedback reward signal \\
DeepEval~\cite{deepeval2026} & Evaluation harness & Scenario and metric traces & Regression suites, gates \\
FeedbackEval~\cite{dai2025feedbackeval} & Repair evaluation benchmark & Feedback-task scores & Failure taxonomy and eval set \\
Langfuse~\cite{langfuse2026} & Observability platform & Spans, cost, latency, evals & Dashboards and replay \\
OpenLLMetry~\cite{openllmetry2026} & Trace instrumentation & OpenTelemetry spans, calls & Harness instrumentation \\
Promptfoo~\cite{promptfoo2026} & Evaluation harness & Scores, regressions, failures & Eval gates and red tests \\
LiteLLM~\cite{litellm2026} & Gateway governance & Routing, budgets, failures & Budgets, fallbacks, tiers \\
\bottomrule
\end{tabular}%
}
\end{table}

\subsubsection{Governed Harness Mutation}
AHE should not be confused with unconstrained self-modification. Because the Evolution Agent changes the harness that controls later task agents, its actions require stronger governance than ordinary code repair. Candidate harness changes should be evaluated inside sandboxes, compared against fixed regression suites, and recorded with auditable rationales. Changes that alter permission boundaries, network access, credential handling, deployment behavior, or human-review requirements should require HITL approval before activation. In this sense, the Evolution Agent is itself subject to the PEV loop: it plans a harness mutation, executes it in an isolated evaluation environment, verifies the result through telemetry and regression tests, and escalates risky changes to humans.

\textbf{\textit{Discussion:}} Agentic Harness Engineering extends the code-as-harness view from operating agents to analyzing the infrastructure that operates them. Deep telemetry provides evidence for locating failures across prompts, tools, memory, sandboxes, validators, permissions, and workflows. Evolution Agents use this evidence to propose and evaluate harness mutations, turning harness design into an iterative and measurable engineering process governed by verification and human approval.

\section{Scaling the Harness: Multi-Agent Orchestration over Code}
\label{sec:mas}

As AI systems tackle increasingly complex problems from
function-level code synthesis to repository-level system
engineering, fundamental limitations for single-agent emerge: (1)
context window constraints prevent a single agent from holding an
entire codebase, long interaction history, and execution trace in
working memory; (2) specialization requirements make it
inefficient to use one generalist agent for planning, synthesis,
testing, review, and debugging simultaneously; and (3) the
absence of independent coordination and verification channels
prevents the agent from reliably detecting and correcting its own
errors during long-horizon execution.
Multi-agent systems introduce a powerful principle: once these
responsibilities are distributed across specialized roles, the
agent harness itself becomes more modular, inspectable, and
adaptable.
Early systems such as ChatDev~\cite{Qian2023ChatDev},
MetaGPT~\cite{Hong2023MetaGPT}, and AgentCoder~\cite{huang2023agentcoder}
demonstrate this shift by dividing software-development
responsibilities among distinct agents such as architect,
programmer, tester, reviewer, and executor.
Coordinated through structured communication protocols and shared
code artifacts, these role-specialized agents turn code from a
mere output target into the shared substrate through which the
overall harness plans, acts, verifies, and improves itself.

In this section, we systematically survey the rapidly growing
direction on using MAS to scale coding harnesses, and we propose
a new position on building shared code-centric harness substrates
for AI agents.

\begin{figure}[t]
    \centering
    \includegraphics[width=1.0\linewidth]{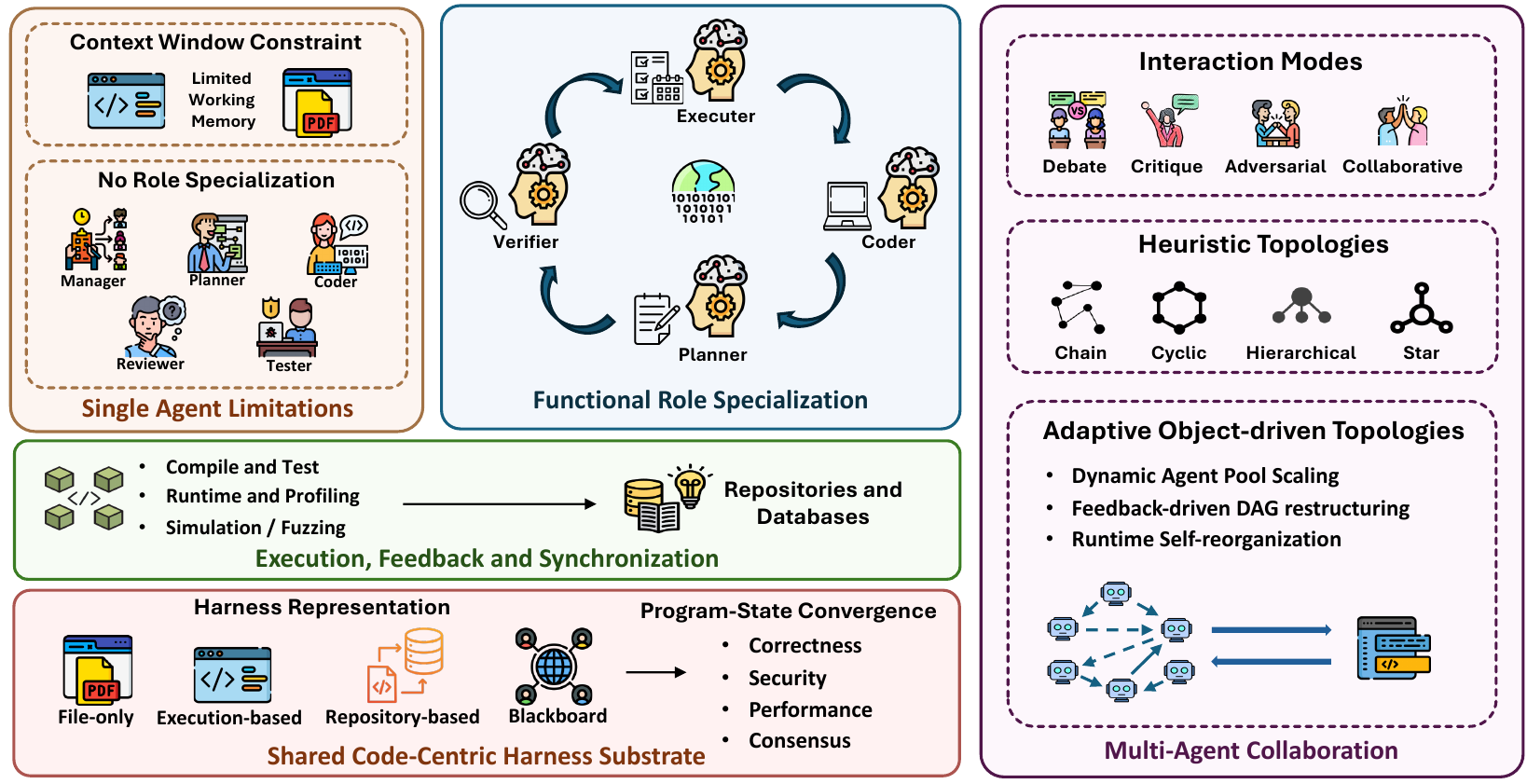}
    \caption{Overview of scaling the agent harness through multi-agent orchestration over code. The figure illustrates how role-specialized agents, shared code-centric substrates, execution feedback, and adaptive collaboration topologies address single-agent limitations in context, specialization, and self-correction.}
    \label{fig:mas-overview}
\end{figure}

\begin{figure}[t!]
    \centering
    \includegraphics[width=0.85\linewidth]{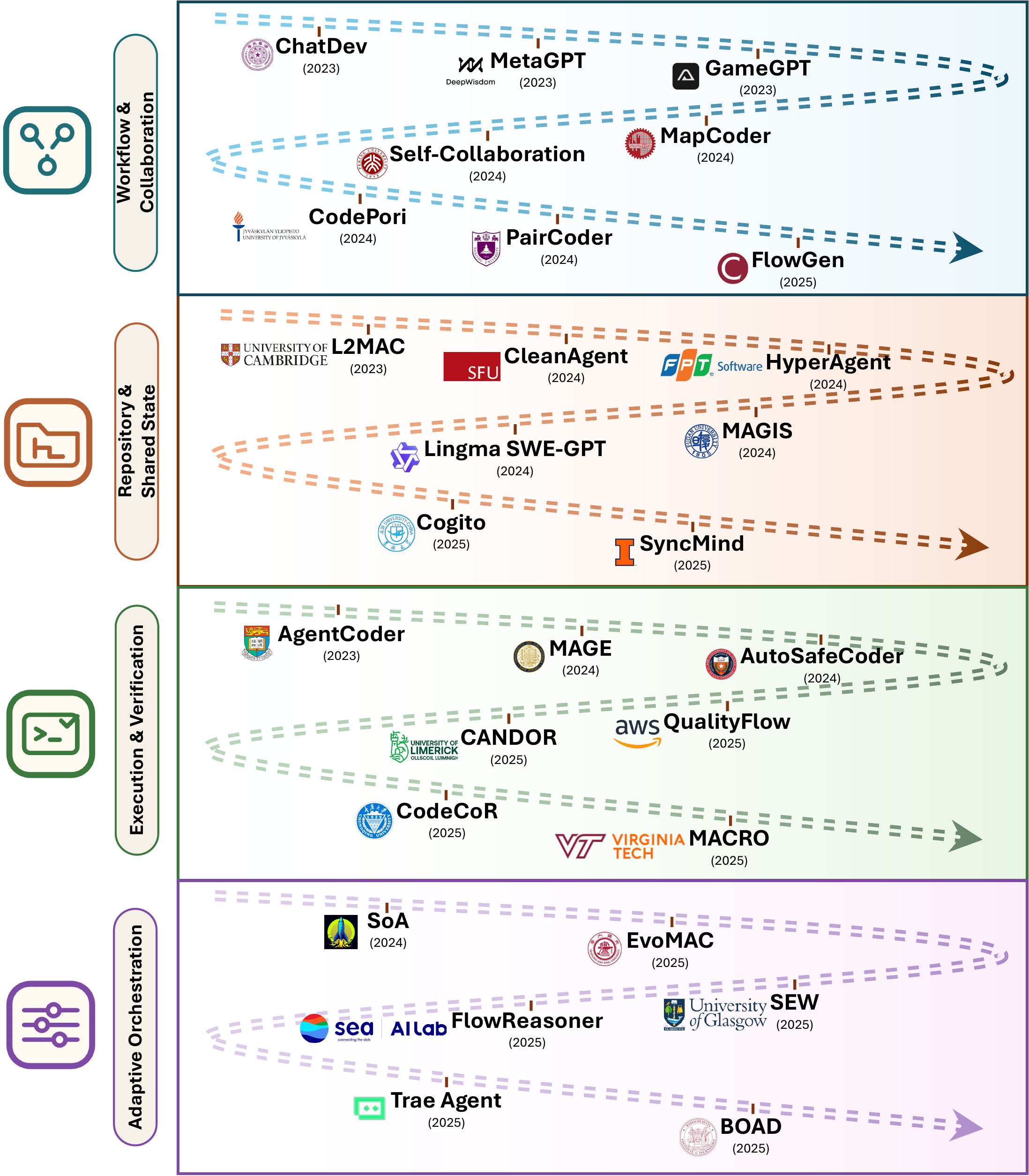}
    \caption{Roadmap of scaling code harnesses for multi-agent orchestration, organized by workflow collaboration, shared repository state, execution verification, and adaptive coordination.}
    \label{fig:roadmap_sec4}
\end{figure}

\subsection{Improved Coding Support through Multi-agent Collaboration}

The most immediate contribution of multi-agent systems is that
they improve coding support by decomposing the harness into
specialized but coordinated components. Instead of integrating
planning, synthesis, execution, and verification into a single
agent loop, these systems distribute responsibility across roles
that interact through shared code artifacts and feedback signals.
This division of labor makes the overall harness more capable of
handling complex software tasks, while also making its internal
workflow more inspectable and controllable. In practice, this
improvement is realized through three closely related design
dimensions: how roles are specialized, how agents interact over
shared program artifacts, and how the workflow topology organizes
their collaboration.

\begin{table}[t]
\centering
\caption{Representative MAS collaboration designs by role specialization and interaction structure. }
\label{tab:mas_collaboration_systems}
\renewcommand{\arraystretch}{1.16}
\setlength{\tabcolsep}{3.5pt}
\footnotesize
\begin{tabularx}{\textwidth}{@{}
  >{\raggedright\arraybackslash}p{2.55cm}
  >{\raggedright\arraybackslash}p{2.85cm}
  >{\raggedright\arraybackslash}p{3.05cm}
  >{\raggedright\arraybackslash}p{3.05cm}
  >{\raggedright\arraybackslash}X@{}}
\toprule
\textbf{System}
& \textbf{Harness Substrate}
& \textbf{Agent Roles}
& \textbf{Interaction Mode}
& \textbf{Topology} \\
\midrule
Self-Collaboration~\cite{Dong2024SelfCollaboration}
& Blackboard, implicit
& Plan, Synth., Verif. (simulated)
& Critique-repair
& Pre-defined cyclic \\ \midrule
CodePori~\cite{Rasheed2024Codepori}
& Implicit
& Plan, Synth., Verif.
& Collab-Synth., critique-repair
& Pre-defined chain, cyclic \\ \midrule
MAGIS~\cite{Tao2024Magis}
& Repository, evolution memory
& Plan, Understand, Synth., Verif.
& Critique-repair, debate, delegation
& Hierarchical, cyclic, dynamic pool \\ \midrule
HyperAgent~\cite{Phan2024HyperAgent}
& Repository, execution
& Plan, Understand, Synth., Exec
& Critique-repair
& Pre-defined hierarchical, cyclic \\ \midrule
PairCoder~\cite{Zhang2024PairProgramming}
& Execution
& Plan-Understand, Synth-Exec
& Collab-Synth., critique-repair
& Pre-defined cyclic with conditional branch \\ \midrule
FlowGen~\cite{Lin2025Soen101}
& Execution, implicit
& Plan, Understand, Synth., Verif.
& Critique-repair, debate
& Pre-defined chain, cyclic (Scrum) \\ \midrule
Trae Agent~\cite{gao2025traeagent}
& Repository, execution
& Generate, Prune, Select
& Collab-Synth., search (selection)
& Pre-defined search pipeline \\ \midrule
BOAD~\cite{xu2025boad}
& Repository, execution
& Orchestrate, Localize, Edit, Validate
& Delegation, adaptive selection
& Adaptive hierarchical \\ \midrule
FlowReasoner~\cite{gao2025flowreasoner}
& Execution, implicit
& Meta-design, Solve
& Runtime workflow generation
& Objective-driven adaptive \\ \midrule
ChatDev~\cite{Qian2023ChatDev}
& Implicit, borderline exec
& Plan, Synth., Verif., Exec
& Critique-repair, debate
& Pre-defined chain (waterfall) \\ \midrule
MetaGPT~\cite{Hong2023MetaGPT}
& Implicit, partial blackboard
& Plan$\times$3, Synth., Verif.
& Critique-repair, pub-sub scheduling
& Pre-defined chain (waterfall) \\ \midrule
GameGPT~\cite{chen2023gamegpt}
& Blackboard (dual collaboration)
& Plan, Synth., Verif.
& Critique-repair, collaborative
& Pre-defined \\
\bottomrule
\end{tabularx}
\end{table}

\subsubsection{Functional Role Specialization and Human-Guided Planning}
In human software development, different roles specialize in
different aspects of the development process. Many MAS naturally
mirror this division of labor by assigning distinct functional
roles to different agents. This specialization allows each agent
to focus on a specific slice of the shared code harness,
leveraging its unique capabilities and perspectives to contribute
to the overall task.
Here, we elaborate on the most common functional roles identified
across the surveyed literature, noting that many systems
implement multiple roles and that the boundaries between them can
be fluid.

\paragraph{Program synthesis agents} Program synthesis agents are
responsible for generating or transforming code. They consume
specifications, plans, or feedback signals and produce or revise
code artifacts. This is the most common role across surveyed
systems. Representative instances include the Coder in
Self-Collaboration~\cite{Dong2024SelfCollaboration}, the
Programmer in AgentCoder~\cite{huang2023agentcoder}, the Engineer
in MetaGPT~\cite{Hong2023MetaGPT}, the Developer in
ChatDev~\cite{Qian2023ChatDev}, and the RTL Generation Agent in
MAGE~\cite{Zhao2024MAGE}.

\paragraph{Program understanding agents} Program understanding
agents analyze existing code or specifications to produce
higher-level representations. They own the interpretation of what
the code means rather than what it does. This category includes
the Repository Custodian in MAGIS~\cite{Tao2024Magis}, the
Navigator in HyperAgent~\cite{Phan2024HyperAgent}, the RepoUer in
Lingma SWE-GPT~\cite{Ma2024Lingma}, and the Column-type Annotator
in CleanAgent~\cite{Qi2024CleanAgent}.

\paragraph{Verification agents} Verification agents evaluate code
quality, typically by generating test cases, running static
analysis, or simulating execution. The Test Designer in
AgentCoder~\cite{huang2023agentcoder} generates test cases
independently of the code to avoid circular reasoning, a design
principle against the mode-collapse problem where an agent's
biased tests pass its own buggy code. The Test Quality Checker in
QualityFlow~\cite{Hu2025QualityFlow} addresses this at a
meta-level, filtering synthesized tests before they are used as
feedback. The Static Analyzer and Fuzzing Agent in
AutoSafeCoder~\cite{Nunez2024AutoSafeCoder} provide
security-oriented verification through static CWE analysis and
dynamic crash detection, respectively. The Panelists in
CANDOR~\cite{Xu2025Hallucination} independently audit oracle
correctness against natural language specifications rather than
against the code itself, deliberately avoiding contamination by
faulty implementations.

\paragraph{Execution agents} Execution agents interface directly
with the program runtime. Critically, the Test Executor in
AgentCoder~\cite{huang2023agentcoder} is a deterministic Python
script (not an LLM) which cleanly separates reasoning from
execution and grounds the feedback signal in objective program
behavior. The Executor in HyperAgent~\cite{Phan2024HyperAgent}
runs unit and integration tests via an interactive bash shell.
The Judge Agent in MAGE~\cite{Zhao2024MAGE} interfaces with RTL
simulation tools to produce per-clock-edge waveform snapshots.

\paragraph{Planning agents} Planning agents decompose the overall
software-development task into subtasks and assign them to
synthesis agents. The Architect and Project Manager in
MetaGPT~\cite{Hong2023MetaGPT}, the Manager in
MAGIS~\cite{Tao2024Magis}, the Scrum Master in
FlowGen~\cite{Lin2025Soen101}, and the Mother agents in
SoA~\cite{Ishibashi2024SelfOrganized} all perform task
decomposition. The Mother agents in
SoA~\cite{Ishibashi2024SelfOrganized} are particularly notable:
they dynamically spawn Child agents at runtime based on the
inferred complexity of each subfunction, making planning and
agent initialization interdependent.

A distinctive feature of EvoMAC~\cite{Hu2025EvoMAC} is the
introduction of two novel meta-roles not present in any other
surveyed system: the Gradient Agent, which reads execution logs
to identify which agents caused failures, and the Updating Agent,
which revises agent prompts and restructures the workflow DAG
accordingly. These roles operate at the level of the MAS itself
rather than the program, enabling the system to adapt its own
structure in response to execution feedback.

\subsubsection{Diverse Interaction Modes Grounded in Shared Program State}
Unlike general MAS where agent interaction is primarily
message-passing, code-centric interaction is characterized by
artifact-mediated communication: agents observe and modify shared
code artifacts, and their interaction is grounded in the
objective state exposed by those artifacts and their execution
results. {These coordination channels are broader
than source code alone: agents communicate through APIs, files,
diffs, tests, logs, schemas, blackboards, and explicit workflow
states. In most systems, these channels are part of the
human-designed harness, while agents dynamically write to or
modify the artifacts circulating within them.} We identify four
interaction modes.

\paragraph{Collaborative synthesis} Collaborative synthesis occurs
when two agents jointly construct a program component, analogous
to pair programming \cite{zou2026recursivemas}. The Navigator--Driver pairing in
PairCoder~\cite{Zhang2024PairProgramming} is the most direct
instantiation: the Navigator generates and selects solution plans
while the Driver implements them, with bidirectional information
flow. CodePori~\cite{Rasheed2024Codepori} implements
collaborative synthesis between Dev\_01 and Dev\_02, who exchange
code drafts across three rounds. This mode is relatively rare among the surveyed system, as most systems prefer a sequential handoff
rather than true co-construction.

\paragraph{Critique and repair} Critique and repair is the
dominant interaction mode across the surveyed literature. A
verification or evaluation agent inspects a code artifact and
produces structured feedback; a synthesis agent then revises the
artifact in response. This pattern appears in some form in
virtually every surveyed system. Its key design decisions are:
(a) whether the critique is LLM-simulated or execution-grounded
(Self-Collaboration~\cite{Dong2024SelfCollaboration} uses a
simulated LLM tester, while
AgentCoder~\cite{huang2023agentcoder} uses a real Python
executor); (b) the richness of the feedback signal (ranging from
binary pass/fail in SEW~\cite{Liu2025SEW} to structured execution
logs enumerating satisfied requirements, function errors, and
unmet requirements in EvoMAC~\cite{Hu2025EvoMAC}); and (c) the
number of repair iterations permitted before fallback.

\paragraph{Adversarial validation} Adversarial validation is a
more active form of verification in which one agent attempts to
break the code through adversarial inputs, rather than passively
reviewing it. AutoSafeCoder~\cite{Nunez2024AutoSafeCoder}
implements this via its Fuzzing Agent, which generates
crash-inducing input seeds using type-aware mutation and executes
the code to produce crash traces. This mode has a fundamentally
different character from critique-and-repair: the fuzzer does not
explain what is wrong, but demonstrates a concrete execution
failure, a counterexample that the coding agent must address.
MAGE~\cite{Zhao2024MAGE} similarly uses simulation mismatch as an
adversarial signal: the Debug Agent receives the exact waveform
window around the first clock-edge failure, enabling targeted
repair.

\paragraph{Reasoning debate} Reasoning debate involves agents
arguing over the correctness of a decision or the interpretation
of a specification, before arriving at a consensus.
ChatDev~\cite{Qian2023ChatDev} introduces communicative
de-hallucination, a mechanism in which the assistant agent
reverses roles to ask clarifying questions before committing to a
response. The Scrum sprint meetings in
FlowGen~\cite{Lin2025Soen101} enable disordered multi-agent
discussion around a shared context buffer before the Scrum Master
synthesizes a decision. CANDOR~\cite{Xu2025Hallucination}
implements the most structured debate mechanism: three
independent Panelists evaluate oracle correctness, and a Curator
aggregates their verdicts via majority vote. The kick-off meeting
in MAGIS~\cite{Tao2024Magis} involves a circular speech among the
Manager and all Developer agents to negotiate task dependencies
and prevent conflicts.

\subsubsection{Optimized Workflow Topology for Agentic Coordination}

The topology of agent interaction, who communicates with whom, in
what order, and how many times, is one of the most consequential
design decisions in a MAS for code generation. We organize
topologies along two primary axes.

\paragraph{Pre-defined Heuristic Topologies}
The majority of surveyed systems use topologies that mirror
established software development life cycle (SDLC) models. These
topologies are fixed at design time and do not change in response
to task complexity or system performance.

\textbf{\textit{Chain (Waterfall) topologies}} sequence agents in
a strict linear order, with artifacts flowing unidirectionally
from planning to synthesis to verification. ChatDev~\cite{Qian2023ChatDev}
and MetaGPT~\cite{Hong2023MetaGPT} are canonical examples,
explicitly modeling the waterfall SDLC: design $\rightarrow$
coding $\rightarrow$ testing. FlowGen~\cite{Lin2025Soen101}
operationalizes three SDLC models as distinct topologies:
FlowWater (strict waterfall chain), FlowTDD (requirement
$\rightarrow$ design $\rightarrow$ test $\rightarrow$
implementation $\rightarrow$ fix, a test-driven reordering), and
FlowScrum (cyclic iterative sprints). This paper is unique in
directly comparing the implications of different
SDLC-mirroring topologies for code quality.
L2MAC~\cite{Holt2023L2MAC} also follows a chain topology but with
a novel twist: each step in the instruction plan is executed by a
fresh-context agent, making the chain a sequence of independent
LLM invocations sharing only the external file store.

\textbf{\textit{Cyclic (Agile/Iterative) topologies}} introduce
back-edges that allow code to be revised in response to
verification feedback. AgentCoder~\cite{huang2023agentcoder}
implements a programmer $\rightarrow$ test executor $\rightarrow$
(if fail) $\rightarrow$ programmer cycle, bounded at 5
iterations. Self-Collaboration~\cite{Dong2024SelfCollaboration}
embeds a coder $\leftrightarrow$ tester back-edge within its
waterfall chain, max 4 iterations.
PairCoder~\cite{Zhang2024PairProgramming} enhances the cyclic
pattern with multi-plan exploration: a pool of $n$ solution plans
is pre-generated via k-means++ clustering for diversity, and the
cycle can switch to the next candidate plan when dead-end is
detected through history-based loop analysis.
MAGE~\cite{Zhao2024MAGE} combines a linear initialization chain
with a cyclic debug-judge loop, and introduces high-temperature
candidate sampling to explore multiple program variants
simultaneously.

\textbf{\textit{Hierarchical topologies}} place one or more
manager agents above a pool of worker agents, enabling
decomposition-and-delegation patterns. MAGIS~\cite{Tao2024Magis}
has a Manager that dynamically instantiates one Developer agent
per candidate file at runtime; each Developer edits its assigned
file and reports back to the manager-review layer. HyperAgent~\cite{Phan2024HyperAgent}
uses a planner above multiple repository navigation and editing
workers, combining top-down decomposition with bottom-up
repository evidence. SoA~\cite{Ishibashi2024SelfOrganized}
pushes this hierarchy further by allowing Mother agents to spawn
Child agents recursively according to inferred subtask
complexity. These systems treat harness orchestration itself as a
resource-allocation problem.

\textbf{\textit{Star topologies}} center on a hub agent that
coordinates multiple parallel worker agents. The
CANDOR~\cite{Xu2025Hallucination} Stage 3 panel is an example: a
Requirement Engineer fans out to three independent
Panelist+Interpreter pipelines, and the Curator aggregates their
outputs. MetaGPT~\cite{Hong2023MetaGPT}'s publish-subscribe
message pool creates a de facto star topology where the shared
pool serves as the hub.

\paragraph{Objective-driven and Adaptive Topologies}
A smaller but rapidly growing class of systems treats the
topology itself as a design variable to be optimized toward a
code quality signal. Recent systems such as
FlowReasoner~\cite{gao2025flowreasoner} and
BOAD~\cite{xu2025boad} further reinforce this trend by treating
multi-agent organization itself as an adaptive object to be
generated, searched, or optimized per task.

\textbf{\textit{Dynamic agent pool scaling}} is the simplest form
of adaptivity: the number of agents scales with task complexity,
but the topology type is fixed.
SoA~\cite{Ishibashi2024SelfOrganized} implements this via a
hierarchical tree of Mother and Child agents, where Mother agents
decide at runtime how many subfunctions to decompose into,
spawning corresponding Child agents. The key insight is that each
agent's context window remains bounded, as complexity is handled
by growing the agent pool rather than growing individual context
windows. MAGIS~\cite{Tao2024Magis} similarly instantiates
Developer agents dynamically based on the number of candidate
files identified during repository analysis. BOAD~\cite{xu2025boad} extends this line of thought from dynamic
scaling to hierarchy discovery: instead of manually fixing the
specialized sub-agent structure, it formulates the selection of
helpful localization, editing, and validation sub-agents as a
bandit-optimization problem, showing that automatically
discovered hierarchical teams can outperform manually designed ones.

\textbf{\textit{Feedback-driven DAG restructuring}} is best
represented by EvoMAC~\cite{Hu2025EvoMAC}. Its workflow is a DAG
whose nodes correspond to agents and whose edges define
information flow. After each iteration, a Gradient Agent reads
execution logs to attribute failures to agents, and an Updating
Agent modifies the prompts and graph structure. This is the only
system in the survey where the harness topology is structurally
modified in response to execution feedback.

\textbf{\textit{Runtime self-reorganization}} is
SEW~\cite{Liu2025SEW}'s approach: the system generates and mutates
entire workflow specifications using Direct Evolution (DE) and
Hyper Evolution (HE) operators applied to LLM-generated workflow
descriptions in structured formats (BPMN, CoRE, Python, YAML).
Rather than optimizing agent parameters, SEW~\cite{Liu2025SEW}
optimizes the workflow structure including the sequence of agent
calls, the routing logic, and the feedback paths. The two
canonical topologies it discovers (a linear chain and a feedback
loop) emerge from optimization rather than being hand-designed. FlowReasoner~\cite{gao2025flowreasoner} pushes this adaptive view
further by training a query-level meta-agent that generates a
tailored multi-agent system for each input problem under external
execution feedback, making topology selection itself part of the
deliberative inference process rather than a fixed system design.

\subsection{Execution Feedback and Shared-Harness Synchronization}

We discuss how a group of agents can exploit the executability of
code, and how they maintain a consistent shared view of the
program state.
This dimension is the defining one for code-centric MAS: the
shared harness is uniquely executable and produces objective
oracle signals. We address two sub-questions: what types of
execution feedback are used, and how is shared state
synchronized across agents.

\begin{table}[t]
\centering
\caption{Representative MAS execution-feedback and convergence designs.}
\label{tab:mas_feedback_systems}
\renewcommand{\arraystretch}{1.2}
\setlength{\tabcolsep}{5pt}
\footnotesize
\begin{tabularx}{\textwidth}{@{}
  >{\raggedright\arraybackslash}p{2.6cm}
  >{\raggedright\arraybackslash}p{3.8cm}
  >{\raggedright\arraybackslash}p{2.3cm}
  >{\raggedright\arraybackslash}p{3.3cm}
  >{\raggedright\arraybackslash}X@{}}
\toprule
\textbf{System} & \textbf{Harness Substrate} & \textbf{Topology} & \textbf{Execution Feedback} & \textbf{Convergence} \\
\midrule
\multicolumn{5}{@{}l}{\textit{Pre-defined topology}} \\
\addlinespace[2pt]
AgentCoder~\cite{huang2023agentcoder} & Execution & Cyclic & Test pass/fail & Correctness (test-gated) \\
MAGE~\cite{Zhao2024MAGE} & Execution (waveform) & Chain-cyclic & Checkpoint waveform & Score-based correctness \\
MapCoder~\cite{islam2024mapcoder} & Execution, implicit & Cyclic & Test pass/fail & Correctness \\
AutoSafeCoder~\cite{Nunez2024AutoSafeCoder} & Execution (static, fuzzer) & Cyclic & CWE warnings, crashes & Security convergence \\
QualityFlow~\cite{Hu2025QualityFlow} & Execution (real, imagined) & Gated cyclic & Pass/fail, imagined exec & Correctness (quality-gated) \\
CodeCoR~\cite{Pan2025CodeCoR} & Execution, implicit & Cyclic & Syntax, test pass/fail & Score-based soft correctness \\
MARCO~\cite{Rahman2025MACRO} & Execution (performance) & 2-node Cyclic & Time, memory, FLOPS & Performance, correctness \\
\midrule
\multicolumn{5}{@{}l}{\textit{Adaptive topology}} \\
\addlinespace[2pt]
SoA~\cite{Ishibashi2024SelfOrganized} & Execution, implicit gap & Hierarchical tree & Test pass/fail & Correctness (implicit fallback) \\
SEW~\cite{Liu2025SEW} & Implicit & Evolution & Test pass/fail & Implicit \\
EvoMAC~\cite{Hu2025EvoMAC} & Execution & Text DAG & Compiler, execution logs & Correctness (fixed-iteration) \\
FlowReasoner~\cite{gao2025flowreasoner} & Execution, implicit & Query workflow & Execution feedback & Objective-driven adaptive \\
Trae Agent~\cite{gao2025traeagent} & Repository, execution & Search pipeline & Test, pruning signals & Score-/selection-based \\
\bottomrule
\end{tabularx}
\end{table}

\subsubsection{Execution Feedback Integration}
\paragraph{Compiler and syntax feedback} Compiler and syntax
feedback catch structural errors before runtime and are used by
many systems. ChatDev~\cite{Qian2023ChatDev} feeds compiler
errors from the testing phase back to the programmer, though only
as one-off corrections within a single phase.
L2MAC~\cite{Holt2023L2MAC} runs syntax checks via its evaluator
module $E(D)$ after every file write, treating them as blocking
conditions that prevent the instruction pipeline from advancing.

\paragraph{Test pass/fail signals} Test pass/fail signals are the
most commonly used execution-feedback type.
AgentCoder~\cite{huang2023agentcoder} centers its entire loop on
whether independently generated test cases pass; the iteration
terminates on full pass or at the 5-iteration budget.
QualityFlow~\cite{Hu2025QualityFlow} introduces a notable
variant: Imagined Execution, in which an LLM simulates the Python
interpreter step-by-step and predicts test outcomes without
actually running the code, achieving 98\%+ precision and recall
on MBPP while avoiding label leakage from visible test cases.
The near-identical performance of
Self-Collaboration~\cite{Dong2024SelfCollaboration}'s simulated
LLM tester and its real-compiler ablation raises a provocative
empirical question: when is actual execution necessary, and when
can linguistic simulation of execution suffice?

\paragraph{Fuzzer crash traces} Fuzzer crash traces represent a
qualitatively different type of feedback: rather than a pass/fail
outcome, they provide a concrete failing input.
AutoSafeCoder~\cite{Nunez2024AutoSafeCoder} uses type-aware
mutation to generate crash-inducing input seeds and passes the
crashing input plus exit code to the Coding Agent. This
adversarial feedback is more informative than a generic failure
signal because it localizes the bug to a specific input category.

\paragraph{Static analysis warnings} Static analysis warnings
provide feedback about code structure and security properties
without execution. AutoSafeCoder~\cite{Nunez2024AutoSafeCoder}
uses CWE-mapped static analysis against the MITRE vulnerability
database, enabling the Static Analyzer Agent to suggest
remediation strategies keyed to specific vulnerability classes.

\paragraph{Performance profiling results} Performance profiling
results are uniquely exploited by MACRO~\cite{Rahman2025MACRO},
which treats code optimization as the primary task rather than
correctness. The Performance Evaluator Agent measures execution
time, memory usage, and FLOPS, and MACRO~\cite{Rahman2025MACRO} uniquely augments this
with real-time web search to retrieve relevant optimization
techniques from the research literature.

\paragraph{Fine-grained simulation feedback} MAGE~\cite{Zhao2024MAGE}'s
distinctive contribution is the finest-grained execution feedback
in the surveyed literature. Rather than reporting only whether a
testbench passes or fails, the State Checkpoint mechanism records
signal values at every clock edge and delivers to the Debug Agent
a waveform window around the first failing clock cycle. This
enables targeted repair at sub-test granularity.

\subsubsection{Shared-Harness Synchronization}

Sequential handoff is the most common synchronization mechanism:
each agent receives the output of its predecessor and passes its
own output to its successor. The program state exists only in the
form of the most recent artifact in the pipeline. This is
sufficient for simple linear pipelines but creates invisible
state divergence in multi-agent settings where multiple agents
modify the codebase in parallel or iteratively. {It
is also where the limits of code-mediated coordination become
clear. Even when agents share executable artifacts, the harness
still imposes information-theoretic constraints: channels have
finite bandwidth, summaries introduce compression loss, logs
become noisy, cached views go stale, and parallel branches raise
unresolved questions of authority and consistency. Code provides
a richer substrate for coordination, but it does not remove
these distributed-systems constraints.}

\paragraph{Shared blackboard} Shared blackboard provides a
globally accessible program state that all agents can read and
update. L2MAC~\cite{Holt2023L2MAC} implements this most cleanly:
the file store $D$ is an external, persistent structure that is
never overwritten but extended and revised. The Control Unit
manages all reads and writes, ensuring that each agent invocation
receives a precisely controlled context window.
MAGIS~\cite{Tao2024Magis}'s repository evolution memory $M$ is a
persistent key-value store mapping file versions to
LLM-generated summaries, updated incrementally via a specialized
blackboard for repository-level reasoning.
Self-Collaboration~\cite{Dong2024SelfCollaboration} is among the
first systems to explicitly name and invoke the blackboard
architecture, establishing a shared memory from which all three
roles read and write.

\paragraph{Parallel branches with merge} Parallel branches with
merge arise when multiple agents modify independent components
simultaneously, with their changes integrated at a later stage.
MAGIS~\cite{Tao2024Magis} instantiates one Developer per
candidate file; each modifies its assigned file independently,
and all changes are merged into the final repository diff.
HyperAgent~\cite{Phan2024HyperAgent} runs multiple Navigator and
Editor instances in parallel via Redis queues, with results
merged at the Planner level.

\paragraph{Structured context scheduling} Structured context
scheduling is the explicit management of what each agent sees and
when. It is the primary innovation of
L2MAC~\cite{Holt2023L2MAC}. The Control Unit resets the context
window between instruction steps, providing each new invocation
with a targeted summary of prior progress $(M_{rs})$ rather than
the full conversation history. When the context window approaches
capacity, the Control Unit stores partial results to $D$ and
re-initializes with a compressed view, explicitly instructing the
LLM which files to read or skip given the remaining context
margin. This mechanism solves the context-window problem not by
expanding the window but by carefully controlling its contents.
MetaGPT~\cite{Hong2023MetaGPT} implements a lighter form of
context scheduling via a publish-subscribe message pool: each
agent subscribes only to the document types relevant to its role,
receiving a filtered view of the shared state.

\paragraph{Hierarchical memory} Hierarchical memory combines
short-term working context with longer-term accumulated
knowledge. ChatDev~\cite{Qian2023ChatDev} explicitly separates
short-term memory (full dialogue within a phase) from long-term
memory (extracted solutions carried across phases).
Cogito~\cite{Li2025Cogito} implements hierarchical memory, drawing on
neurobiological architecture: short-term memory for immediate
task state, a long-term knowledge base for accumulated expertise,
and growth units for evolving abstractions that improve over
time. HyperAgent~\cite{Phan2024HyperAgent} uses a lightweight
LLaMA-3.1-8B summarizer to condense execution logs before storing
them in hierarchical memory, preventing context bloat.

\paragraph{Agent pool scaling} Agent pool scaling addresses the
context-management problem orthogonally: rather than managing
what a single agent sees, it distributes the context load across
more agents. SoA~\cite{Ishibashi2024SelfOrganized} is the
canonical example: by spawning more agents as task complexity
grows, each agent's context remains bounded. This is a
structural solution to the harness-state problem: instead of
building a shared representation that all agents can query,
SoA~\cite{Ishibashi2024SelfOrganized} partitions the task state
across agents, each holding a bounded slice. The limitation is
that global consistency is sacrificed: agents cannot reason about
the full program, only their assigned sub-function.

\paragraph{Other}
QualityFlow~\cite{Hu2025QualityFlow}'s revert mechanism
represents a synchronization pattern: the initial code artifact is never overwritten,
enabling the system to roll back to a prior shared harness state
if the debugging trajectory degrades quality. This is the only
work among the surveyed system that explicitly manages state history rather than always
moving forward.

\subsection{Position: The Shared Code-Centric Harness Substrate}

We propose a new position for the next generation of multi-agent
intelligence: the shared code-centric harness substrate. This
position is motivated by the central gap identified in the
literature: the lack of formal, persistent representations of the
shared code state that agents can query and update across
iterations. We argue that building such a harness substrate is
both feasible and necessary for achieving robust, scalable
multi-agent intelligence.

\begin{table}[t]
\centering
\caption{Representative MAS designs centered on shared program-state representation and synchronization.}
\label{tab:mas_world_state_systems}
\renewcommand{\arraystretch}{1.16}
\setlength{\tabcolsep}{4pt}
\footnotesize
\begin{tabularx}{\textwidth}{@{}
  >{\raggedright\arraybackslash}p{2.55cm}
  >{\raggedright\arraybackslash}p{3.05cm}
  >{\raggedright\arraybackslash}p{3.15cm}
  >{\raggedright\arraybackslash}p{3.05cm}
  >{\raggedright\arraybackslash}X@{}}
\toprule
\textbf{System}
& \textbf{Harness Substrate}
& \textbf{Agent Roles}
& \textbf{Execution Feedback}
& \textbf{Convergence / Synchronization} \\
\midrule
L2MAC~\cite{Holt2023L2MAC}
& Blackboard, repository, execution
& Plan, Synth, Verif (evaluator)
& Syntax, test pass/fail
& Correctness per instruction step \\ \midrule
Cogito~\cite{Li2025Cogito}
& Blackboard (3-tier memory)
& Neurobiological model
& NA
& Hierarchical memory synchronization \\ \midrule
CleanAgent~\cite{Qi2024CleanAgent}
& Execution (weak), implicit
& Plan, Understand, Synth, Exec
& Runtime errors
& Correctness through execution success \\ \midrule
Lingma SWE-GPT~\cite{Ma2024Lingma}
& Repository, execution
& Understand, Synth-Verif
& Syntax, git apply, tests
& Fixed-limit implicit convergence \\ \midrule
SyncMind~\cite{Guo2025SyncMind}
& Repository, execution (formal $S_k/B_k$)
& Synth-Understand, oracle Understand
& Test pass/fail, runtime errors
& Correctness, resource-constrained synchronization \\ \midrule
BOAD~\cite{xu2025boad}
& Repository, execution
& Orchestrator with specialized sub-agents
& Test pass/fail, validation reward
& Hierarchy discovery, coordination \\ \midrule
CANDOR~\cite{Xu2025Hallucination}
& Execution (Java, JaCoCo)
& Plan, Synth, Verif, Understand, Debate
& Compiler, coverage, tests
& Correctness, coverage, consensus \\
\bottomrule
\end{tabularx}
\end{table}

\subsubsection{Shared Harness Representation}

A foundational question for any MAS is: what is the substrate
these agents inhabit? In code as agent harness, the natural
answer is the shared program environment, namely the collection
of artifacts, execution contexts, and quality signals that agents
collectively act upon and that evolve as agents produce, revise,
and evaluate code. We call this the shared harness substrate, and
we distinguish four levels of formalization with which existing
systems represent it.

\paragraph{Implicit / File-only Representation}

The most common and least formalized category treats the shared
harness as simply the current code file or set of code files.
Agents receive the latest code artifact as part of their input
context and produce a modified or evaluated version. There is no
persistent, queryable representation: the shared state is
reconstructed implicitly at each agent invocation from the
conversational history. This category encompasses many
foundational systems: ChatDev~\cite{Qian2023ChatDev},
MetaGPT~\cite{Hong2023MetaGPT}, FlowGen~\cite{Lin2025Soen101},
MapCoder~\cite{islam2024mapcoder},
CodeCoR~\cite{Pan2025CodeCoR}, SEW~\cite{Liu2025SEW}, and
CodePori~\cite{Rasheed2024Codepori}. While this representation is
simple to implement, it entails a fundamental limitation: agents
cannot reason about the shared substrate except through the
narrow lens of their most recent context window.
State divergence~\cite{Guo2025SyncMind}, in which an agent's
internal belief about the code state diverges from the true
state, is invisible to the system and cannot be detected or
corrected.

\paragraph{Repository-based Representation}

A richer class of systems represents the shared harness as a
navigable repository: a file system with directory structure,
inter-file dependency graphs, call hierarchies, and version
history. This representation supports agents that reason about
where in the codebase a change needs to be made, what other
components depend on the changed function, and how the codebase
has evolved over time. MAGIS~\cite{Tao2024Magis} introduces a
repository evolution memory that caches file-level summaries and
incrementally updates them via git diff as files change across
issue-resolution episodes. HyperAgent~\cite{Phan2024HyperAgent}
provides agents with repository navigation tools
(get\_tree\_structure, go\_to\_definition, code\_search,
get\_all\_references), treating the repository as a structured
knowledge base. Lingma SWE-GPT~\cite{Ma2024Lingma} compresses the
repository view via abstract syntax tree (AST) skeletons,
preserving function signatures and class definitions to enable
efficient navigation. SyncMind~\cite{Guo2025SyncMind} is the only
work to formally define the repository substrate as a ground-truth
state $S_k$ and measure the divergence between $S_k$ and an
agent's belief state $B_k$.

\paragraph{Execution-based Representation}

Execution-based representation is the most distinctive category
for code generation. It has no direct parallel in general MAS and
represents the shared substrate through execution behavior. The
state is not what the code looks like but what the code does:
whether it compiles, which tests it passes, what vulnerabilities
a fuzzer uncovers, how fast it runs, and whether its runtime
behavior matches its specification. This execution-based
representation provides an objective oracle signal, a ground
truth that is not subject to the hallucination or bias that
affects purely linguistic agent evaluations. Systems that exploit
this representation include
AgentCoder~\cite{huang2023agentcoder},
AutoSafeCoder~\cite{Nunez2024AutoSafeCoder},
QualityFlow~\cite{Hu2025QualityFlow},
MACRO~\cite{Rahman2025MACRO}, EvoMAC~\cite{Hu2025EvoMAC},
CANDOR~\cite{Xu2025Hallucination}, and
MAGE~\cite{Zhao2024MAGE}. Notably,
MAGE~\cite{Zhao2024MAGE} achieves the finest-grained execution
feedback in the literature, operating at clock-edge granularity
via \textit{State Checkpoint} waveform snapshots.

\paragraph{Blackboard / Shared-State Representation}

A fourth category introduces an explicit, globally accessible
data structure that all agents can read from and write to (akin
to the classical blackboard architecture in
AI~\cite{erman1980hearsay}). This shared state is the closest
approximation in the literature to a formal harness substrate: it
persists across agent invocations, can be queried and updated,
and provides a consistent view of the program state to all
agents. Self-Collaboration~\cite{Dong2024SelfCollaboration} is
among the first systems to explicitly invoke the blackboard
metaphor, establishing a shared memory from which all three roles
(Analyst, Coder, Tester) read and write.
L2MAC~\cite{Holt2023L2MAC} implements the most principled
blackboard in the literature: a persistent file store $D$ with
semantically meaningful paths, accessed through a Control Unit
that explicitly manages which slice of state each agent
invocation sees. GameGPT~\cite{chen2023gamegpt} uses a shared
context buffer to reduce redundant information retransmission in
multi-round game development. Cogito~\cite{Li2025Cogito} draws on
neurobiological architecture to implement a three-tier memory:
short-term working state, long-term knowledge base, and growth
units for evolving abstractions, as a structured harness
representation.

\paragraph{The Central Gap}

The distribution of systems across these four categories reveals
a striking pattern: the majority of the literature resides in the
implicit/file-only category, lacking any formal model of the
shared harness substrate. This is the central gap that motivates
the code as agent harness framing. The program, uniquely among
multi-agent domains, is an artifact that executes. It produces
objective, non-linguistic signals that could in principle anchor
a formal shared substrate. Yet most systems fail to exploit this
property at the architectural level, instead relying on agents to
reason about code quality through natural language alone.

\subsubsection{Harness-State Convergence}

Convergence determines when a multi-agent coding harness should
stop iterating and accept its current program state as a
satisfactory outcome. In many existing MAS, convergence is still
defined implicitly, either by consensus among agents or by an
external iteration budget. However, code as agent harness has a
distinctive advantage: because the shared substrate is
executable, convergence can be grounded in objective behavioral
signals rather than in conversational agreement alone. We identify six convergence patterns, ranging from widely used test-gated and implicit convergence to less common security-, performance-, and consensus-based criteria.

\paragraph{Correctness convergence} Correctness convergence
(test-gated) is the most principled and widely used objective
criterion: the system terminates successfully when all test cases
pass. AgentCoder~\cite{huang2023agentcoder},
L2MAC~\cite{Holt2023L2MAC}, SyncMind~\cite{Guo2025SyncMind}, and
CANDOR~\cite{Xu2025Hallucination} implement test-gated
convergence. PairCoder~\cite{Zhang2024PairProgramming} augments
this with dead-end detection: if the same buggy code or feedback
appears in the iteration history, the system switches to the next
candidate plan rather than looping. FlowGen~\cite{Lin2025Soen101}
uses test-gated convergence but on LLM-generated tests rather
than ground-truth tests, introducing a potential quality concern:
a system can converge on code that passes its own biased tests
but fails on external evaluation.

\paragraph{Security convergence} Security convergence is uniquely
implemented by AutoSafeCoder~\cite{Nunez2024AutoSafeCoder}: the
system terminates successfully when no CWE vulnerabilities are
flagged by static analysis and no crashes are induced by the
fuzzer. This multi-criteria convergence is a strong argument for
the execution-based harness framing. Both convergence criteria
are grounded in objective program behavior, not agent opinions.

\paragraph{Performance convergence} Performance convergence is the
focus of MACRO~\cite{Rahman2025MACRO}: the optimization loop
terminates when user-defined runtime and memory thresholds are
satisfied, as measured by the Performance Evaluator against
actual execution benchmarks. This is the only system that treats
performance as the primary convergence criterion rather than
correctness.

\paragraph{Score-based convergence} Score-based convergence uses
quantitative quality scores computed by agents evaluating
intermediate outputs to determine when to stop.
MAGE~\cite{Zhao2024MAGE} ranks candidate programs by their
simulation mismatch score $s(r) = 1 - m(r) / tc(r)$ and
continues iterating until the maximum score reaches 1.0.
CodeCoR~\cite{Pan2025CodeCoR} uses a four-criteria binary score
(clarity, relevance, conciseness, context) to prune intermediate
outputs at each agent stage and selects the highest-ranked code
in its Ranked Code Set as the final output. It sets a soft
correctness convergence that submits the best available result
rather than waiting for a perfect solution. Trae Agent~\cite{gao2025traeagent} introduces a closely related
search-and-selection view at repository scale: it formulates
issue resolution as an optimal solution search problem and uses
modular generation, pruning, and selection agents to navigate a
large ensemble space of candidate patches. In this setting,
convergence is not only a matter of repeated repair, but also of
ranking, filtering, and selecting among competing solutions under
repository-aware evidence.

\paragraph{Consensus convergence} Consensus convergence aggregates
judgments from multiple reviewer agents.
CANDOR~\cite{Xu2025Hallucination} implements majority voting
among three Panelists on oracle correctness.
MAGIS~\cite{Tao2024Magis} uses LLM-judgment from the QA Engineer
as the acceptance signal, though this is a single-agent consensus
rather than a multi-agent vote. QualityFlow~\cite{Hu2025QualityFlow}
uses its Code Quality Checker as the single gating signal. It is
an efficient design where the quality checker serves as both a
convergence oracle and the system controller, enabling early exit
(75--84\% of problems converge after the first generator call).

\paragraph{Implicit convergence} Pipeline termination after a
fixed number of stages or iterations with no objective quality
criterion is the most prevalent convergence pattern in the
literature and represents the most significant gap in the field.
ChatDev~\cite{Qian2023ChatDev} terminates after a fixed number of
phases, or when two consecutive rounds produce identical code, or
after 10 rounds, none of which is an objective quality signal.
MetaGPT~\cite{Hong2023MetaGPT} terminates after completing the
fixed SOP stages.
Self-Collaboration~\cite{Dong2024SelfCollaboration} falls back to
implicit convergence after $n = 4$ iterations if the tester never
approves. EvoMAC~\cite{Hu2025EvoMAC} runs a fixed $K$ iterations
of the textual backpropagation loop. The prevalence of implicit
convergence is a direct consequence of the lack of formal shared
substrates: without an objective representation of the program
state, systems have no principled criterion for convergence.

\subsection{Patterns and Trends}

Across systems, differences in role specialization, shared-state representation,
execution grounding, and workflow topology are not independent
engineering choices; they interact to determine how reliably a
group of agents can maintain coherence over long-horizon coding
tasks. In this subsection, we distill the main
trends that emerge from the surveyed systems, highlighting both the common
structural bottlenecks of current systems and the design
principles that point toward more robust shared harnesses.

\paragraph{The implicit-harness-state constraint} The majority of
surveyed systems (ChatDev~\cite{Qian2023ChatDev},
MetaGPT~\cite{Hong2023MetaGPT}, FlowGen~\cite{Lin2025Soen101},
CodePori~\cite{Rasheed2024Codepori}, SEW~\cite{Liu2025SEW},
MapCoder~\cite{islam2024mapcoder},
CodeCoR~\cite{Pan2025CodeCoR}) operate without explicit
representations of the shared code harness. These systems rely on
agents to reconstruct state implicitly from conversational
history at each invocation. This design choice works for
function-level tasks where the program state is simple and does
not fragment across agents. However, this implicit approach
creates a fundamental vulnerability: without a formal shared
substrate, agents cannot reliably detect when their internal
understanding diverges from the true program
state~\cite{Guo2025SyncMind}. From the code as agent harness
perspective, the reliance on implicit state representations is
the technical root of system brittleness rather than a
scalability convenience.

\paragraph{Code-mediated channels do not eliminate coordination bottlenecks}
{The shift from free-form dialogue to code-mediated
coordination is a genuine architectural advance, but it should
not be overstated. Files, APIs, diffs, tests, logs, schemas,
blackboards, and workflow states are all partial channels
through which task state is encoded, transmitted, and
reconstructed. Each channel trades off fidelity, latency, and
scope: tests compress semantics into pass/fail, summaries save
context at the cost of detail, logs are grounded but noisy, and
shared blackboards improve persistence while creating authority
and consistency problems. The central design question is
therefore not merely whether code is present, but which
artifacts are authoritative, how they are compressed, and how
conflicts across channels are resolved.}

\paragraph{Execution feedback as the bridge between linguistic and formal reasoning}
The deepest divide in the literature is between systems that use
execution as ground truth and those that rely on linguistic model
judgments. Systems that ground shared state in execution
(AgentCoder~\cite{huang2023agentcoder},
AutoSafeCoder~\cite{Nunez2024AutoSafeCoder},
QualityFlow~\cite{Hu2025QualityFlow}, EvoMAC~\cite{Hu2025EvoMAC},
MAGE~\cite{Zhao2024MAGE}) have access to objective oracle
signals, signals that cannot hallucinate. Yet a surprising
finding complicates this picture:
Self-Collaboration~\cite{Dong2024SelfCollaboration} and
QualityFlow~\cite{Hu2025QualityFlow} demonstrate that
LLM-simulated execution can achieve 98\%+ precision and recall in
predicting actual outcomes without running code. This suggests
that execution feedback's value is not uniform across all failure
modes. It excels at detecting the corner cases that linguistic
simulation structurally cannot imagine (runtime crashes, resource
exhaustion, boundary condition errors, performance regressions),
but for many correctable bugs, simulated reasoning may suffice. A
mature harness would integrate both: using linguistic reasoning
as the fast path and delegating to execution as the verification
oracle only for the failure modes that require it.

\paragraph{Two complementary representations of the shared harness}
The surveyed systems reveals two conceptually orthogonal views:
repository-based representation (structure: what functions call
what, where does data flow, what are the dependencies) and
execution-based representation (behavior: what does the code do
when run, how does state evolve at runtime, what emergent
failures occur under different inputs). MAGIS~\cite{Tao2024Magis}
and HyperAgent~\cite{Phan2024HyperAgent} operate primarily in the
repository view, enabling agents to reason about codebase
architecture. AgentCoder~\cite{huang2023agentcoder} and
MAGE~\cite{Zhao2024MAGE} operate primarily in the execution view,
grounding shared state in runtime signals. Yet none of the
surveyed systems fully unifies both views into a single harness
substrate where agents can reason across both the static
structure of code and its dynamic behavior. The deepest harness
would integrate these two perspectives, answering questions like
``which components are slow'' (requires both call graphs and
profiling data) or ``does this refactoring break APIs that
external code depends on'' (requires both static analysis and
dynamic testing).

\paragraph{Topology complexity inversely correlates with harness-state formality}
Systems with explicit, formal shared substrates use simpler
topologies, while systems lacking formal shared state employ
increasingly complex topology patterns as a structural
workaround. L2MAC~\cite{Holt2023L2MAC}, which has the clearest
formal harness substrate (a persistent file store with explicit
context scheduling), uses a simple sequential chain with
sophisticated state management. By contrast,
implicit-state systems like EvoMAC~\cite{Hu2025EvoMAC} and
SEW~\cite{Liu2025SEW} develop elaborate adaptive topologies
(dynamic DAGs, workflow mutation, agent pool scaling) that
attempt to optimize the collaboration structure in the absence of
a principled shared representation. This suggests that topology
complexity is partially a symptom: when the substrate is
formally represented and queryable, agents can coordinate through
simple, transparent protocols. When the substrate is implicit,
agents require richer interaction patterns to compensate for
missing state information.

\paragraph{Context management is the tax of implicit shared state}
A striking pattern is that many systems have
developed sophisticated context-management mechanisms precisely
because they lack a formal shared substrate.
L2MAC~\cite{Holt2023L2MAC}'s Control Unit,
MetaGPT~\cite{Hong2023MetaGPT}'s publish-subscribe pool,
SoA~\cite{Ishibashi2024SelfOrganized}'s agent-pool scaling, and
Cogito~\cite{Li2025Cogito}'s three-tier memory are all responses
to the same underlying problem: how to give agents a coherent
view of a code harness that is too large to fit in any one
context window. A mature harness substrate could unify these
disparate solutions by providing a principled, queryable
representation of task state that agents access on demand,
rather than forcing the system to carefully manage what each
agent sees at every step.

\paragraph{Agent specialization increases the criticality of shared state metrics}
As agent role diversity increases, from basic coder-tester pairs
to systems with Architect, Manager, Navigator, Executor, and
Verifier roles, the need for a unified shared substrate becomes
urgent. Without shared understanding of code state, the Planning
Agent may decompose tasks based on an outdated codebase snapshot,
the Execution Agent may run tests against a different version
than the Synthesis Agent intended, and the Verification Agent's
feedback may misfire. EvoMAC~\cite{Hu2025EvoMAC} addresses this
through its Gradient and Updating agents that explicitly monitor
failure attribution at the MAS level.
SyncMind~\cite{Guo2025SyncMind} formalizes the problem as agent
belief divergence $|B_k - S_k|$, proposing explicit
synchronization protocols. The proliferation of agent roles is
thus not merely an engineering choice. It is a forcing function
for developing more mature shared harnesses. Multi-agent systems
with rich role repertoires cannot function robustly without them.

\section{Emerging Fields and Open Problems}

Having characterized code as an agent harness through its interfaces, mechanisms, and orchestration patterns, we now examine how this paradigm materializes in concrete application domains and what open problems it exposes. Across coding assistants, GUI/OS agents, scientific discovery, personalization, and embodied agents, code serves not only as a model output, but also as the operational substrate for state representation, action execution, memory, feedback, and governance. These domains make the promise of code-centric agentic systems tangible, while revealing a common set of unresolved challenges around evaluation, verification, safety, coordination, multimodal grounding, and harness evolution.

\begin{figure}[t]
    \centering
    \includegraphics[width=1.0\linewidth]{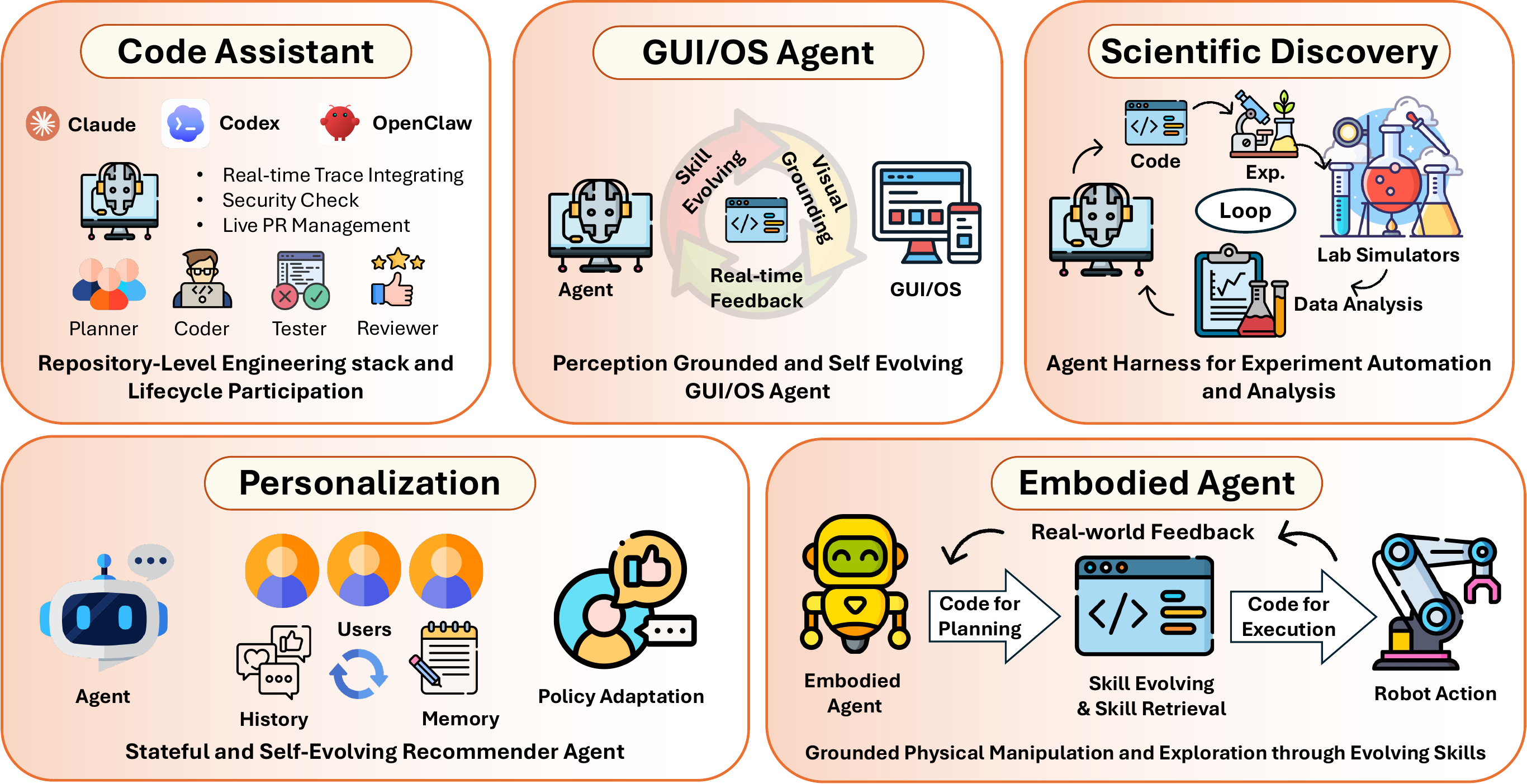}
    \caption{Overview of code as an agent harness across five emerging domains, including coding assistants, GUI/OS agents, scientific discovery, personalization, and embodied agents.}
    \label{fig:applications}
\end{figure}

\subsection{Emerging Fields and Tangible Applications}
This subsection surveys five application domains where code-as-harness systems have become especially visible. Code assistants operate over repositories, tests, development tools, and collaborative workflows; GUI and OS agents manipulate rendered interfaces through executable actions and programmatic checkers; scientific agents organize hypotheses, experiments, analyses, and laboratory protocols as executable pipelines; personalization agents adapt recommendation policies through structured user feedback and editable preference states; and embodied agents ground high-level intent in executable skills subject to physical constraints. Together, these domains show how code connects model outputs to real-world systems, and how the design of the surrounding harness shapes reliability, controllability, and long-horizon autonomy.

\subsubsection{Code Assistants}
Code assistants provide one of the clearest application domains where
code-centric agentic systems become operational. Early systems mainly supported
localized completion or single-turn code generation. Recent assistants instead
operate across repository-level workflows, where editing, tool use, validation,
and pull-request interaction form a closed-loop agent process. This shift is
reflected in research systems such as SWE-agent~\citep{yang2024swe} and
OpenHands~\citep{wang2024openhands}, as well as production-oriented platforms
such as Claude Code~\citep{claudecode2025}, Codex~\citep{codex2025}, GitHub
Copilot coding agents~\citep{copilotagent2025}, and
DeepAgents~\citep{deepagents2025}. In these systems, the assistant is no longer
a standalone code generator. It is embedded in a development environment where
repository state, tools, validation routines, and collaboration workflows
provide the operational context for action and feedback.

\paragraph{Repository-centered Workspace}
Modern code assistants operate over repositories rather than isolated code
snippets. Source files, tests, build scripts, dependency metadata, issues,
branches, and pull requests form a persistent workspace that the agent can
inspect, modify, and validate over multiple steps. This makes repository-level
assistance less a matter of placing relevant files in the prompt, and more a
matter of constructing a task-specific working view over a large and evolving
codebase. Systems such as RepoCoder~\cite{zhang2023repocoder}, CodexGraph~\cite{liu2024codexgraphbridginglargelanguage}, and AutoCodeRover~\cite{zhang2024autocoderover} address this
problem through repository indexing, dependency-aware retrieval, graph-based
code representations, and agentic localization before editing. In this sense, the repository
becomes the operational substrate on which code assistants plan, act, and
receive feedback.

\paragraph{Executable Development Harnesses.}
Executable development harnesses are becoming the runtime and control plane of
code assistants. Rather than exposing the model to a flat list of tools, recent
systems wrap it in a managed development loop that controls repository access,
file edits, command execution, approval boundaries, context isolation, logging,
and validation. This trend is visible in production systems: Claude Code
packages local terminal/IDE/browser coding into a tool-mediated loop with
editing, command execution, permissions, hooks, memory, and subagents; Codex and
GitHub Copilot coding agents move similar loops into managed cloud or
GitHub-native workspaces with sandboxes, branches, approvals, and auditable
pull-request outputs; and DeepAgents exposes planning, filesystem-backed state,
context management, code execution, and subagent delegation as reusable harness
components~\citep{claudecode2025,codex2025,deepagents2025,copilotagent2025}.
Such loops are increasingly mediated by open protocols such as the Model Context Protocol~\citep{anthropic2024mcp,hou2025model}, 
which standardize how harnesses expose tools, context, and resources to the model and enable cross-system tool reuse. 
In parallel, recent research treats the harness itself as an object of optimization rather than a fixed wrapper: AutoHarness~\citep{lou2026autoharness} synthesizes
code harnesses from environment feedback, Meta-Harness~\citep{lee2026metaharness} searches over harness
code using prior candidates and execution traces, Agentic Harness Engineering~\citep{lin2026agentic}
evolves coding-agent harnesses through observability, and Natural-Language
Agent Harnesses~\citep{pan2026natural} externalize roles, contracts, adapters, and state conventions
into editable harness specifications. Together, these developments suggest that practical progress in code assistants
is increasingly shaped not only by improvements in the base model, but also by
the surrounding execution runtime, including its sandbox, permissions, context
plumbing, telemetry, and verification hooks.

\paragraph{Execution Feedback as Grounded Verification}
A distinguishing property of code assistants is the availability of machine-checkable feedback: compiler diagnostics, test outcomes, linter warnings, and runtime traces.
Agentless~\cite{xia2024agentless} shows that a fault-localization and patch-generation pipeline guided by test execution achieves competitive results on SWE-bench~\cite{jimenez2024swebench} without elaborate agentic control.
RepairAgent~\cite{bouzenia2025repairagent} and Live-SWE-agent~\cite{xia2025live} extend this loop into autonomous program repair driven by test results, while AlphaCodium~\cite{ridnik2024alphacodium} demonstrates that test-driven flow engineering substantially improves competitive programming performance over single-shot prompting.
Execution thus converts each candidate edit from a textual hypothesis into a verifiable transformation of the program world.

\paragraph{Memory and Context Management at Repository Scale}
Repositories routinely exceed any plausible context window, forcing code assistants to maintain explicit, structured memory.
Retrieval-augmented completion~\cite{zhang2023repocoder}, graph-based code indexing~\cite{liu2024codexgraphbridginglargelanguage}, documentation-oriented agents such as RepoAgent~\cite{luo-etal-2024-repoagent}, and recent context-retrieval benchmarks such as ContextBench~\cite{li2026contextbench} instantiate the memory abstractions of \S\ref{sec:memory} with a code-specific twist: stored items such as functions, tests, traces, and retrieved issue contexts are themselves executable or directly tied to executable states, and can be re-run, checked, or localized rather than merely re-read.
Recent memory systems further extend this view by storing reusable agent procedures or repository experience as procedural and experiential memory~\cite{gaurav2025codemem,wang2026memgovern}.
This narrows the gap between memory and environment found in conventional agent architectures, and makes abstraction management particularly acute, since the assistant must select the right scale of code and experience to surface for a given subtask.

\paragraph{Developer Intent and Project Conventions as Latent State}
Beyond explicit repository state, practical coding assistants must reason about
latent developer intent and project conventions. A useful patch should not only
pass visible tests, but also align with the repository's architecture, coding
style, and internal API reuse, properties that recent work describes as the
\emph{organicity} of generated code~\cite{li2026learning}. Agents that ignore
these constraints can produce technically correct patches that maintainers still
reject~\cite{li2026learning,thillen2026codetaste}, while benchmark analyses show
that some seemingly solved SWE-bench issues rely on solution leakage in the
issue text rather than genuine intent inference~\cite{aleithan2024swe}. Coding
assistance is therefore a partially observable program world problem: files,
tests, and tool outputs provide observable state, while design rationales,
implicit constraints, and team conventions must be inferred from issue threads,
prior commits, code reviews, and accumulated interaction history. This extends
the belief state divergence studied in SyncMind from shared multi agent state to
individual agent and user alignment~\cite{Guo2025SyncMind}. Modeling this latent
state is essential for moving from functional code generation toward trustworthy
developer collaboration.

\paragraph{From Inline Completion to Autonomous SWE Agents}
The evolution of code assistants can be viewed as an expansion of the development harness around the model.
Early systems such as Codex-based completion~\cite{chen2021evaluating} and commercial assistants such as Copilot~\cite{peng2023copilot} rely on a lightweight IDE harness, where local context is surfaced, an inline suggestion is generated, and the developer remains the primary executor, verifier, and state manager.
Productivity~\cite{peng2023copilot} and usability~\cite{vaithilingam2022expectation,mozannar2022reading} studies show that even this lightweight harness matters, since the value of a suggestion depends on its alignment with the developer's evolving program state and intent.
At the autonomous end, systems such as SWE-agent, OpenHands, AutoCodeRover, and Agentless operate within a repository-level harness, shifting from isolated code generation to stateful inspection, editing, execution, and revision.

\paragraph{From Patch Generation to Software Lifecycle Participation}
Code assistants are also moving from isolated patch generation toward broader
software lifecycle participation. SWE-bench framed repository-level assistance
as an issue-to-patch task~\citep{jimenez2023swe}, while newer benchmarks such as
SWE-Lancer~\citep{miserendino2025swe} and SWE-Bench Pro~\citep{deng2025swe} evaluate longer-horizon, economically meaningful
software deliverables that span multiple files and require professional
engineering effort. Related benchmarks
such as Terminal-Bench~\citep{merrill2026terminal} and AppWorld~\citep{trivedi2024appworld} further reflect the same shift toward
interactive environments where agents must operate through commands, tools, and
executable application states~\citep{xie2024osworld,yao2025taubench}. In deployment, this trend appears as agents that work
inside persistent engineering workflows rather than static repository snapshots,
including pull-request review, CI/CD feedback, and production issue resolution
~\citep{tang2024codeagent,Baqar_2025}. At production scale,
LingmaAgent reports that an autonomously deployed issue-resolution agent at
Alibaba Cloud resolves 16.9\% of in-house issues fully autonomously and 43.3\%
with manual intervention~\citep{ma2025alibaba,li2026advances}. This suggests
that code assistants are becoming workflow participants, not merely patch
generators.

\paragraph{Multi-Agent Code Assistance and Shared Repositories}
At the upper end of the spectrum, code assistance increasingly takes a multi-agent form, with planner, coder, tester, and reviewer roles operating over a shared repository.
ChatDev~\cite{Qian2023ChatDev}, MetaGPT~\cite{Hong2023MetaGPT},
CodeAgent~\cite{zhang2024codeagent}, and METAL~\cite{li2025metal} show how role specialization combined with a shared executable artifact enables coordination patterns that single agents struggle to sustain over long horizons.
The repository, together with its tests and execution traces, becomes both the medium of communication and the convergence target, directly instantiating the shared program world of \S\ref{sec:mas}.
Concurrent edits, however, can silently invalidate assumptions held by other agents, exposing the world-state synchronization challenges discussed in the same section.

\paragraph{The Harness as a Distillation Surface}
A defining 2026 development is that production harnesses are no longer only deployment infrastructure; they are becoming a dominant source of training data for the next generation of code-assistant models.
Cursor's Composer is trained with continuous online reinforcement learning on real Cursor usage traces, tightening the loop between deployed agent behavior and model updates~\citep{cursor2025composer,cursor2025rtrl}.
OpenAI's codex-1 (an o3 derivative)~\citep{codex2025}, GPT-5-Codex~\citep{openai2025gpt5codexcard}, and GPT-5.1-Codex-Max~\citep{openai2026codexmax} are explicitly trained on long-horizon, multi-turn coding interactions that mirror the Codex harness loop, while Anthropic's internal Claude Code dogfooding contributes a similar feedback channel documented in their teams-using-Claude-Code whitepaper~\citep{anthropic2025teams}.
At the same time, the harness itself is becoming an explicit optimization object: AutoHarness~\citep{lou2026autoharness} synthesizes harness code with a smaller LLM that filters illegal actions, Agentic Harness Engineering~\citep{lin2026agentic} closes an observability-driven evolution loop over harness components, Meta-Harness~\citep{lee2026metaharness} formalizes joint model--harness optimization, and Live-SWE-agent~\citep{xia2025live} edits its own scaffolding at runtime---together suggesting that the boundary between ``the agent'' and ``the harness around the agent'' is becoming a learnable surface in its own right.

\paragraph{Open Challenges for Code-Assistant Harnesses}
The maturation of production harnesses surfaces several coding-specific open problems that complement the cross-domain agenda discussed in the next subsection.
First, verification beyond unit tests remains largely unsolved: the oracle-adequacy crisis exposed by PatchDiff~\citep{wang2025solved} and SWE-Bench++~\citep{anonymous2025swebenchpp}, the security-correctness gap addressed by Aardvark~\citep{openai2025aardvark} and Codex Security~\citep{openai2026codexsecurity}, and the organicity gap between functional and accepted patches~\citep{li2026learning,thillen2026codetaste} all point to a verifier surface that current harnesses underspecify.
Second, failure attribution in long-horizon agent loops is still immature: empirical studies such as ``Why do multi-agent systems fail?''~\citep{cemri2025whymas}, the Who\&When attribution dataset~\citep{zhang2025whoandwhen}, AgenTracer~\citep{agentracer2025}, and AgentDebug~\citep{zhu2025llm} report best step-level attribution accuracies in the 14--53\% range, suggesting that production harnesses lack the structured traces needed for principled debugging.
Third, safety governance of autonomous code execution requires capability-based primitives that remain rare in practice: Aethelgard's learned capability governor~\citep{anonymous2026aethelgard}, fault-tolerant transactional sandboxing~\citep{anonymous2025faultsandbox}, and Microsoft's Agent Governance Toolkit~\citep{microsoft2026governance} represent early steps toward enforcing least privilege under concurrent agent action.
Fourth, harness self-evolution at production scale---demonstrated only in narrow settings by AutoHarness, AHE, and Live-SWE-agent---raises stability and rollback questions absent from non-self-modifying harnesses.
Fifth, multi-agent state synchronization on live repositories generalizes the SyncMind belief-state divergence problem~\citep{Guo2025SyncMind} to settings where humans, autonomous agents, and CI systems concurrently mutate shared program state.
Finally, trust calibration in pair programming user experience remains an under studied human factors problem, including decisions about when to interrupt, when to checkpoint, when to delegate, and when to defer, despite its centrality to whether harness driven autonomy can be safely scaled to enterprise workflows.

Code assistants are thus the clearest production instantiation of code-centric agentic systems and the most demanding testbed for the harness-engineering discipline now emerging across industry and academia.

\subsubsection{GUI/OS Agents as a Program World}

Graphical user interfaces and operating systems constitute, perhaps more than any other tangible application of foundation-model agents, a \textit{program world} in the most literal sense: every observation an agent receives is the rendered output of executable code (HTML, CSS, layout XML, accessibility APIs, framebuffers driven by window managers), and every action it takes is a call into another piece of code (a DOM event, an \texttt{adb} shell command, a keystroke captured by the OS event loop, a Playwright script). For this reason, GUI/OS agents have become the canonical testbed for the central thesis that code is the unifying substrate through which perception, action, environment dynamics, and memory can be represented, executed, and verified. Below we develop this view systematically.

\paragraph{GUI/OS as a Partially Observable Program World}

We model a GUI/OS environment as a Partially Observable Markov Decision Process $\langle \mathcal{S}, \mathcal{A}, \mathcal{O}, T, R\rangle$ in which the latent state \textit{s} $\in \mathcal{S}$ is the full program state of one or more processes (a browser's full DOM and JavaScript heap, an Android emulator's Activity stack and content providers, a Linux VM's filesystem and window tree). The agent never observes \textit{s} directly; it observes \textit{o} $\in \mathcal{O}$, which in modern systems takes one of four code-defined forms: (i) a serialized DOM or HTML subtree as in WebArena and Mind2Web \citep{zhou2024webarenarealisticwebenvironment,deng2023mind2webgeneralistagentweb}; (ii) an accessibility tree (AXTree) exposed by Android's UIAutomator or by macOS/Windows accessibility APIs as in AndroidWorld and WindowsAgentArena, for example, adopted by AgentOccam \citep{rawles2025androidworlddynamicbenchmarkingenvironment,bonatti2024windowsagentarenaevaluating,yang2024agentoccam}; (iii) a screenshot annotated with bounding-box or Set-of-Mark coordinates, the representation adopted by SeeAct, WebVoyager, OSWorld, and most recent native models \citep{zheng2024gpt4visiongeneralistwebagent,he2024webvoyagerbuildingendtoendweb,xie2024osworldbenchmarkingmultimodalagents,yang2023setofmarkpromptingunleashesextraordinary}; or (iv) hybrid representations that interleave pixels, accessibility metadata and HTML, as in WebArena's BrowserGym observation space and in CogAgent's dual-resolution encoder \citep{drouin2024workarenacapablewebagents,hong2024cogagentvisuallanguagemodel}. The action space $\mathcal{A}$ is likewise code: a tuple $\langle action\_type, target, value\rangle$ that compiles either to a DOM/accessibility call (\texttt{element.click()}, \texttt{setText(node\_id, ``...'')}) or to OS-level keyboard/mouse primitives (\texttt{pyautogui.click(x,y)}, \texttt{xdotool key}). Crucially, the transition function $T$ is not learned but \textit{executed}: the browser engine, the Android runtime, or the host OS deterministically produces the next observation. Agents are commonly framed as human-like computer users: they perceive the visual interface, reason over the user instruction, and execute actions through the same graphical channel available to humans. The agent's policy $\pi(a|h)$ is therefore best thought of as a \textit{program synthesizer} that, conditioned on a history \textit{h}, emits the next snippet of executable code; the environment is the interpreter.

\paragraph{Code as a Bridge Between User Interfaces and GUI Agents} 
Recent works treat code as an intermediate interface between high-level model reasoning and low-level UI execution \citep{xie2024osworldbenchmarkingmultimodalagents, wang2025guiagentsfoundationmodels,
xu2024androidlabtrainingsystematicbenchmarking}. This interface provides two main advantages: First, it abstracts away noisy visual details, and creates a natural boundary between the model's semantic planning and the system's executable control layer. Second, it fuses the perception, action, and evaluation in to a single code-as-harness pipeline.

On the action side, this is the GUI specialization of the broader CodeAct paradigm \citep{wang2024executablecodeactionselicit}: rather than emitting JSON tool calls, agents emit Python or JavaScript snippets that compose primitives such as \texttt{click(x, y)}, \texttt{type(text)}, \texttt{scroll(dx, dy)}, \texttt{key(``Enter'')}, and arbitrary library calls (e.g., \texttt{requests}, \texttt{subprocess}, \texttt{selenium}). Cradle makes this explicit by having an LMM output executable Python that drives keyboard and mouse for any application, including AAA games, achieving generalization across previously unseen software through skill curation and self-reflection rather than task-specific APIs \citep{tan2024cradleempoweringfoundationagents}. WebArena, BrowserGym, and TheAgentCompany similarly expose Playwright-style code actions whose execution is the ground truth of progress \citep{zhou2024webarenarealisticwebenvironment,drouin2024workarenacapablewebagents,xu2025theagentcompanybenchmarkingllmagents}.

On the perception side, recent native GUI models such as SeeClick, CogAgent, Ferret-UI, OS-Atlas, ShowUI, Aria-UI, UGround, UI-TARS, and GUI-Libra treat grounding as a \textit{function from pixels to executable coordinates}, training large vision-language models to emit $(x, y)$ or \texttt{bbox} tokens that can be directly piped into an action API \citep{cheng2024seeclickharnessingguigrounding,hong2024cogagentvisuallanguagemodel,you2024ferretuigroundedmobileui,wu2024osatlasfoundationactionmodel,lin2024showuivisionlanguageactionmodelgui,yang2025ariauivisualgroundinggui,gou2025navigatingdigitalworldhumans,qin2025uitarspioneeringautomatedgui,yang2026guilibratrainingnativegui}. By collapsing the planner→grounder→executor pipeline into a single VLA model whose output token stream is itself runnable code, these systems eliminate the brittle string-matching layer that historically separated language plans from grounded actions, as documented in SeeAct's analysis showing that grounding, rather than planning, is the dominant bottleneck on Mind2Web \citep{zheng2024gpt4visiongeneralistwebagent}.

On the evaluation side, code-defined environments enable \textit{executable feedback}: success is determined not by a learned reward model but by running an evaluator script over the post-action system state. WebArena's URL/string assertions, OSWorld's per-task Python checkers operating over OS file I/O and application state, AndroidWorld's \texttt{adb}-based state inspection, and Spider2-V's enterprise-tool checks all share the same pattern, an evaluator is itself a piece of code that interrogates the program world after the agent has finished \citep{zhou2024webarenarealisticwebenvironment,xie2024osworldbenchmarkingmultimodalagents,rawles2025androidworlddynamicbenchmarkingenvironment,cao2024spider2vfarmultimodalagents}. This closes the loop: code generates the environment, code is the agent's action, and code adjudicates the result.

\paragraph{Memory as Persistent Programmatic State}
For code-grounded GUI agents, memory is best understood as a \textit{persistent programmatic state layer}: structured artifacts that outlive the current UI state and can be retrieved, composed, or executed in later interactions. Recent works explore different line of memory: (i) \textit{Working memory of UI state} compresses the current observation to a task-relevant abstraction: Synapse's state-abstraction module filters HTML to a few task-relevant elements, allowing trajectory-as-exemplar prompting and an exemplar memory that retrieves prior trajectories by similarity \citep{zheng2024synapsetrajectoryasexemplarpromptingmemory}. (ii) \textit{Long-term cross-app/session memory} is implemented as structured documents and skill libraries: AppAgent compiles an exploration document per application that records the learned function of each UI element, which is then consulted on subsequent tasks \citep{zhang2023appagentmultimodalagentssmartphone}; Mobile-Agent-v2 introduces a dedicated planning agent whose memory tracks long-horizon progress across sub-tasks \citep{wang2024mobileagentv2mobiledeviceoperation}; Cradle maintains an explicit skill-curation module that promotes successful code snippets to a reusable library \citep{tan2024cradleempoweringfoundationagents}. Whereas these designs are tightly coupled to the host application's UI ontology, PlugMem proposes a \textit{task-agnostic} plugin memory module that distils raw interaction traces into a compact knowledge-centric memory graph of propositional and prescriptive knowledge, transferring unchanged from web agents to long-horizon dialogue and multi-hop retrieval \citep{yang2026plugmemtaskagnosticpluginmemory}. (iii) \textit{Self-evolving GUI agents} (already cited in this survey as UI-Voyager \citep{lin2026uivoyagerselfevolvingguiagent}) and AutoGLM extend this idea with online curriculum reinforcement learning that continuously grows a library of grounded behaviors, while OS-Genesis and UI-TARS use reflective trace collection on hundreds of virtual machines as a form of distilled memory \citep{liu2024autoglmautonomousfoundationagents,sun2025osgenesisautomatingguiagent,qin2025uitarspioneeringautomatedgui}. In all three regimes the memory is itself a code artifact, for example, a JSON document, a Python skill module, or a vector index of code-formatted trajectories, directly executable or directly composable into the agent's next action.

\paragraph{UI Simulators and Sandboxes as Executable Dynamics}

The simulator stack for GUI/OS agents is perhaps the clearest demonstration that environment dynamics in this domain \textit{is} code. Early benchmarks such as MiniWoB++ defined each task as a self-contained HTML/JavaScript page with a programmatic reward function \citep{liu2018reinforcementlearningwebinterfaces}; WebShop scaled this to 1.18M real Amazon products inside a self-hosted shopping site \citep{yao2023webshopscalablerealworldweb}. Mind2Web cached real-world traces for offline evaluation, while WebArena and VisualWebArena fork four full-stack open-source sites into Docker containers with deterministic resets and per-task functional checkers \citep{deng2023mind2webgeneralistagentweb,zhou2024webarenarealisticwebenvironment,koh2024visualwebarenaevaluatingmultimodalagents}. OSWorld pushes this further to 369 real Ubuntu/Windows/macOS tasks in disposable VMs whose initial state, golden actions, and Python evaluation scripts are all version-controlled artifacts \citep{xie2024osworldbenchmarkingmultimodalagents}; WindowsAgentArena specializes the same architecture for Windows 11 with Azure-parallel execution \citep{bonatti2024windowsagentarenaevaluating}; and Spider2-V extends OSWorld to professional data-engineering pipelines spanning BigQuery, dbt, and Airbyte \citep{cao2024spider2vfarmultimodalagents}. On mobile, AndroidWorld provides 116 programmatic tasks dynamically parameterized from natural-language templates with reward signals derived from device system state, while AndroidArena and AndroidLab supply complementary cross-app evaluations \citep{rawles2025androidworlddynamicbenchmarkingenvironment,xing2024understandingweaknesslargelanguage,xu2024androidlabtrainingsystematicbenchmarking}. BrowserGym and WorkArena unify many of these under a common Gym-style API and add 23,150 enterprise ServiceNow task instances \citep{drouin2024workarenacapablewebagents}, while AgentBench's OS and web tracks and the OpenHands-driven TheAgentCompany benchmark situate GUI control inside broader knowledge-work simulations \citep{liu2025agentbenchevaluatingllmsagents,xu2025theagentcompanybenchmarkingllmagents}. Most recently, Code2World makes the program-world stance explicit at the model level by training a vision-language coder that predicts the next GUI state as renderable HTML, turning the world model itself into an executable artifact and using rendered outcomes as reinforcement signals \citep{zheng2026code2worldguiworldmodel}. Together, these sandboxes embody the survey's claim that environment dynamics in agentic systems are increasingly authored as code: they are forkable, diffable, version-controlled, and reproducible in ways that no learned simulator can match.

\paragraph{From Simulation to Production: Executable Feedback Loops}

The same code-as-harness interface that makes simulators tractable has enabled an unusually rapid jump to production deployment, because the agent's input/output contract: screenshots in, code (or coordinate-typed function calls) out, is identical in both settings. Anthropic's Claude Computer Use exposes a public-beta API in which the model takes screenshots of a sandboxed desktop and emits keyboard/mouse actions as structured tool calls \citep{anthropic2024computeruse}. OpenAI's Operator and the underlying Computer-Using Agent (CUA) followed, combining GPT-4o's vision with reinforcement-learned reasoning over a unified click/scroll/type action space \citep{openai2025operator}. Google DeepMind's Project Mariner ships a Gemini-powered Chrome extension that observes the rendered DOM, plans, and executes browser actions on behalf of the user, and is being integrated into Search's AI Mode and the Gemini app \citep{deepmind2025mariner}. ByteDance's UI-TARS-1.5/2 and the associated UI-TARS-desktop product, Zhipu's AutoGLM (web browser plug-in and Android app), and Tencent's AppAgent lineage demonstrate that the same architecture transfers from the lab to consumer devices \citep{qin2025uitarspioneeringautomatedgui,liu2024autoglmautonomousfoundationagents,zhang2023appagentmultimodalagentssmartphone}. AutoWebGLM, the production sibling of CogAgent, exemplifies the route from arXiv preprint to deployed browser agent through an ``intermediate interface'' that decouples planning from grounding \citep{lai2024autowebglmlargelanguagemodelbased}. Earlier industrial efforts, like Adept's ACT-1/ACT-2 and Rabbit's Large Action Model, anticipated this trajectory but predated the executable-feedback infrastructure that has since made the loop reliable enough for deployment.

Looking forward, the literature converges on three frontiers, all expressed in code-as-harness terms. \textit{First}, native end-to-end agents that internalize perception, planning, grounding, and action into a single VLA model are displacing the modular planner+grounder pipeline. \textit{Second}, executable world models promise to give agents human-like foresight by predicting the next UI state as renderable code rather than as pixels or unstructured text. \textit{Third}, embodied, instruction-following GUI agents treat the entire device (e.g., terminal, browser, native apps, and peripherals) as a unified program world. The common thread is that code is the lingua franca: it defines observations, actions, evaluation, memory, and increasingly the world model itself.

\subsubsection{Autonomous Embodied Agents}

Embodied agent operates in the physical world or its simulation, perceiving the environment through structured outputs from vision and force sensors, and acting through motor commands subject to physical constraints such as reachability, collision, and dynamics.

\paragraph{Code as the Control Boundary that Connecting Agents and the World}
Unlike purely reasoning agents, embodied agents operate under physical constraints that may fail silently when violated: a robot may attempt to grasp an object outside its workspace without producing any explicit failure signal~\citep{liang2023codepolicieslanguagemodel}. This shifts the burden of correctness from runtime to action-generation time, where the agent's output must already be expressive enough to compose verified operation intents before reaching the actuator. 
Code naturally satisfies the requirements by serving both as the grounding interface and as the safety boundary. As a grounding interface, it translates high-level intent from LLMs into embodiment-respecting commands through primitive skill calls \citep{ahn2022can, ren2023robots, zhai2026skillvla, zhang2023bootstrap}, synthesized Python control policies \citep{liang2023code, mu2024robocodex, xie2025robotic, wang2025llm, ji2026genswarm}, and structured behavior-tree programs \citep{zhang2025codebt}. As a safety boundary, it constrains admissible actions at execution time \citep{guan2025normcode, szeider2025cp, miculicich2025veriguard}.

\paragraph{Layered Harness for Grounded and Verifiable Embodied Actions} 
Embodied agents require a layered harness that separates semantic reasoning from executable, physically grounded, and human-governed control~\citep{vemprala2024chatgpt}. Foundation models handle the semantic layer of embodied agency: interpreting goals, decomposing tasks, inferring affordances, selecting skills, proposing actions, and replanning under changing observations~\citep{huang2022inner, wang2023voyager}.  Code and classical robotics software define the admissibility boundary by exposing typed robot APIs, parameterizing primitive skills, calling geometric libraries, invoking motion planners, and supporting inspection, replay, versioning, and verification \citep{xie2025robotic, liang2023code, huang2023voxposer, macenski2020nav2}. Perception models and state estimators convert raw sensor streams into structured state that planners and controllers can use ~\citep{driess2023palme, deepmind2025geminirobotics}. Physical systems and low-level controllers then enforce embodiment-specific constraints such as kinematics, dynamics, collision avoidance, workspace limits, contact forces, timing, and stability. 

\paragraph{Reusable Skills as Embodied Memory} While code grounds a single action in physical feasibility, embodied agents operating over long horizons must also accumulate experience across tasks. In this regime, code takes on a second role: the same executable form that makes an action verifiable also makes it storable and reusable. 
Memory therefore naturally takes the form of a skill library, a collection of code artifacts that record past behavior and can be called as actions in future tasks.
This dual identity distinguishes embodied memory from other memory abstractions in \S\ref{sec:memory}: a skill is not merely something the agent reads, but something the agent re-executes. Voyager pioneered this paradigm with an growing skill library for open-ended tasks in Minecraft \citep{wang2023voyager}, and other work extends the same idea along several directions: tabletop manipulation \citep{tziafas2024lifelong}, human correction \citep{meng2025growing}, vision-grounded replanning \citep{kagaya2025vireskill}, and continual learning \citep{wang2026lifelong}. The principle has even crossed into the GUI domain \citep{lin2026uivoyagerselfevolvingguiagent}. Across these systems, the challenge has shifted from generating skills to governing the library: handling forgetting, abstraction, and grounding alignment.

\paragraph{Coordinated and Auditable Real-World Deployment} Moving from simulation to real-world deployment introduces challenges that go beyond a single agent: multiple robots must coordinate, behaviors must be auditable, and skills must transfer across embodiments. Code naturally extends to address all three. For coordination, it provides the substrate for multi-robot policy synthesis \citep{ji2026genswarm} and robot-agnostic cooperative architectures \citep{ashley2026racas}. For auditability, it supports governance mechanisms for industrial safety \citep{guan2025normcode, liu2026agents4plc} and verified closed-loop control \citep{santos2026alrm}. For cross-embodiment transfer, the same code-based skill abstraction enables combinatorial reuse on dual-arm systems \citep{zhai2026skillvla}. Open challenges remain in reducing the sim-to-real gap, scaling multi-agent coordination, and maintaining safety as environments evolve.

\subsubsection{Agents for Scientific Discovery as Program Worlds}

Scientific research is among the most natural testbeds for code as an agent harness: the scientific method is itself a closed loop of \textit{hypothesize → design → execute → observe → revise}, in which each transition is mediated by an artifact that is, increasingly, a program. Modern science can already be digital end-to-end, for example, hypotheses are encoded as differential equations or generative models, experimental protocols are written as XDL or Opentrons scripts, instruments are driven through Python APIs, and analyses live in Jupyter notebooks whose cells form a verifiable trace of reasoning. This makes scientific discovery an ideal domain to instantiate the three-fold role of code: code as the medium of \textit{reasoning} (e.g., symbolic derivations, formal proofs, hypothesis-as-program), code as the substrate of \textit{acting} (e.g., calls to wet-lab robots, simulators, statistical pipelines), and code as the executable \textit{environment} itself (e.g., molecular-dynamics engines, autonomous laboratories, virtual research teams). Recent systems, like \textit{AI Scientist v1/v2}, \citep{lu2024aiscientistfullyautomated,yamada2025aiscientistv2workshoplevelautomated} AI co-scientist \citep{gottweis2025aicoscientist}, Virtual Lab \citep{swanson2025virtual} and Biomni \citep{huang2025biomni}, make this code-as-harness framing concrete by elevating the entire research workflow to a single, executable program graph.

\paragraph{Scientific Discovery as a Partially Observable Program World}

We treat a research project as a partially observable program world $\langle \mathcal{S}, \mathcal{A}, T, \mathcal{O}, R\rangle$. The state $\mathcal{S}$ is a structured program memory containing the current best hypotheses, accumulated literature, code artifacts, intermediate datasets, and experimental observations. Actions $\mathcal{A}$ are typed code expressions: literature-search queries, calls to symbolic or numerical solvers, generation of new experimental scripts, modifications to a training pipeline, or robot-control commands. The transition function $T$ is realized by a Python interpreter, a Lean kernel, a quantum-chemistry package, a robotic synthesizer, or, in fully end-to-end systems such as the AI Scientist v2 \citep{yamada2025aiscientistv2workshoplevelautomated}, by a tree-search experiment manager that orchestrates all of these. Observations $\mathcal{O}$ correspond to execution outputs (numerical results, plots, error messages, peer-review scores), and the latent reward $R$ encodes desiderata such as novelty, reproducibility, and statistical significance. Crucially, the policy of a scientific agent is itself a program: ChemCrow \citep{bran2023chemcrowaugmentinglargelanguagemodels} composes 18 expert-designed chemistry tools through structured tool calls; Coscientist \citep{boiko2023autonomous} interleaves Python execution, web search, and robotic-API actions; and \textit{AlphaProof} \citep{hubert2025olympiad} expresses each ``reasoning step'' as a Lean tactic that the proof assistant verifies before transitioning the state. This view recasts traditionally informal categories (e.g., hypothesis, protocol, claim) as concrete program objects whose execution traces can be logged, replayed, and audited.

\paragraph{Unifying Ideation, Experimentation, Analysis, and Communication}

Traditional accounts of science separate ideation, experiment design, data analysis, and dissemination into distinct workflows with distinct tools. Code-centric agents collapse these into a single executable pipeline. ResearchAgent \citep{baek2025researchagentiterativeresearchidea} and SciAgents-style systems iteratively refine hypotheses by traversing entity graphs over the literature, with each candidate idea materialized as a structured object that can be passed to downstream planners. BioPlanner \citep{odonoghue2023bioplannerautomaticevaluationllms} formalizes wet-lab protocols as pseudocode whose admissible functions can be type-checked, retrieved, and composed, providing the same compositional substrate for biology that XDL provides for chemistry \citep{mehr2020universal}. Agent Laboratory \citep{schmidgall2025agentlaboratoryusingllm} and its preprint-sharing extension AgentRxiv \citep{schmidgall2025agentrxivcollaborativeautonomousresearch} explicitly factor research into three program-level phases: literature review, experimentation, report writing, orchestrated by specialized PhD, postdoc, and engineer agents that exchange Python files, LaTeX, and arXiv records. The AI Scientist \citep{lu2024aiscientistfullyautomated,yamada2025aiscientistv2workshoplevelautomated} goes further by representing an entire ML paper as a single executable trace: the system writes the experimental code with a coding assistant, executes it, reads the figures with a vision-language model, and emits a LaTeX manuscript that includes the very plots it generated. In all of these systems, what used to be a heterogeneous pipeline of natural-language artifacts becomes a homogeneous flow of typed code objects, enabling end-to-end optimization and automatic verification at every stage \citep{ren2026scientificintelligencesurveyllmbased,wang2024executablecodeactionselicit}.

\paragraph{Memory as Persistent Program State}

Long-horizon research depends on memory: prior experiments, failed attempts, citation graphs, and tacit lab know-how. Code-centric agents externalize this memory as persistent program state. At the \textit{working-memory} level, agents maintain executable scratchpads,  typically a Jupyter kernel or a CodeAct-style Python REPL \citep{jiang2025aideaidrivenexplorationspace},  whose live variables, dataframes, and figures form the immediate context for reasoning. El Agente Q \citep{Zou_2025} and Biomni \citep{huang2025biomni} exemplify hierarchical memory: short-lived tool outputs are cached in an episodic buffer, while structured artifacts (plasmid maps, optimized geometries, fitted models) are written to durable file stores that subsequent agent steps can re-load. At the \textit{long-term} level, PaperQA / PaperQA2 \citep{lala2023paperqaretrievalaugmentedgenerativeagent} and Google's AI co-scientist \citep{gottweis2025aicoscientist} treat the scientific literature itself as an indexed knowledge base, accessed through tool calls that retrieve passages, expand citations, and detect contradictions; this enables hypothesis evaluation against millions of prior results without inflating the prompt. AgentRxiv \citep{schmidgall2025agentrxivcollaborativeautonomousresearch} takes the idea one step further by giving autonomous research agents a shared preprint server: hypotheses, code, and findings produced by one run are uploaded as durable program artifacts that future runs can build on, instantiating cumulative scientific progress as a globally shared, version-controlled program state. Biomni's action-discovery agent \citep{huang2025biomni} mines tens of thousands of bioRxiv papers to populate a unified tool registry across 25 biomedical subfields, so that ``remembering how to clone a plasmid'' becomes the concrete act of importing a verified, code-level protocol from persistent storage.

\paragraph{Simulators as Executable Dynamics}

Scientific agents rely on simulators of physical and computational reality, and the code-as-harness view treats these uniformly as executable transition models. In computational chemistry, El Agente Q \citep{Zou_2025} wraps DFT engines, geometry optimizers, and thermochemistry tools as callable functions that the LLM invokes to roll out alternative reaction trajectories; on six university-level benchmarks it exceeds 87\% task success while emitting a transparent action-trace log of every simulation. ChemCrow \citep{bran2023chemcrowaugmentinglargelanguagemodels} similarly integrates RDKit, retrosynthesis engines, and reaction predictors so that an agent can ``execute'' a candidate synthesis virtually before committing to a wet-lab run. In structural and systems biology, the Virtual Lab \citep{swanson2025virtual} composes ESM, AlphaFold-Multimer, and Rosetta into a Python pipeline through which an LLM Principal-Investigator agent and its subordinate scientist agents jointly designed 92 SARS-CoV-2 nanobodies, two of which showed validated binding to JN.1 and KP.3 variants,  all in a few days of simulated meetings. For algorithmic and mathematical science, AlphaProof \citep{hubert2025olympiad} uses the Lean theorem prover as the executable environment, formally verifying every candidate proof step before reinforcing the language model, and AlphaEvolve \citep{novikov2025alphaevolvecodingagentscientific} orchestrates an evolutionary loop in which Gemini-generated code edits are executed and scored by automated evaluators, yielding new matrix-multiplication algorithms and mathematical constructions. In each case the simulator is the world: program states evolve only through verified executions, eliminating much of the hallucination that plagues purely textual scientific reasoning \citep{ren2026scientificintelligencesurveyllmbased}.

\paragraph{From Simulation to Production: Self-Driving Labs as Executable Feedback Loops}

The decisive test of a scientific agent is whether its closed loop crosses the boundary into physical reality. Self-driving laboratories (SDLs) are the production systems of this domain: they expose real instruments, like liquid handlers, XRD scanners, spectrometers, robotic arms, through code APIs, and accept agent-generated programs as their primary input. Berkeley's A-Lab \citep{szymanski2023autonomous} combines machine-learned synthesis recipes with autonomous robotics to synthesize 41 novel inorganic compounds from a target list of 58 in 17 days of continuous operation, while early thin-film SDLs \citep{macleod2020selfdrivinglaboratoryaccelerateddiscovery} established that Bayesian optimization loops can be wrapped as Python services and run unattended. Coscientist \citep{boiko2023autonomous} crossed this threshold for organic chemistry by autonomously planning, executing, and analyzing palladium-catalyzed Suzuki and Sonogashira couplings on the Emerald Cloud Lab and an in-house liquid-handling platform from a single English prompt. The Cronin group's Chemputer and its XDL chemical-description language \citep{mehr2020universal} formalize this contract: any synthesis published in the literature can be parsed into hardware-independent XDL code that compiles, like LLVM IR for chemistry, onto any compliant robotic platform. In biology, Biomni \citep{huang2025biomni} generates end-to-end molecular-cloning protocols that human reviewers rated comparable to a senior Stanford postdoc, while Google's AI co-scientist's drug-repurposing and antimicrobial-resistance hypotheses were experimentally validated in collaborator wet labs at Imperial College and Stanford \citep{gottweis2025aicoscientist}. MatPilot \citep{ni2024matpilotllmenabledaimaterials} explicitly couples a hypothesis-generation cognition module to an autonomous experimental-verification module driving physical synthesis robots, instantiating a complete generate–execute–feedback loop for materials. These systems make the survey's central thesis tangible: in a self-driving lab, the agent's policy \textit{is} the code, the lab \textit{is} the runtime, and the publication record \textit{is} the log.

\paragraph{Toward Agentic and Instruction-Following Science}

A final dimension of code-as-harness scientific agents is controllability: the ability to steer them with high-level scientific intent while preserving rigorous execution semantics. Benchmarks have rapidly emerged to measure this capability. MLAgentBench \citep{huang2024mlagentbenchevaluatinglanguageagents} evaluates language agents on 13 open-ended ML research tasks, requiring agents to read code, run experiments, and improve metrics. MLE-bench \citep{chan2025mlebenchevaluatingmachinelearning} scales this to 75 Kaggle ML-engineering competitions; the best-performing scaffold at release (OpenAI o1-preview with the Weco AIDE tree-search agent \citep{hu2025surveyscientificlargelanguage}) reaches Kaggle bronze-medal level on 16.9\% of competitions, and AIDE achieves roughly three times the medal rate of the next agent. ScienceAgentBench \citep{chen2025scienceagentbenchrigorousassessmentlanguage} compiles 102 tasks adapted from peer-reviewed publications across bioinformatics, computational chemistry, GIS, and cognitive neuroscience, \textit{unifying every target output as a self-contained Python program}, which is an explicit endorsement of code as the universal interface to data-driven science. DiscoveryBench \citep{majumder2024discoverybenchdatadrivendiscoverylarge} complements this with 264 multi-step hypothesis-search tasks across six domains, exposing failure modes of current agents (best system score $\sim$25\%). On the controllability side, instruction-following progress is visible in systems such as the AI co-scientist \citep{gottweis2025aicoscientist}, where scientists steer the multi-agent debate via natural-language research goals and constraints, in Biomni \citep{huang2025biomni}, whose graphical interface accepts natural-language queries and returns auditable code execution, and in the Virtual Lab \citep{swanson2025virtual}, where a human PI specifies high-level objectives and the AI PI dynamically configures a team of expertise-specific agents. AlphaEvolve \citep{novikov2025alphaevolvecodingagentscientific} and AlphaProof \citep{hubert2025olympiad} represent the goal-conditioned extreme: the agent is given only an objective function or a theorem statement, and the closed code-execution loop searches for any program that satisfies the verifier. Across these systems, instruction-following is realized by translating user goals into typed program specifications that the runtime can rigorously enforce.

Taken together, recent work on agents for scientific discovery exemplifies the survey's central shift: from static prediction toward interactive, stateful, and executable decision making. Hypotheses cease to be free-floating sentences and become parameterized programs; experiments cease to be lab notebooks and become version-controlled code; analyses cease to be one-off scripts and become reproducible artifacts that downstream agents can re-execute; and laboratories cease to be opaque physical sites and become production runtimes addressable through documented APIs. The result is a closed generate–execute–feedback loop in which a single substrate, code, carries scientific reasoning, scientific action, and the scientific environment itself, providing a unified foundation on which agents like the AI Scientist \citep{lu2024aiscientistfullyautomated,yamada2025aiscientistv2workshoplevelautomated}, AI co-scientist \citep{gottweis2025aicoscientist}, Virtual Lab \citep{swanson2025virtual}, Biomni \citep{huang2025biomni}, Coscientist \citep{boiko2023autonomous}, and AlphaEvolve \citep{novikov2025alphaevolvecodingagentscientific} can be compared, composed, and progressively improved. As benchmarks such as MLAgentBench \citep{huang2024mlagentbenchevaluatinglanguageagents}, MLE-bench \citep{chan2025mlebenchevaluatingmachinelearning}, ScienceAgentBench \citep{chen2025scienceagentbenchrigorousassessmentlanguage}, and DiscoveryBench \citep{majumder2024discoverybenchdatadrivendiscoverylarge} make precise, the open challenge is not whether code-as-harness agents can imitate isolated scientific tasks, but whether they can be trusted to drive the full loop autonomously, which is a challenge for which the program-world abstraction provides both the right ontology and the right experimental harness.

\subsubsection{Agent Personalization}

Personalization and recommender systems offer a distinctive setting for code-centric agentic systems. Unlike coding, GUI control, or scientific discovery, the environment here is not only a software system but also a human user whose intent, satisfaction, and long-term goals are only partially observed. As recommendation moves from static ranking toward interactive agents, the central challenge becomes how to maintain, update, and govern a user model through repeated interaction. Code is useful in this setting not simply because it executes recommendation policies, but because it provides an inspectable substrate for preference representation, feedback processing, constraint enforcement, and policy adaptation.

\paragraph{From Static Recommendation to Interactive Personalization}
Traditional recommender systems usually treat personalization as a prediction problem: given historical interactions, the system scores candidate items and returns a ranked list~\citep{he2020lightgcn,guo2017deepfm}. LLM-based recommenders broaden this view by enabling conversational preference elicitation, explanation, and multi-step refinement. Early prompting-based approaches query an LLM with user history and ask it to produce recommendations directly \citep{hou2024large, dai2023uncovering}. More agentic systems instead decompose recommendation into candidate retrieval, filtering, re-ranking, explanation, and feedback collection.
The emerging agentic recommendation~\citep{liu2025recoworld,wang2024recmind,huang2025recommender} instantiate this direction by using LLMs to coordinate recommendation sub-tasks through tool calls and structured intermediate states. Agent4Rec \citep{zhang2024generative} and iAgent~\cite{xu2025iagent} further simulates recommendation sessions with synthetic users, enabling offline evaluation of interactive policies. These systems mark a shift from recommendation as one-shot scoring to an adaptive process, where each interaction may revise the system's belief about the user.

\paragraph{Preference State as an Editable Artifact}
A key difference between personalization agents and other agentic systems is that the most important state is not fully observable. User preferences are latent, contextual, and often unstable. A user may click an item for convenience rather than genuine interest, skip an item because of timing rather than dislike, or change goals across sessions. Therefore, personalization agents need explicit preference states that can absorb noisy behavioral signals while remaining interpretable and correctable.
Code-centric representations provide a practical way to structure this state. Short-term interests can be stored as recent interaction logs, contextual summaries, or session-level preference vectors. Long-term preferences can be maintained as structured memory objects that record stable interests, constraints, and user-provided corrections. AMem \citep{xu2026mem} and related memory-based systems~\citep{wei2025evo,chhikara2025mem0} show how long-term user information can be maintained as editable documents or structured records. MemRec \citep{chen2026memrec} further studies how collaborative signals can support memory management for personalized recommendation. Compared with opaque embedding-only memory, structured preference memory is easier to inspect, revise, and reuse. A user can correct a stored preference in natural language, and the system can update the corresponding state before generating future recommendations.

\paragraph{Feedback as Policy Adaptation}
Personalization agents are driven by feedback, but the feedback is often sparse, delayed, and ambiguous. Clicks, dwell time, ratings, purchases, skips, and conversational corrections all provide partial evidence about user satisfaction. Production recommender systems already rely on code-defined feedback pipelines that log interactions, compute metrics, run A/B tests, and trigger model or policy updates. In an agentic setting, these pipelines become part of the personalization harness: they determine what signals are recorded, how they are interpreted, and when the agent should adapt.
User simulators~\citep{zhang2025llm,wang2025user,liu2025recoworld} provide an offline way to study such adaptation. They allow recommendation policies to be tested under controlled behavioral assumptions before real deployment. Recent LLM-based simulators extend this idea by generating richer synthetic user profiles and interaction traces. However, the central difficulty remains that simulated feedback may not match real user behavior, especially when recommendations themselves influence future preferences.

\paragraph{Controllable and Instruction-Following Personalization}
A major opportunity for agentic personalization is to move beyond optimizing implicit engagement signals toward following explicit user instructions. Users may want recommendations that satisfy constraints such as avoiding certain sources, limiting repeated categories, balancing exploration and familiarity, or prioritizing long-term goals over short-term engagement. These requirements are hard to express through a single learned score but can be represented as structured constraints, filters, or reward functions.
LLM-based conversational recommenders can elicit such preferences in natural language and translate them into policy specifications \citep{hou2024large}. Constraint-based recommendation further shows how fairness, diversity, and exposure requirements can be enforced at serving time rather than hidden inside model parameters \citep{lei2020conversational}. Explanation-based systems provide another path toward controllability: if a system explains why an item was recommended, the user can correct the rationale, and the corrected explanation can update the preference state. This makes personalization more interactive and auditable, since the user can shape not only outputs but also the logic behind future outputs.

\paragraph{Open Challenges for Personalization Harnesses}
Personalization raises several challenges that are sharper than in other domains. First, preference grounding remains unresolved. Unlike code assistants, which can rely on tests, or GUI agents, which can check interface states, personalization agents lack a reliable oracle for true user satisfaction. Proxy metrics such as clicks and engagement can be misleading or even harmful when optimized too aggressively.
Second, preference memory introduces privacy and governance risks. Long-term user models may contain sensitive behavioral patterns, so the harness must specify what is stored, where it is stored, how it is updated, and how users can inspect or delete it. Third, personalization is inherently multi-stakeholder. A platform may optimize engagement, a creator may seek exposure, and a user may value welfare or autonomy. Reducing these objectives to a single reward function can obscure conflicts of interest.

\subsection{Open Problems}

Code-as-harness systems shift the central challenge of agentic AI from isolated model generation to the reliability of the complete execution loop. Once agents act through tools, memory, code execution, shared state, and environment feedback, failures may arise from weak verifiers, stale context, unsafe tool access, inconsistent multi-agent state, insufficient multimodal grounding, or poorly governed self-improvement. These issues cannot be diagnosed by final task success alone. This section outlines the key open problems that emerge when the harness is treated as a first-class system component, with the goal of building agentic systems that are executable, inspectable, stateful, verifiable, and governed in long-horizon real-world environments.

\subsubsection{Harness-Level Evaluation and Oracle Adequacy}

Evaluation becomes difficult once an LLM is embedded in a code-agent harness. In this setting, performance is no longer determined by the base model alone, but also by the surrounding runtime: which repository files are retrieved, which tools are exposed, how many retries are allowed, whether the agent can execute tests, how failures are summarized, and what verifier decides success. However, most existing evaluations measure end-task success: whether a generated solution passes tests, solves an issue, or completes an interactive task. Such metrics conflate the capabilities of the base model, the quality of the harness, the reliability of tools, the informativeness of feedback, and the difficulty of the environment. This is especially visible in repository-level software engineering, where an agent may pass visible tests while exploiting weak or incomplete test suites; in GUI/OS tasks, where a scripted checker may miss unsafe or undesirable intermediate actions; and in scientific or embodied settings, where successful execution in a simulator may not imply that the result is scientifically valid or physically safe \citep{jimenez2024swebench,deng2025swe,miserendino2025swe,merrill2026terminal,jain2024livecodebench,chen2025scienceagentbenchrigorousassessmentlanguage}.

A key open problem is therefore to define \emph{harness-level metrics} that evaluate the operational substrate itself. These metrics should complement final task accuracy with measurements of execution reliability, feedback quality, context sustainability, safety, coordination, and reproducibility. Useful dimensions include: (i) \emph{trajectory efficiency}, such as number of tool calls, tokens, edits, executions, and wall-clock time; (ii) \emph{verification strength}, such as test coverage, oracle diversity, and rate of false acceptance; (iii) \emph{recovery ability}, such as whether the agent can diagnose and repair failures after invalid actions; (iv) \emph{state consistency}, such as whether memory, repository state, execution traces, and agent beliefs remain synchronized; (v) \emph{safety compliance}, such as whether permissions, sandboxes, and human-approval gates are respected; and (vi) \emph{replayability}, such as whether the full trajectory can be reconstructed and audited from logs and artifacts~\citep{anthropic2026agentevals}. A central bottleneck in this agenda is \emph{oracle adequacy}: whether the evaluator captures the intended task rather than only a narrow executable proxy. The open problem is not merely to build harder benchmarks, but to evaluate the code-agent harness as an executable runtime system.

\subsubsection{Semantic Verification Beyond Executable Feedback}

Oracle adequacy becomes especially challenging because execution feedback, while central to code-centric agents, can create a false sense of correctness: code can be run, traces can be inspected, tests can be checked, and failures can be fed back into revision. However, execution is only as reliable as the oracle attached to it. Unit tests may be incomplete, static analyzers may over-approximate, GUI checkers may miss unacceptable intermediate actions, scientific scripts may encode invalid assumptions, and robot simulators may hide physical risks. As a result, a harness can become overconfident precisely because it has executable feedback: the agent sees a green test, but the green test is not the full specification.

The central missing abstraction is a verification stack with explicit scope. Instead of treating pass/fail as a single terminal signal, future harnesses should compose multiple verification artifacts: unit tests, integration tests, property-based tests, fuzzers, static analyzers, type checkers, security scanners, runtime monitors, coverage reports, formal specifications, model-based critiques, and human review. Each artifact should declare what it verifies, what it cannot verify, and what confidence it provides. This is especially important for self-repair and self-evolving harnesses: if the verifier is weak, the agent will learn to optimize against the wrong signal. A useful direction is to make every accepted action carry an evidence bundle containing the checks run, the assumptions preserved, the untested regions, and the remaining risks. In this view, verification is not a final gate; it is an evolving, inspectable contract between the agent, the harness, and the environment.

Other promising directions include feedback calibration, independent verification, metamorphic testing, differential testing, property-based test generation, execution-trace summarization, and uncertainty-aware critics~\citep{ni2023lever,jung-etal-2025-code,tang2026execverify}. Reliable feedback should also be routed differently depending on its type: compiler errors may trigger local syntax repair, test failures may trigger behavioral diagnosis, coverage gaps may trigger test generation, and inconsistent reviewer comments may trigger arbitration. The broader goal is to build feedback loops that are not merely reactive, but epistemically aware: the harness should know when a signal is strong enough to act on, when it is weak, and when additional evidence is required.

\subsubsection{Self-Evolving Harnesses without Regression}

Most current harnesses are manually designed: developers choose the planning loop, memory format, tool set, permission rules, debugging procedure, and agent topology. However, as tasks become longer and more diverse, fixed harnesses may be suboptimal. A harness that works well for competitive programming may fail for repository repair; a harness tuned for GUI navigation may be inefficient for scientific workflows; and a multi-agent topology that succeeds on one task distribution may waste computation on another. This suggests that future systems should treat the harness itself as a programmable component that can adapt to new environments, rather than a fixed wrapper around the base model.

Automatic harness evolution is already underway. AutoHarness synthesizes code harnesses that constrain invalid actions~\citep{lou2026autoharness}, MetaHarness searches over harness code~\citep{lee2026metaharness}, Agentic Harness Engineering evolves harness components from observability signals~\citep{lin2026agentic}, and related methods optimize prompts, contexts, and workflows through reflection, search, or execution feedback \citep{agrawal2025gepa,Liu2025SEW,zhang2025agentic}. These systems point toward a broader paradigm in which an overarching optimization process analyzes runtime feedback, such as computational cost, decision paths, tool-use traces, memory pressure, and specific failure cases, and proposes modifications to the harness itself. Such modifications may reorganize communication among sub-agents, adjust memory allocation, revise retrieval or verification policies, or change how execution feedback is routed through the system. Therefore, ``automated harness evolution'' is not itself the open problem. The harder problem is whether a harness can improve itself without overfitting, weakening safety, increasing cost, hiding failures, or regressing on rare but important tasks.

The central insight is that a harness mutation should be treated like a code change to a safety-critical runtime. Every proposed edit should carry a change contract: which component is modified, which failure mode it targets, what improvement it predicts, which invariants it must preserve, which evaluation can falsify it, and how it can be rolled back. This is especially important because harness changes affect the future distribution of agent behavior. A new retrieval policy may improve benchmark accuracy while increasing hallucinated evidence; a new tool schema may reduce token cost while weakening permission boundaries; a new verifier may improve pass rate by accepting underspecified solutions. Future work should develop evidence-carrying harness evolution, held-out regression suites, safety invariants, canary deployment, rollback semantics, and causal evidence for why a harness edit helped. The goal is not a harness that changes often, but one that changes only when it can justify the change. A practical research agenda includes: defining mutation operators for harness components; building telemetry standards; evaluating evolved harnesses across diverse tasks; enforcing safety invariants during evolution; and separating improvements in the harness from improvements in the base model.

\subsubsection{Transactional Shared Program State and Semantic Conflict Resolution}

Scaling from single agents to multi-agent systems turns the codebase into a shared harness substrate. Planners, coders, testers, reviewers, security agents, and humans may all read and modify overlapping artifacts. Prior sections show that many systems still rely on sequential handoff, shared logs, or file-only state, while newer systems introduce blackboards, repository memories, execution feedback, and explicit belief-state synchronization \citep{Qian2023ChatDev,Hong2023MetaGPT,huang2023agentcoder,wang2025openhands,Guo2025SyncMind}. The open problem is that synchronization alone does not provide transactional semantics or assumption-level consistency: these mechanisms often synchronize artifacts but not assumptions. One agent may plan from an old repository snapshot, another may test a newer patch, a third may remember an obsolete invariant, and a human reviewer may introduce a new constraint that is not propagated to the rest of the system.

The missing abstraction is transactional shared program state. Agents should not merely append messages to a common log; each action should declare its read set, write set, assumptions, version dependencies, verifier obligations, and conflict policy. Conflicts should be detected not only at the level of file diffs, but also at the level of plans, tests, retrieved evidence, permissions, memory entries, and latent user requirements. Future harnesses need conflict-resolution mechanisms that are semantic rather than purely textual, including semantic merge, rollback, dependency-aware locking, belief-state reconciliation, conflict explanation, and re-verification after merge. Classical version control, databases, CRDTs, and build systems provide useful analogies, but agentic systems add conflicts that conventional tools do not see: incompatible plans, stale memories, duplicated subtasks, inconsistent tool authority, and divergent interpretations of the user's goal. A key research challenge is to determine when a conflict can be resolved automatically and when it requires external judgment. Such mechanisms also require metrics beyond merge correctness, including merge success, semantic regression rate, rollback frequency, conflict recurrence, and the cost of human intervention.

\subsubsection{Human-in-the-Loop Safety and Accountability as Harness State}

As code-as-agent-harness systems are used in increasingly consequential settings, safety cannot be delegated to the base model or encoded only as a natural-language instruction. In critical domains such as software deployment, cybersecurity, finance, healthcare, scientific experimentation, enterprise automation, and embodied control, agent actions may affect production systems, private data, external users, physical devices, or institutional compliance. A harness therefore needs to function not only as a context manager or tool executor, but also as a safety governor between model intent and real-world consequence. It should classify proposed actions by risk, enforce permission tiers, deny actions that violate hard constraints, and require human approval for irreversible or externally consequential transitions. For example, when an agent requests credentials, modifies security-critical code, accesses user data, deploys a service, issues financial or medical recommendations, or controls physical equipment, the harness should be able to override the base model and suspend autonomy until a human decision is made \citep{Nunez2024AutoSafeCoder,vijayvargiya2025openagentsafety,guan2025normcode}.

Future harnesses need explicit governance mechanisms that mediate between model intent and environmental action. A useful design pattern is a multi-tier permission model. At the lowest tier, agents may read files, inspect logs, and run static analysis. At higher tiers, they may edit local files, execute sandboxed code, access the network, call external APIs, modify shared repositories, or affect production systems. Each tier should specify its allowed actions, constraints, audit logs, rollback mechanisms, and human-in-the-loop gates for high-risk operations. Such governance must also be context-sensitive. The same command may be safe in a disposable sandbox but unsafe in a production repository, and the same network request may be benign during documentation retrieval but risky when it transmits local state. Therefore, permissions should depend not only on tool identity, but also on arguments, environment state, data sensitivity, and expected side effects. Open problems include policy specification, side-effect prediction, sandbox escape prevention, secret handling, secure tool schemas, reversible execution, and measuring the tradeoff between autonomy and safety.

This safety role also changes how human feedback should be represented. Human-in-the-loop control should not appear only as an occasional prompt interruption; it should become durable harness state. Each approval, rejection, policy exception, or reviewer correction should update the harness's permission rules, escalation policy, verification criteria, and future memory retrieval. Likewise, high-stakes approvals should be auditable state transitions: what action was proposed, what evidence was shown, what risks were surfaced, who approved or rejected it, and what responsibility boundary changed afterward. The open problem is to design harnesses that can decide when autonomy is appropriate and when human judgment is mandatory. In this view, reliable code-as-agent-harness systems require not only executable code and verifiable feedback, but also executable accountability: a safety layer that filters, vetoes, escalates, and records agent actions before they reach the real world.

\subsubsection{Multimodal Code-Harness Systems}

Most code-agent harnesses are still designed around textual state: prompts, files, logs, tool outputs, tests, and execution traces. However, many emerging agentic systems operate in environments where the critical state is multimodal. GUI agents observe screenshots, accessibility trees, and rendered interface states; embodied agents rely on egocentric images, depth, force, tactile signals, object poses, and simulator or robot states; scientific agents inspect plots, microscope images, molecular structures, and experimental readouts. In these settings, the harness can no longer treat perception as a passive input to the model. It must manage multimodal observations as persistent, queryable, and verifiable state.

A central challenge is multimodal context compression. Visual observations are large, redundant, and often only partially relevant to the task. A GUI screenshot may contain hundreds of elements, while only one button matters; an embodied trajectory may contain thousands of frames, while only a few reveal task-critical object relations, contact events, or failure causes. Future harnesses need compression mechanisms that preserve task-relevant visual evidence rather than merely reduce token cost. This suggests a multi-level memory design: raw images or frames are stored as immutable evidence; object-, region-, element-, and pose-level annotations provide structured intermediate state; and compact textual or symbolic summaries expose only the information needed for skill retrieval and planning. The open problem is to decide what multimodal information should be retained, abstracted, forgotten, or promoted into long-term memory, especially when later failures reveal that an earlier visual or physical detail was important.

Visual grounding introduces a second challenge: aligning observations with actions. In text-centric harnesses, an action can often be checked against a file, command, or test result. In visual environments, the agent must map language goals to image regions, interface elements, objects, coordinates, poses, and executable actions. A GUI agent must know that a planned click corresponds to the correct rendered button; an embodied agent must know that a grasp command targets the intended object under the current camera view and physical configuration. This requires harness-level grounding contracts that connect perception, action, and verification. Each action should carry not only a natural-language rationale, but also a grounded reference to the evidence it depends on, such as a bounding box, object identifier, UI element, frame index, region feature, object position, or orientation. After execution, the harness should verify whether the intended grounded state changed as expected, rather than relying only on the model's self-report.

Reliable feedback is also harder in multimodal settings. A textual error message or unit-test failure provides an explicit signal, but visual and physical feedback is often implicit, delayed, or ambiguous. A button may look clicked without triggering the right state transition; a robot may appear to hold an object while the grasp is unstable; a chart may seem to support a conclusion while its axis scale changes the interpretation. Future harnesses therefore need multimodal verification stacks that combine visual state checks, object tracking, OCR or UI-tree inspection, simulator state, physical sensors, tactile feedback, and task-specific validators. More importantly, each feedback signal should expose its scope and uncertainty. For example, a bounding-box detector verifies localization but not task completion; a simulator state verifies object position but not physical robustness; an OCR result verifies visible text but not semantic correctness. This also calls for tighter integration between world modeling and action modeling: the harness should predict how the visual or physical world is expected to change after an action, compare that prediction with the observed outcome, and use the mismatch to diagnose failures. In embodied and robotic settings, such prediction-error signals are especially important for recovery, since failures may arise from occlusion, slippage, collision, unreachable poses, or violated preconditions rather than from an explicit error message. Treating multimodal feedback as calibrated evidence, rather than as a binary success signal, is essential for safe long-horizon autonomy.

Multimodal memory should also support skill evolution. In visual-centric domains such as GUI control and embodied manipulation, reusable skills cannot be represented only as text or code snippets. A useful skill often couples a multimodal precondition, an executable action pattern, and an expected postcondition: what the agent should see or sense before acting, what program, UI command, or motor primitive it should execute, and what visual, physical, or state change should follow. For example, a GUI skill may encode how to locate a settings menu from a screenshot, click the correct region, and verify that a new panel appears. An embodied skill may encode how to identify a graspable object, choose an approach pose, execute a primitive controller, and confirm through vision, force, or tactile feedback that the object has moved into the gripper. Such skills should evolve from successful trajectories, failed attempts, and human corrections, while retaining their grounding evidence. The harness must therefore decide when a visual-action pattern is reusable, how abstractly it should be stored, and how to adapt it across layouts, viewpoints, embodiments, sensors, or tasks.

\subsubsection{Toward a Science of Harness Engineering}

Taken together, these open problems suggest that code-as-harness research is moving toward a broader science of harness engineering. The central object of study is no longer only the model or the generated program, but the complete closed-loop system: context, memory, tools, execution, feedback, safety, coordination, and evaluation. Progress will require benchmarks that expose long-horizon failures, telemetry that makes trajectories auditable, metrics that isolate harness components, and design principles that allow agents to operate safely in persistent program worlds.

The most important future systems will likely be those that combine four properties. First, they will be \emph{executable}, grounding decisions in code, tools, tests, and environments. Second, they will be \emph{inspectable}, exposing plans, state, provenance, and failure causes. Third, they will be \emph{stateful}, preserving task-relevant information across long trajectories and multiple agents. Fourth, they will be \emph{governed}, ensuring that autonomy is constrained by permissions, verification, and accountability. These properties define the next frontier for reliable, long-horizon agentic AI.

\clearpage
\bibliographystyle{unsrtnat}
\bibliography{reference}

@article{li2025graphcodeagent,
  title={GraphCodeAgent: Dual Graph-Guided LLM Agent for Retrieval-Augmented Repo-Level Code Generation},
  author={Li, Jia and Shi, Xianjie and Zhang, Kechi and Li, Ge and Jin, Zhi and Li, Lei and Zhang, Huangzhao and Liu, Fang and Zhang, Yuwei and Tao, Zhengwei and others},
  journal={arXiv preprint arXiv:2504.10046},
  year={2025}
}

@article{erdogan2025plan,
  title={Plan-and-act: Improving planning of agents for long-horizon tasks},
  author={Erdogan, Lutfi Eren and Lee, Nicholas and Kim, Sehoon and Moon, Suhong and Furuta, Hiroki and Anumanchipalli, Gopala and Keutzer, Kurt and Gholami, Amir},
  journal={arXiv preprint arXiv:2503.09572},
  year={2025}
}

@misc{openai2025execplans,
  author       = {Aaron Friel},
  title        = {Using {PLANS.md} for Multi-Hour Problem Solving},
  howpublished = {OpenAI Cookbook},
  year         = {2025},
  month        = oct,
  day          = {7},
  url          = {https://developers.openai.com/cookbook/articles/codex_exec_plans},
  note         = {Accessed: 2026-05-11}
}

@misc{openai2026codexlonghorizon,
  author       = {Derrick Choi},
  title        = {Run Long Horizon Tasks with Codex},
  howpublished = {OpenAI Developers Blog},
  year         = {2026},
  month        = feb,
  day          = {23},
  url          = {https://developers.openai.com/blog/run-long-horizon-tasks-with-codex},
  note         = {Accessed: 2026-05-11}
}

@misc{openai2026harnessengineering,
  author       = {Ryan Lopopolo},
  title        = {Harness Engineering: Leveraging Codex in an Agent-First World},
  howpublished = {OpenAI Engineering Blog},
  year         = {2026},
  month        = feb,
  day          = {11},
  url          = {https://openai.com/index/harness-engineering/},
  note         = {Accessed: 2026-05-11}
}

@misc{agentsmd2025,
  author       = {{AGENTS.md Contributors}},
  title        = {{AGENTS.md}: A Simple, Open Format for Guiding Coding Agents},
  howpublished = {Project Website},
  year         = {2025},
  url          = {https://agents.md/},
  note         = {Accessed: 2026-05-11}
}

@misc{openai2026agentsmd,
  author       = {{OpenAI}},
  title        = {Custom Instructions with {AGENTS.md}},
  howpublished = {OpenAI Codex Documentation},
  year         = {2026},
  url          = {https://developers.openai.com/codex/guides/agents-md},
  note         = {Accessed: 2026-05-11}
}

@misc{anthropic2025claudememory,
  author       = {{Anthropic}},
  title        = {Best Practices for Claude Code},
  howpublished = {Claude Code Documentation},
  year         = {2025},
  url          = {https://code.claude.com/docs/en/best-practices},
  note         = {Accessed: 2026-05-11}
}

@misc{anthropic2025longrunning,
  author       = {Justin Young},
  title        = {Effective Harnesses for Long-Running Agents},
  howpublished = {Anthropic Engineering Blog},
  year         = {2025},
  month        = nov,
  day          = {26},
  url          = {https://www.anthropic.com/engineering/effective-harnesses-for-long-running-agents},
  note         = {Accessed: 2026-05-11}
}

@misc{anthropic2026longrunningapps,
  author       = {Prithvi Rajasekaran},
  title        = {Harness Design for Long-Running Application Development},
  howpublished = {Anthropic Engineering Blog},
  year         = {2026},
  month        = mar,
  day          = {24},
  url          = {https://www.anthropic.com/engineering/harness-design-long-running-apps},
  note         = {Accessed: 2026-05-11}
}

@misc{cursor2026scalingagents,
  author       = {Wilson Lin},
  title        = {Scaling Long-Running Autonomous Coding},
  howpublished = {Cursor Blog},
  year         = {2026},
  month        = jan,
  day          = {14},
  url          = {https://cursor.com/blog/scaling-agents},
  note         = {Accessed: 2026-05-11}
}

@misc{pan2026nlah,
  author        = {Linyue Pan and Lexiao Zou and Shuo Guo and Jingchen Ni and Hai-Tao Zheng},
  title         = {Natural-Language Agent Harnesses},
  year          = {2026},
  eprint        = {2603.25723},
  archivePrefix = {arXiv},
  primaryClass  = {cs.CL},
  doi           = {10.48550/arXiv.2603.25723},
  url           = {https://arxiv.org/abs/2603.25723}
}

@misc{sweSearch2024,
  author        = {Antonis Antoniades and Albert {\"O}rwall and Kexun Zhang and Yuxi Xie and Anirudh Goyal and William Wang},
  title         = {{SWE-Search}: Enhancing Software Agents with Monte Carlo Tree Search and Iterative Refinement},
  year          = {2024},
  eprint        = {2410.20285},
  archivePrefix = {arXiv},
  primaryClass  = {cs.AI},
  doi           = {10.48550/arXiv.2410.20285},
  url           = {https://arxiv.org/abs/2410.20285},
  note          = {Revised Apr. 2, 2025}
}

@misc{cursor2025composer,
  title={Composer: Building a fast frontier model with reinforcement learning},
  author={{Cursor Team}},
  year={2025},
  month={October},
  howpublished={\url{https://cursor.com/blog/composer}},
  note={Cursor blog}
}

@misc{cursor2025rtrl,
  title={Improving {Composer} through real-time reinforcement learning},
  author={{Cursor Team}},
  year={2025},
  howpublished={\url{https://cursor.com/blog/real-time-rl-for-composer}},
  note={Cursor blog}
}

@misc{codex2025,
  title={Introducing {Codex}},
  author={{OpenAI}},
  year={2025},
  month={May},
  howpublished={\url{https://openai.com/index/introducing-codex/}},
  note={OpenAI announcement}
}

@techreport{openai2025gpt5codexcard,
  title={Addendum to {GPT-5} system card: {GPT-5-Codex}},
  author={{OpenAI}},
  year={2025},
  month={September},
  institution={OpenAI},
  note={\url{https://cdn.openai.com/pdf/97cc5669-7a25-4e63-b15f-5fd5bdc4d149/gpt-5-codex-system-card.pdf}}
}

@misc{openai2026codexmax,
  title={Building more with {GPT-5.1-Codex-Max}},
  author={{OpenAI}},
  year={2025},
  howpublished={\url{https://openai.com/index/gpt-5-1-codex-max/}},
  note={OpenAI announcement}
}

@techreport{anthropic2025teams,
  title={How {Anthropic} teams use {Claude Code}},
  author={{Anthropic}},
  year={2025},
  institution={Anthropic},
  note={\url{https://www-cdn.anthropic.com/58284b19e702b49db9302d5b6f135ad8871e7658.pdf}}
}

@article{wang2025solved,
  title={Are" solved issues" in swe-bench really solved correctly? an empirical study},
  author={Wang, You and Pradel, Michael and Liu, Zhongxin},
  journal={arXiv preprint arXiv:2503.15223},
  year={2025}
}

@article{anonymous2025swebenchpp,
  title={SWE-Bench++: A Framework for the Scalable Generation of Software Engineering Benchmarks from Open-Source Repositories},
  author={Wang, Lilin and Ramalho, Lucas and Celestino, Alan and Pham, Phuc Anthony and Liu, Yu and Sinha, Umang Kumar and Portillo, Andres and Osunwa, Onassis and Maduekwe, Gabriel},
  journal={arXiv preprint arXiv:2512.17419},
  year={2025}
}

@misc{openai2025aardvark,
  title={Introducing {Aardvark}: {OpenAI}'s agentic security researcher},
  author={{OpenAI}},
  year={2025},
  month={October},
  howpublished={\url{https://openai.com/index/introducing-aardvark/}}
}

@misc{openai2026codexsecurity,
  title={{Codex Security}: now in research preview},
  author={{OpenAI}},
  year={2026},
  month={March},
  howpublished={\url{https://openai.com/index/codex-security-now-in-research-preview/}}
}

@article{cemri2025whymas,
  title={Why Do Multi-Agent {LLM} Systems Fail?},
  author={Cemri, Mert and Pan, Melissa Z. and Yang, Shuyi and Agrawal, Lakshya A. and Chopra, Bhavya and Tiwari, Rishabh and Keutzer, Kurt and Parameswaran, Aditya and Klein, Dan and Ramchandran, Kannan and Zaharia, Matei and Gonzalez, Joseph E. and Stoica, Ion},
  journal={arXiv preprint arXiv:2503.13657},
  year={2025}
}

@article{agentracer2025,
  title={AgenTracer: Who Is Inducing Failure in the LLM Agentic Systems?},
  author={Zhang, Guibin and Wang, Junhao and Chen, Junjie and Zhou, Wangchunshu and Wang, Kun and Yan, Shuicheng},
  journal={arXiv preprint arXiv:2509.03312},
  year={2025}
}

@article{zhu2025llm,
  title={Where llm agents fail and how they can learn from failures},
  author={Zhu, Kunlun and Liu, Zijia and Li, Bingxuan and Tian, Muxin and Yang, Yingxuan and Zhang, Jiaxun and Han, Pengrui and Xie, Qipeng and Cui, Fuyang and Zhang, Weijia and others},
  journal={arXiv preprint arXiv:2509.25370},
  year={2025}
}

@article{anonymous2026aethelgard,
  title={Beyond static sandboxing: Learned capability governance for autonomous ai agents},
  author={Sidik, Bronislav and Rokach, Lior},
  journal={arXiv preprint arXiv:2604.11839},
  year={2026}
}

@article{anonymous2025faultsandbox,
  title={Fault-Tolerant Sandboxing for {AI} Coding Agents: A Transactional Approach to Safe Autonomous Execution},
  author={Boyang Yan},
  journal={arXiv preprint arXiv:2512.12806},
  year={2025}
}

@misc{microsoft2026governance,
  title={Introducing the {Agent Governance Toolkit}: open-source runtime security for {AI} agents},
  author={{Microsoft}},
  year={2026},
  month={April},
  howpublished={\url{https://opensource.microsoft.com/blog/2026/04/02/introducing-the-agent-governance-toolkit-open-source-runtime-security-for-ai-agents/}}
}

@article{peng2023copilot,
  title={The Impact of AI on Developer Productivity: Evidence from GitHub Copilot},
  author={Peng, Sida and Kalliamvakou, Eirini and Cihon, Peter and Demirer, Mert},
  journal={arXiv preprint arXiv:2302.06590},
  year={2023}
}

@inproceedings{wang2024openhands,
  title={{OpenHands}: An Open Platform for {AI} Software Developers as Generalist Agents},
  author={Wang, Xingyao and Li, Boxuan and Song, Yufan and Xu, Frank F. and Tang, Xiangru and Zhuge, Mingchen and Pan, Jiayi and Song, Yueqi and Li, Bowen and Singh, Jaskirat and Tran, Hoang H. and Li, Fuqiang and Ma, Ren and Zheng, Mingzhang and Qian, Bill and Shao, Yanjun and Muennighoff, Niklas and Zhang, Yizhe and Hui, Binyuan and Lin, Junyang and Brennan, Robert and Peng, Hao and Ji, Heng and Neubig, Graham},
  booktitle={International Conference on Learning Representations (ICLR)},
  year={2025}
}

@article{xia2024agentless,
  title={Agentless: Demystifying {LLM}-based Software Engineering Agents},
  author={Xia, Chunqiu Steven and Deng, Yinlin and Dunn, Soren and Zhang, Lingming},
  journal={arXiv preprint arXiv:2407.01489},
  year={2024}
}

@article{ridnik2024alphacodium,
  title={Code Generation with {AlphaCodium}: From Prompt Engineering to Flow Engineering},
  author={Ridnik, Tal and Kredo, Dedy and Friedman, Itamar},
  journal={arXiv preprint arXiv:2401.08500},
  year={2024}
}

@inproceedings{vaithilingam2022expectation,
  title={Expectation vs.\ Experience: Evaluating the Usability of Code Generation Tools Powered by Large Language Models},
  author={Vaithilingam, Priyan and Zhang, Tianyi and Glassman, Elena L.},
  booktitle={Extended Abstracts of the 2022 CHI Conference on Human Factors in Computing Systems (CHI EA)},
  year={2022},
  doi={10.1145/3491101.3519665}
}

@article{mozannar2022reading,
  title={Reading Between the Lines: Modeling User Behavior and Costs in {AI}-Assisted Programming},
  author={Mozannar, Hussein and Bansal, Gagan and Fourney, Adam and Horvitz, Eric},
  journal={arXiv preprint arXiv:2210.14306},
  year={2022}
}

@article{lu2025requirements,
  title={Requirements Development and Formalization for Reliable Code Generation: A Multi-Agent Vision},
  author={Lu, Xu and Sun, Weisong and Zhang, Yiran and Hu, Ming and Tian, Cong and Jin, Zhi and Liu, Yang},
  journal={arXiv preprint arXiv:2508.18675},
  year={2025}
}

@inproceedings{wang2024executable,
  title={Executable code actions elicit better LLM agents},
  author={Wang, Xingyao and Chen, Yangyi and Yuan, Lifan and Zhang, Yizhe and Li, Yunzhu and Peng, Hao and Ji, Heng},
  booktitle={Proceedings of the 41st International Conference on Machine Learning},
  pages={50208--50232},
  year={2024}
}

@article{meng2026agent,
  title={Agent Harness for Large Language Model Agents: A Survey},
  author={Meng, Qianyu and Wang, Yanan and Chen, Liyi and Wang, Qimeng and Lu, Chengqiang and Wu, Wei and Gao, Yan and Wu, Yi and Hu, Yao},
  year={2026},
  publisher={Preprints}
}

@article{zhang2026sgagent,
  title={SGAgent: Suggestion-Guided LLM-Based Multi-Agent Framework for Repository-Level Software Repair},
  author={Zhang, Quanjun and Gao, Chengyu and Han, Yu and Shang, Ye and Fang, Chunrong and Chen, Zhenyu and Xiao, Liang},
  journal={arXiv preprint arXiv:2602.23647},
  year={2026}
}

@article{khan2025macog,
  title={Multi-Agent Code-Orchestrated Generation for Reliable Infrastructure-as-Code},
  author={Khan, Rana Nameer Hussain and Wasif, Dawood and Cho, Jin-Hee and Butt, Ali},
  journal={arXiv preprint arXiv:2510.03902},
  year={2025},
}

@article{doualgoforge,
  title={AlgoForge: Specializing Code Generation Agents through Collaborative Reinforcement Learning},
  author={Dou, Zhihao and Zhao, Qinjian and Biswas, Sumon}
}

@article{mao2025blueprint2code,
  title={Blueprint2Code: a multi-agent pipeline for reliable code generation via blueprint planning and repair},
  author={Mao, Kehao and Hu, Baokun and Lin, Ruixin and Li, Zewen and Lu, Guanyu and Zhang, Zhengyu},
  journal={Frontiers in Artificial Intelligence},
  volume={8},
  pages={1660912},
  year={2025},
  publisher={Frontiers Media SA}
}

@inproceedings{ukai2024adacoder,
  title={Adacoder: Adaptive prompt compression for programmatic visual question answering},
  author={Ukai, Mahiro and Kurita, Shuhei and Hashimoto, Atsushi and Ushiku, Yoshitaka and Inoue, Nakamasa},
  booktitle={Proceedings of the 32nd ACM International Conference on Multimedia},
  pages={9234--9243},
  year={2024}
}

@article{linearplan1,
  title={PaT: Planning-after-Trial for Efficient Code Generation},
  author={Yoon, Youngsik and Lee, Sungjae and Song, Seockbean and Wang, Siwei and Chen, Wei and Ok, Jungseul}
}

@article{zhang2025linearplan2,
  title={A Little Help Goes a Long Way: Tutoring LLMs in Solving Competitive Programming through Hints},
  author={Zhang, Yating and Dong, Wei and Liu, Jiaxin and Wang, Shangwen and Wang, Deze and Ma, Tiecheng and Li, Yiwei and Yang, Kang},
  journal={IEEE Transactions on Software Engineering},
  year={2025},
  publisher={IEEE}
}

@article{jiang2024selfplanning,
  title={Self-planning code generation with large language models},
  author={Jiang, Xue and Dong, Yihong and Wang, Lecheng and Fang, Zheng and Shang, Qiwei and Li, Ge and Jin, Zhi and Jiao, Wenpin},
  journal={ACM Transactions on Software Engineering and Methodology},
  volume={33},
  number={7},
  pages={1--30},
  year={2024},
  publisher={ACM New York, NY}
}

@article{gur2023webagent,
  title={A real-world webagent with planning, long context understanding, and program synthesis},
  author={Gur, Izzeddin and Furuta, Hiroki and Huang, Austin and Safdari, Mustafa and Matsuo, Yutaka and Eck, Douglas and Faust, Aleksandra},
  journal={arXiv preprint arXiv:2307.12856},
  year={2023}
}

@inproceedings{huang2024knowledge,
  title={Knowledge-aware code generation with large language models},
  author={Huang, Tao and Sun, Zhihong and Jin, Zhi and Li, Ge and Lyu, Chen},
  booktitle={Proceedings of the 32nd IEEE/ACM International Conference on Program Comprehension},
  pages={52--63},
  year={2024}
}

@article{lyu2025reloc,
  title={Let's Revise Step-by-Step: A Unified Local Search Framework for Code Generation with LLMs},
  author={Lyu, Zhiyi and Huang, Jianguo and Deng, Yanchen and Hoi, Steven and An, Bo},
  journal={arXiv preprint arXiv:2508.07434},
  year={2025}
}

@inproceedings{light2025sfs,
  title={SFS: Smarter code space search improves LLM inference scaling},
  author={Light, Jonathan and Wu, Yue and Sun, Yiyou and Yu, Wenchao and Liu, Yanchi and Zhao, Xujiang and Hu, Ziniu and Chen, Haifeng and Cheng, Wei},
  booktitle={The Thirteenth International Conference on Learning Representations},
  year={2025}
}

@inproceedings{li2025codetree,
  title={Codetree: Agent-guided tree search for code generation with large language models},
  author={Li, Jierui and Le, Hung and Zhou, Yingbo and Xiong, Caiming and Savarese, Silvio and Sahoo, Doyen},
  booktitle={Proceedings of the 2025 Conference of the Nations of the Americas Chapter of the Association for Computational Linguistics: Human Language Technologies (Volume 1: Long Papers)},
  pages={3711--3726},
  year={2025}
}

@article{ni2024treeofcode,
  title={Tree-of-code: A tree-structured exploring framework for end-to-end code generation and execution in complex task handling},
  author={Ni, Ziyi and Li, Yifan and Yang, Ning and Shen, Dou and Lv, Pin and Dong, Daxiang},
  journal={arXiv preprint arXiv:2412.15305},
  year={2024}
}

@article{dainese2024codegenerating,
  title={Generating code world models with large language models guided by monte carlo tree search},
  author={Dainese, Nicola and Merler, Matteo and Alakuijala, Minttu and Marttinen, Pekka},
  journal={Advances in Neural Information Processing Systems},
  volume={37},
  pages={60429--60474},
  year={2024}
}

@inproceedings{aggarwal2025dars,
  title={Dars: Dynamic action re-sampling to enhance coding agent performance by adaptive tree traversal},
  author={Aggarwal, Vaibhav and Kamal, Ojasv and Japesh, Abhinav and Jin, Zhijing and Sch{\"o}lkopf, Bernhard},
  booktitle={Proceedings of the 63rd Annual Meeting of the Association for Computational Linguistics (Volume 1: Long Papers)},
  pages={19808--19855},
  year={2025}
}

@article{lumer2025tool,
  title={Tool-to-agent retrieval: Bridging tools and agents for scalable llm multi-agent systems},
  author={Lumer, Elias and Nizar, Faheem and Gulati, Anmol and Basavaraju, Pradeep Honaganahalli and Subbiah, Vamse Kumar},
  journal={arXiv preprint arXiv:2511.01854},
  year={2025}
}

@article{wang2024planning,
  title={Planning in natural language improves llm search for code generation},
  author={Wang, Evan and Cassano, Federico and Wu, Catherine and Bai, Yunfeng and Song, Will and Nath, Vaskar and Han, Ziwen and Hendryx, Sean and Yue, Summer and Zhang, Hugh},
  journal={arXiv preprint arXiv:2409.03733},
  year={2024}
}

@inproceedings{li2025rethinkmcts,
  title={Rethinkmcts: Refining erroneous thoughts in monte carlo tree search for code generation},
  author={Li, Qingyao and Xia, Wei and Dai, Xinyi and Du, Kounianhua and Liu, Weiwen and Wang, Yasheng and Tang, Ruiming and Yu, Yong and Zhang, Weinan},
  booktitle={Proceedings of the 2025 Conference on Empirical Methods in Natural Language Processing},
  pages={8103--8121},
  year={2025}
}

@inproceedings{ho2025verilogcoder,
  title={Verilogcoder: Autonomous verilog coding agents with graph-based planning and abstract syntax tree (ast)-based waveform tracing tool},
  author={Ho, Chia-Tung and Ren, Haoxing and Khailany, Brucek},
  booktitle={Proceedings of the AAAI Conference on Artificial Intelligence},
  volume={39},
  number={1},
  pages={300--307},
  year={2025}
}

@article{wang2026domagent,
  title={DomAgent: Leveraging Knowledge Graphs and Case-Based Reasoning for Domain-Specific Code Generation},
  author={Wang, Shuai and Parthasarathy, Dhasarathy and Feldt, Robert and Yu, Yinan},
  journal={arXiv preprint arXiv:2603.21430},
  year={2026}
}

@article{luo2025rpg,
  title={RPG: A Repository Planning Graph for Unified and Scalable Codebase Generation},
  author={Luo, Jane and Zhang, Xin and Liu, Steven and Wu, Jie and Liu, Jianfeng and Huang, Yiming and Huang, Yangyu and Yin, Chengyu and Xin, Ying and Zhan, Yuefeng and others},
  journal={arXiv preprint arXiv:2509.16198},
  year={2025}
}

@inproceedings{chen2025locagent,
  title={Locagent: Graph-guided llm agents for code localization},
  author={Chen, Zhaoling and Tang, Robert and Deng, Gangda and Wu, Fang and Wu, Jialong and Jiang, Zhiwei and Prasanna, Viktor and Cohan, Arman and Wang, Xingyao},
  booktitle={Proceedings of the 63rd Annual Meeting of the Association for Computational Linguistics (Volume 1: Long Papers)},
  pages={8697--8727},
  year={2025}
}

@article{tao2025cgm,
  title={Code graph model (cgm): A graph-integrated large language model for repository-level software engineering tasks},
  author={Tao, Hongyuan and Zhang, Ying and Tang, Zhenhao and Peng, Hongen and Zhu, Xukun and Liu, Bingchang and Yang, Yingguang and Zhang, Ziyin and Xu, Zhaogui and Zhang, Haipeng and others},
  journal={arXiv preprint arXiv:2505.16901},
  year={2025}
}

@article{wei2022chain,
  title={Chain-of-thought prompting elicits reasoning in large language models},
  author={Wei, Jason and Wang, Xuezhi and Schuurmans, Dale and Bosma, Maarten and Xia, Fei and Chi, Ed and Le, Quoc V and Zhou, Denny and others},
  journal={Advances in neural information processing systems},
  volume={35},
  pages={24824--24837},
  year={2022}
}

@misc{
shi2025from,
title={From Code to Correctness: Closing the Last Mile of Code Generation with Hierarchical Debugging},
author={Yuling Shi and Songsong Wang and Chengcheng Wan and Xiaodong Gu},
year={2025},
url={https://openreview.net/forum?id=dwQIVcW1du}
}

@misc{liu2024codexgraphbridginglargelanguage,
      title={CodexGraph: Bridging Large Language Models and Code Repositories via Code Graph Databases}, 
      author={Xiangyan Liu and Bo Lan and Zhiyuan Hu and Yang Liu and Zhicheng Zhang and Fei Wang and Michael Shieh and Wenmeng Zhou},
      year={2024},
      eprint={2408.03910},
      archivePrefix={arXiv},
      primaryClass={cs.SE},
      url={https://arxiv.org/abs/2408.03910}, 
}

@inproceedings{Baqar_2025,
   title={AI Augmented CI/CD Pipelines: From Code Commit to Production with Autonomous Decisions},
   url={http://dx.doi.org/10.1109/FLLM67465.2025.11391007},
   DOI={10.1109/fllm67465.2025.11391007},
   booktitle={2025 3rd International Conference on Foundation and Large Language Models (FLLM)},
   publisher={IEEE},
   author={Baqar, Mohammad and Naqvi, Saba and Khanda, Rajat},
   year={2025},
   month=nov, pages={1041–1048} }

@inproceedings{luo-etal-2024-repoagent,
    title = "{R}epo{A}gent: An {LLM}-Powered Open-Source Framework for Repository-level Code Documentation Generation",
    author = "Luo, Qinyu  and
      Ye, Yining  and
      Liang, Shihao  and
      Zhang, Zhong  and
      Qin, Yujia  and
      Lu, Yaxi  and
      Wu, Yesai  and
      Cong, Xin  and
      Lin, Yankai  and
      Zhang, Yingli  and
      Che, Xiaoyin  and
      Liu, Zhiyuan  and
      Sun, Maosong",
    editor = "Hernandez Farias, Delia Irazu  and
      Hope, Tom  and
      Li, Manling",
    booktitle = "Proceedings of the 2024 Conference on Empirical Methods in Natural Language Processing: System Demonstrations",
    month = nov,
    year = "2024",
    address = "Miami, Florida, USA",
    publisher = "Association for Computational Linguistics",
    url = "https://aclanthology.org/2024.emnlp-demo.46/",
    doi = "10.18653/v1/2024.emnlp-demo.46",
    pages = "436--464"
}

@article{Wu2025IterPrefFP,
  title={IterPref: Focal Preference Learning for Code Generation via Iterative Debugging},
  author={Jie Wu and Haoling Li and Xin Zhang and Jianwen Luo and Yangyu Huang and Ruihang Chu and Yujiu Yang and Scarlett Li},
  journal={ArXiv},
  year={2025},
  volume={abs/2503.02783},
  url={https://api.semanticscholar.org/CorpusID:282402257}
}

@inproceedings{huang2025mldebugging,
  title={Mldebugging: Towards benchmarking code debugging across multi-library scenarios},
  author={Huang, Jinyang and Feng, Xiachong and Chen, Qiguang and Zhao, Hanjie and Cheng, Zihui and Bai, Jiesong and Zhou, Jingxuan and Li, Min and Qin, Libo},
  booktitle={Findings of the Association for Computational Linguistics: ACL 2025},
  pages={5866--5879},
  year={2025}
}

@article{adnan2025debugging,
  title={The Debugging Decay Index: Rethinking Debugging Strategies for Code LLMs},
  author={Adnan, Muntasir and Kuhn, Carlos CN},
  journal={arXiv preprint arXiv:2506.18403},
  year={2025}
}

@article{gu2024testart,
  title={Testart: Improving llm-based unit testing via co-evolution of automated generation and repair iteration},
  author={Gu, Siqi and Zhang, Quanjun and Li, Kecheng and Fang, Chunrong and Tian, Fangyuan and Zhu, Liuchuan and Zhou, Jianyi and Chen, Zhenyu},
  journal={arXiv preprint arXiv:2408.03095},
  year={2024}
}

@inproceedings{ross2023programmer,
  title={The programmer’s assistant: Conversational interaction with a large language model for software development},
  author={Ross, Steven I and Martinez, Fernando and Houde, Stephanie and Muller, Michael and Weisz, Justin D},
  booktitle={Proceedings of the 28th international conference on intelligent user interfaces},
  pages={491--514},
  year={2023}
}

@article{fakhoury2024llm,
  title={Llm-based test-driven interactive code generation: User study and empirical evaluation},
  author={Fakhoury, Sarah and Naik, Aaditya and Sakkas, Georgios and Chakraborty, Saikat and Lahiri, Shuvendu K},
  journal={IEEE Transactions on Software Engineering},
  volume={50},
  number={9},
  pages={2254--2268},
  year={2024},
  publisher={IEEE}
}

@article{chen2023teaching,
  title={Teaching large language models to self-debug},
  author={Chen, Xinyun and Lin, Maxwell and Sch{\"a}rli, Nathanael and Zhou, Denny},
  journal={arXiv preprint arXiv:2304.05128},
  year={2023}
}

@article{sun2024llm,
  title={Llm as runtime error handler: A promising pathway to adaptive self-healing of software systems},
  author={Sun, Zhensu and Zhu, Haotian and Xu, Bowen and Du, Xiaoning and Li, Li and Lo, David},
  journal={arXiv preprint arXiv:2408.01055},
  year={2024}
}

@misc{zou2025latentmas,
      title={Latent Collaboration in Multi-Agent Systems}, 
      author={Jiaru Zou and Xiyuan Yang and Ruizhong Qiu and Gaotang Li and Katherine Tieu and Pan Lu and Ke Shen and Hanghang Tong and Yejin Choi and Jingrui He and James Zou and Mengdi Wang and Ling Yang},
      year={2025},
      eprint={2511.20639},
      archivePrefix={arXiv},
      primaryClass={cs.CL},
      url={https://arxiv.org/abs/2511.20639}, 
}

@inproceedings{blyth2025static,
  title={Static analysis as a feedback loop: Enhancing llm-generated code beyond correctness},
  author={Blyth, Scott and Licorish, Sherlock A and Treude, Christoph and Wagner, Markus},
  booktitle={2025 IEEE International Conference on Source Code Analysis \& Manipulation (SCAM)},
  pages={100--109},
  year={2025},
  organization={IEEE}
}

@inproceedings{bi2024iterative,
  title={Iterative refinement of project-level code context for precise code generation with compiler feedback},
  author={Bi, Zhangqian and Wan, Yao and Wang, Zheng and Zhang, Hongyu and Guan, Batu and Lu, Fangxin and Zhang, Zili and Sui, Yulei and Jin, Hai and Shi, Xuanhua},
  booktitle={Findings of the Association for Computational Linguistics: ACL 2024},
  pages={2336--2353},
  year={2024}
}

@article{shinn2023reflexion,
  title={Reflexion: Language agents with verbal reinforcement learning},
  author={Shinn, Noah and Cassano, Federico and Gopinath, Ashwin and Narasimhan, Karthik and Yao, Shunyu},
  journal={Advances in neural information processing systems},
  volume={36},
  pages={8634--8652},
  year={2023}
}

@inproceedings{wu2024autogen,
  title={Autogen: Enabling next-gen LLM applications via multi-agent conversations},
  author={Wu, Qingyun and Bansal, Gagan and Zhang, Jieyu and Wu, Yiran and Li, Beibin and Zhu, Erkang and Jiang, Li and Zhang, Xiaoyun and Zhang, Shaokun and Liu, Jiale and others},
  booktitle={First conference on language modeling},
  year={2024}
}

@misc{zou2026recursivemas,
      title={Recursive Multi-Agent Systems}, 
      author={Jiaru Zou and Rui Pan and Ruizhong Qiu and Pan Lu and Shizhe Diao and Jindong Jiang and Hanghang Tong and Tong Zhang and Markus J. Buehler and Jingrui He and James Zou},
      year={2026},
      eprint={2604.25917},
      archivePrefix={arXiv},
      primaryClass={cs.AI},
      url={https://arxiv.org/abs/2604.25917}, 
}

@article{vijayvargiya2025openagentsafety,
  title={Openagentsafety: A comprehensive framework for evaluating real-world ai agent safety},
  author={Vijayvargiya, Sanidhya and Soni, Aditya Bharat and Zhou, Xuhui and Wang, Zora Zhiruo and Dziri, Nouha and Neubig, Graham and Sap, Maarten},
  journal={arXiv preprint arXiv:2507.06134},
  year={2025}
}

@article{hou2025model,
  title={Model context protocol (mcp): Landscape, security threats, and future research directions},
  author={Hou, Xinyi and Zhao, Yanjie and Wang, Shenao and Wang, Haoyu},
  journal={ACM Transactions on Software Engineering and Methodology},
  year={2025},
  publisher={ACM New York, NY}
}

@article{sergeyuk2026human,
  title={Human-AI experience in integrated development environments: a systematic literature review},
  author={Sergeyuk, Agnia and Zakharov, Ilya and Koshchenko, Ekaterina and Izadi, Maliheh},
  journal={Empirical Software Engineering},
  volume={31},
  number={3},
  pages={55},
  year={2026},
  publisher={Springer}
}

@article{li2025glue,
  title={From glue-code to protocols: A critical analysis of a2a and mcp integration for scalable agent systems},
  author={Li, Qiaomu and Xie, Ying},
  journal={arXiv preprint arXiv:2505.03864},
  year={2025}
}

@inproceedings{RayASO,
  title={A Survey on Model Context Protocol: Architecture, State-of-the-art, Challenges and Future Directions},
  author={Partha Pratim Ray},
  url={https://api.semanticscholar.org/CorpusID:281419186}
}

@inproceedings{gao2023pal,
  title={Pal: Program-aided language models},
  author={Gao, Luyu and Madaan, Aman and Zhou, Shuyan and Alon, Uri and Liu, Pengfei and Yang, Yiming and Callan, Jamie and Neubig, Graham},
  booktitle={International conference on machine learning},
  pages={10764--10799},
  year={2023},
  organization={PMLR}
}

@article{chen2022program,
  title={Program of thoughts prompting: Disentangling computation from reasoning for numerical reasoning tasks},
  author={Chen, Wenhu and Ma, Xueguang and Wang, Xinyi and Cohen, William W},
  journal={arXiv preprint arXiv:2211.12588},
  year={2022}
}

@article{li2023chain,
  title={Chain of code: Reasoning with a language model-augmented code emulator},
  author={Li, Chengshu and Liang, Jacky and Zeng, Andy and Chen, Xinyun and Hausman, Karol and Sadigh, Dorsa and Levine, Sergey and Fei-Fei, Li and Xia, Fei and Ichter, Brian},
  journal={arXiv preprint arXiv:2312.04474},
  year={2023}
}

@article{wang2023mathcoder,
  title={Mathcoder: Seamless code integration in llms for enhanced mathematical reasoning},
  author={Wang, Ke and Ren, Houxing and Zhou, Aojun and Lu, Zimu and Luo, Sichun and Shi, Weikang and Zhang, Renrui and Song, Linqi and Zhan, Mingjie and Li, Hongsheng},
  journal={arXiv preprint arXiv:2310.03731},
  year={2023}
}

@article{ye2023satlm,
  title={Satlm: Satisfiability-aided language models using declarative prompting},
  author={Ye, Xi and Chen, Qiaochu and Dillig, Isil and Durrett, Greg},
  journal={Advances in Neural Information Processing Systems},
  volume={36},
  pages={45548--45580},
  year={2023}
}

@inproceedings{zhao2024expel,
  title={Expel: Llm agents are experiential learners},
  author={Zhao, Andrew and Huang, Daniel and Xu, Quentin and Lin, Matthieu and Liu, Yong-Jin and Huang, Gao},
  booktitle={Proceedings of the AAAI Conference on Artificial Intelligence},
  volume={38},
  number={17},
  pages={19632--19642},
  year={2024}
}

@article{xia2025demystifying,
  title={Demystifying llm-based software engineering agents},
  author={Xia, Chunqiu Steven and Deng, Yinlin and Dunn, Soren and Zhang, Lingming},
  journal={Proceedings of the ACM on Software Engineering},
  volume={2},
  number={FSE},
  pages={801--824},
  year={2025},
  publisher={ACM New York, NY, USA}
}

@article{deng2026your,
  title={Your Code Agent Can Grow Alongside You with Structured Memory},
  author={Deng, Yi-Xuan and Liu, Xiaoqin and Zhang, Yi and Yang, Guo-Wei and Yang, Shuojin},
  journal={arXiv preprint arXiv:2603.13258},
  year={2026}
}

@article{shen2025talm,
  title={TALM: Dynamic Tree-Structured Multi-Agent Framework with Long-Term Memory for Scalable Code Generation},
  author={Shen, Ming-Tung and Joung, Yuh-Jzer},
  journal={arXiv preprint arXiv:2510.23010},
  year={2025}
}

@inproceedings{li2025survey,
  title={A Survey of RAG-Reasoning Systems in Large Language Models},
  author={Li, Yangning and Zhang, Weizhi and Yang, Yuyao and Huang, Wei-Chieh and Wu, Yaozu and Luo, Junyu and Bei, Yuanchen and Zou, Henry Peng and Luo, Xiao and Zhao, Yusheng and others},
  booktitle={Findings of the Association for Computational Linguistics: EMNLP 2025},
  pages={12120--12145},
  year={2025}
}

@inproceedings{qian2024chatdev,
  title={Chatdev: Communicative agents for software development},
  author={Qian, Chen and Liu, Wei and Liu, Hongzhang and Chen, Nuo and Dang, Yufan and Li, Jiahao and Yang, Cheng and Chen, Weize and Su, Yusheng and Cong, Xin and others},
  booktitle={Proceedings of the 62nd annual meeting of the association for computational linguistics (volume 1: Long papers)},
  pages={15174--15186},
  year={2024}
}

@article{zhang2025ragsurvey,
  title={A survey of graph retrieval-augmented generation for customized large language models},
  author={Zhang, Qinggang and Chen, Shengyuan and Bei, Yuanchen and Yuan, Zheng and Zhou, Huachi and Hong, Zijin and Chen, Hao and Xiao, Yilin and Zhou, Chuang and Dong, Junnan and others},
  journal={arXiv preprint arXiv:2501.13958},
  year={2025}
}

@inproceedings{yuan2025easytool,
  title={Easytool: Enhancing llm-based agents with concise tool instruction},
  author={Yuan, Siyu and Song, Kaitao and Chen, Jiangjie and Tan, Xu and Shen, Yongliang and Ren, Kan and Li, Dongsheng and Yang, Deqing},
  booktitle={Proceedings of the 2025 Conference of the Nations of the Americas Chapter of the Association for Computational Linguistics: Human Language Technologies (Volume 1: Long Papers)},
  pages={951--972},
  year={2025}
}

@misc{su2025toolorchestra,
      title={ToolOrchestra: Elevating Intelligence via Efficient Model and Tool Orchestration}, 
      author={Hongjin Su and Shizhe Diao and Ximing Lu and Mingjie Liu and Jiacheng Xu and Xin Dong and Yonggan Fu and Peter Belcak and Hanrong Ye and Hongxu Yin and Yi Dong and Evelina Bakhturina and Tao Yu and Yejin Choi and Jan Kautz and Pavlo Molchanov},
      year={2025},
      eprint={2511.21689},
      archivePrefix={arXiv},
      primaryClass={cs.CL},
      url={https://arxiv.org/abs/2511.21689}, 
}

@misc{zou2025autotool,
      title={AutoTool: Dynamic Tool Selection and Integration for Agentic Reasoning}, 
      author={Jiaru Zou and Ling Yang and Yunzhe Qi and Sirui Chen and Mengting Ai and Ke Shen and Jingrui He and Mengdi Wang},
      year={2025},
      eprint={2512.13278},
      archivePrefix={arXiv},
      primaryClass={cs.CL},
      url={https://arxiv.org/abs/2512.13278}, 
}

@article{yuan2025graphs,
  title={Graphs meet ai agents: Taxonomy, progress, and future opportunities},
  author={Bei, Yuanchen and Zhang, Weizhi and Wang, Siwen and Chen, Weizhi and Zhou, Sheng and Chen, Hao and Li, Yong and Bu, Jiajun and Pan, Shirui and Yu, Yizhou and others},
  journal={arXiv preprint arXiv:2506.18019},
  year={2025}
}

@inproceedings{liang2023code,
  title={Code as policies: Language model programs for embodied control},
  author={Liang, Jacky and Huang, Wenlong and Xia, Fei and Xu, Peng and Hausman, Karol and Ichter, Brian and Florence, Pete and Zeng, Andy},
  booktitle={2023 IEEE International conference on robotics and automation (ICRA)},
  pages={9493--9500},
  year={2023},
  organization={IEEE}
}

@article{huang2022inner,
  title={Inner monologue: Embodied reasoning through planning with language models},
  author={Huang, Wenlong and Xia, Fei and Xiao, Ted and Chan, Harris and Liang, Jacky and Florence, Pete and Zeng, Andy and Tompson, Jonathan and Mordatch, Igor and Chebotar, Yevgen and others},
  journal={arXiv preprint arXiv:2207.05608},
  year={2022}
}

@article{wang2023voyager,
  title={Voyager: An open-ended embodied agent with large language models, 2023},
  author={Wang, Guanzhi and Xie, Yuqi and Jiang, Yunfan and Mandlekar, Ajay and Xiao, Chaowei and Zhu, Yuke and Fan, Linxi and Anandkumar, Anima},
  journal={URL https://arxiv. org/abs/2305.16291},
  volume={2},
  number={11},
  year={2023}
}

@article{vemprala2024chatgpt,
  title={Chatgpt for robotics: Design principles and model abilities},
  author={Vemprala, Sai H and Bonatti, Rogerio and Bucker, Arthur and Kapoor, Ashish},
  journal={Ieee Access},
  volume={12},
  pages={55682--55696},
  year={2024},
  publisher={IEEE}
}

@article{huang2023voxposer,
  title={Voxposer: Composable 3d value maps for robotic manipulation with language models},
  author={Huang, Wenlong and Wang, Chen and Zhang, Ruohan and Li, Yunzhu and Wu, Jiajun and Fei-Fei, Li},
  journal={arXiv preprint arXiv:2307.05973},
  year={2023}
}

@inproceedings{chen2023vistruct,
  title={Vistruct: Visual structural knowledge extraction via curriculum guided code-vision representation},
  author={Chen, Yangyi and Wang, Xingyao and Li, Manling and Hoiem, Derek and Ji, Heng},
  booktitle={Proceedings of the 2023 Conference on Empirical Methods in Natural Language Processing},
  pages={13342--13357},
  year={2023}
}

@article{ding2024semcoder,
  title={Semcoder: Training code language models with comprehensive semantics},
  author={Ding, Yangruibo and Peng, Jinjun and Min, Marcus J and Kaiser, Gail and Yang, Junfeng and Ray, Baishakhi},
  journal={arXiv preprint arXiv:2406.01006},
  volume={47},
  year={2024}
}

@article{armengol2025cannot,
  title={What I cannot execute, I do not understand: Training and Evaluating LLMs on Program Execution Traces},
  author={Armengol-Estap{\'e}, Jordi and Carbonneaux, Quentin and Zhang, Tianjun and Markosyan, Aram H and Seeker, Volker and Cummins, Chris and Kambadur, Melanie and O'Boyle, Michael FP and Wang, Sida and Synnaeve, Gabriel and others},
  journal={arXiv preprint arXiv:2503.05703},
  year={2025}
}

@article{tang2024worldcoder,
  title={Worldcoder, a model-based llm agent: Building world models by writing code and interacting with the environment},
  author={Tang, Hao and Key, Darren and Ellis, Kevin},
  journal={Advances in Neural Information Processing Systems},
  volume={37},
  pages={70148--70212},
  year={2024}
}

@article{jimenez2023swe,
  title={Swe-bench: Can language models resolve real-world github issues?},
  author={Jimenez, Carlos E and Yang, John and Wettig, Alexander and Yao, Shunyu and Pei, Kexin and Press, Ofir and Narasimhan, Karthik},
  journal={arXiv preprint arXiv:2310.06770},
  year={2023}
}

@article{yang2024swe,
  title={Swe-agent: Agent-computer interfaces enable automated software engineering},
  author={Yang, John and Jimenez, Carlos E and Wettig, Alexander and Lieret, Kilian and Yao, Shunyu and Narasimhan, Karthik and Press, Ofir},
  journal={Advances in Neural Information Processing Systems},
  volume={37},
  pages={50528--50652},
  year={2024}
}

@article{liu2023agentbench,
  title={Agentbench: Evaluating llms as agents},
  author={Liu, Xiao and Yu, Hao and Zhang, Hanchen and Xu, Yifan and Lei, Xuanyu and Lai, Hanyu and Gu, Yu and Ding, Hangliang and Men, Kaiwen and Yang, Kejuan and others},
  journal={arXiv preprint arXiv:2308.03688},
  year={2023}
}

@article{yang2023intercode,
  title={Intercode: Standardizing and benchmarking interactive coding with execution feedback},
  author={Yang, John and Prabhakar, Akshara and Narasimhan, Karthik and Yao, Shunyu},
  journal={Advances in Neural Information Processing Systems},
  volume={36},
  pages={23826--23854},
  year={2023}
}

@article{dong2025survey,
  title={A survey on code generation with llm-based agents},
  author={Dong, Yihong and Jiang, Xue and Qian, Jiaru and Wang, Tian and Zhang, Kechi and Jin, Zhi and Li, Ge},
  journal={arXiv preprint arXiv:2508.00083},
  year={2025}
}

@article{zhang2025survey,
  title={A survey on the memory mechanism of large language model-based agents},
  author={Zhang, Zeyu and Dai, Quanyu and Bo, Xiaohe and Ma, Chen and Li, Rui and Chen, Xu and Zhu, Jieming and Dong, Zhenhua and Wen, Ji-Rong},
  journal={ACM Transactions on Information Systems},
  volume={43},
  number={6},
  pages={1--47},
  year={2025},
  publisher={ACM New York, NY}
}

@article{huang2026rethinking,
  title={Rethinking Memory Mechanisms of Foundation Agents in the Second Half},
  author={Huang, Wei-Chieh and Zhang, Weizhi and Liang, Yueqing and Bei, Yuanchen and Chen, Yankai and Feng, Tao and Pan, Xinyu and Tan, Zhen and Wang, Yu and Wei, Tianxin and others},
  journal={arXiv preprint arXiv:2602.06052},
  year={2026}
}

@article{jiang2026survey,
  title={A survey on large language models for code generation},
  author={Jiang, Juyong and Wang, Fan and Shen, Jiasi and Kim, Sungju and Kim, Sunghun},
  journal={ACM Transactions on Software Engineering and Methodology},
  volume={35},
  number={2},
  pages={1--72},
  year={2026},
  publisher={ACM New York, NY}
}

@article{huang2025language,
  title={On the Failure of Latent State Persistence in Large Language Models},
  author={Huang, Jen-tse and Sun, Kaiser and Wang, Wenxuan and Dredze, Mark},
  journal={arXiv preprint arXiv:2505.10571},
  year={2025}
}

@article{wei2025evo,
  title={Evo-memory: Benchmarking llm agent test-time learning with self-evolving memory},
  author={Wei, Tianxin and Sachdeva, Noveen and Coleman, Benjamin and He, Zhankui and Bei, Yuanchen and Ning, Xuying and Ai, Mengting and Li, Yunzhe and He, Jingrui and Chi, Ed H and others},
  journal={arXiv preprint arXiv:2511.20857},
  year={2025}
}

@article{dong2025towards,
  title={Towards large language models with human-like episodic memory},
  author={Dong, Cody V and Lu, Qihong and Norman, Kenneth A and Michelmann, Sebastian},
  journal={Trends in Cognitive Sciences},
  year={2025},
  publisher={Elsevier}
}

@inproceedings{
huet2025episodic,
title={Episodic Memories Generation and Evaluation Benchmark for Large Language Models},
author={Alexis Huet and Zied Ben Houidi and Dario Rossi},
booktitle={The Thirteenth International Conference on Learning Representations},
year={2025}
}

@inproceedings{
zhang2025gmemory,
title={G-Memory: Tracing Hierarchical Memory for Multi-Agent Systems},
author={Guibin Zhang and Muxin Fu and Kun Wang and Guancheng Wan and Miao Yu and Shuicheng YAN},
booktitle={The Thirty-ninth Annual Conference on Neural Information Processing Systems},
year={2025}
}

@article{bei2026mem,
  title={Mem-gallery: Benchmarking multimodal long-term conversational memory for mllm agents},
  author={Bei, Yuanchen and Wei, Tianxin and Ning, Xuying and Zhao, Yanjun and Liu, Zhining and Lin, Xiao and Zhu, Yada and Hamann, Hendrik and He, Jingrui and Tong, Hanghang},
  journal={arXiv preprint arXiv:2601.03515},
  year={2026}
}

@misc{zhao2026papermind,
      title={PaperMind: Benchmarking Agentic Reasoning and Critique over Scientific Papers in Multimodal LLMs}, 
      author={Yanjun Zhao and Tianxin Wei and Jiaru Zou and Xuying Ning and Yuanchen Bei and Lingjie Chen and Simmi Rana and Wendy H. Yang and Hanghang Tong and Jingrui He},
      year={2026},
      eprint={2604.21304},
      archivePrefix={arXiv},
      primaryClass={cs.IR},
      url={https://arxiv.org/abs/2604.21304}, 
}

@inproceedings{maharana2024evaluating,
  title={Evaluating very long-term conversational memory of llm agents},
  author={Maharana, Adyasha and Lee, Dong-Ho and Tulyakov, Sergey and Bansal, Mohit and Barbieri, Francesco and Fang, Yuwei},
  booktitle={Proceedings of the 62nd Annual Meeting of the Association for Computational Linguistics (Volume 1: Long Papers)},
  pages={13851--13870},
  year={2024}
}

@inproceedings{kang2025memory,
  title={Memory os of ai agent},
  author={Kang, Jiazheng and Ji, Mingming and Zhao, Zhe and Bai, Ting},
  booktitle={Proceedings of the 2025 Conference on Empirical Methods in Natural Language Processing},
  pages={25972--25981},
  year={2025}
}

@article{xia2025live,
  title={Live-SWE-agent: Can Software Engineering Agents Self-Evolve on the Fly?},
  author={Xia, Chunqiu Steven and Wang, Zhe and Yang, Yan and Wei, Yuxiang and Zhang, Lingming},
  journal={arXiv preprint arXiv:2511.13646},
  year={2025}
}

@article{li2025swe,
  title={Swe-debate: Competitive multi-agent debate for software issue resolution},
  author={Li, Han and Shi, Yuling and Lin, Shaoxin and Gu, Xiaodong and Lian, Heng and Wang, Xin and Jia, Yantao and Huang, Tao and Wang, Qianxiang},
  journal={arXiv preprint arXiv:2507.23348},
  year={2025}
}

@inproceedings{bouzenia2025repairagent,
  title={Repairagent: An autonomous, llm-based agent for program repair},
  author={Bouzenia, Islem and Devanbu, Premkumar and Pradel, Michael},
  booktitle={2025 IEEE/ACM 47th International Conference on Software Engineering (ICSE)},
  pages={2188--2200},
  year={2025},
  organization={IEEE}
}

@inproceedings{zhang2025coderag,
  title={CodeRAG: Finding Relevant and Necessary Knowledge for Retrieval-Augmented Repository-Level Code Completion},
  author={Zhang, Sheng and Ding, Yifan and Lian, Shuquan and Song, Shun and Li, Hui},
  booktitle={Proceedings of the 2025 Conference on Empirical Methods in Natural Language Processing},
  pages={23289--23299},
  year={2025}
}

@inproceedings{phan2025repohyper,
  title={Repohyper: Search-expand-refine on semantic graphs for repository-level code completion},
  author={Phan, Huy N and Phan, Hoang N and Nguyen, Tien N and Bui, Nghi DQ},
  booktitle={2025 IEEE/ACM Second International Conference on AI Foundation Models and Software Engineering (Forge)},
  pages={14--25},
  year={2025},
  organization={IEEE}
}

@article{gaurav2025codemem,
  title={CodeMem: Architecting Reproducible Agents via Dynamic MCP and Procedural Memory},
  author={Gaurav, Nishant and Akarsh, Adit and Ravishankar, Tejas and Bajaj, Manoj},
  journal={arXiv preprint arXiv:2512.15813},
  year={2025}
}

@article{wu2025human,
  title={From human memory to ai memory: A survey on memory mechanisms in the era of llms},
  author={Wu, Yaxiong and Liang, Sheng and Zhang, Chen and Wang, Yichao and Zhang, Yongyue and Guo, Huifeng and Tang, Ruiming and Liu, Yong},
  journal={arXiv preprint arXiv:2504.15965},
  year={2025}
}

@inproceedings{zhang2024autocoderover,
  title={Autocoderover: Autonomous program improvement},
  author={Zhang, Yuntong and Ruan, Haifeng and Fan, Zhiyu and Roychoudhury, Abhik},
  booktitle={Proceedings of the 33rd ACM SIGSOFT International Symposium on Software Testing and Analysis},
  pages={1592--1604},
  year={2024}
}

@inproceedings{zhang2024codeagent,
  title={Codeagent: Enhancing code generation with tool-integrated agent systems for real-world repo-level coding challenges},
  author={Zhang, Kechi and Li, Jia and Li, Ge and Shi, Xianjie and Jin, Zhi},
  booktitle={Proceedings of the 62nd Annual Meeting of the Association for Computational Linguistics (Volume 1: Long Papers)},
  pages={13643--13658},
  year={2024}
}

@article{wang2025mirix,
  title={Mirix: Multi-agent memory system for llm-based agents},
  author={Wang, Yu and Chen, Xi},
  journal={arXiv preprint arXiv:2507.07957},
  year={2025}
}

@article{chen2023gamegpt,
  title={Gamegpt: Multi-agent collaborative framework for game development},
  author={Chen, Dake and Wang, Hanbin and Huo, Yunhao and Li, Yuzhao and Zhang, Haoyang},
  journal={arXiv preprint arXiv:2310.08067},
  year={2023}
}

@article{huang2023agentcoder,
  title={Agentcoder: Multi-agent-based code generation with iterative testing and optimisation},
  author={Huang, Dong and Zhang, Jie M and Luck, Michael and Bu, Qingwen and Qing, Yuhao and Cui, Heming},
  journal={arXiv preprint arXiv:2312.13010},
  year={2023}
}

@inproceedings{
jimenez2024swebench,
title={{SWE}-bench: Can Language Models Resolve Real-world Github Issues?},
author={Carlos E Jimenez and John Yang and Alexander Wettig and Shunyu Yao and Kexin Pei and Ofir Press and Karthik R Narasimhan},
booktitle={The Twelfth International Conference on Learning Representations},
year={2024}
}

@article{feng2026longcli,
  title={LongCLI-Bench: A Preliminary Benchmark and Study for Long-horizon Agentic Programming in Command-Line Interfaces},
  author={Feng, Yukang and Sun, Jianwen and Yang, Zelai and Ai, Jiaxin and Li, Chuanhao and Li, Zizhen and Zhang, Fanrui and He, Kang and Ma, Rui and Lin, Jifan and others},
  journal={arXiv preprint arXiv:2602.14337},
  year={2026}
}

@article{Dong2024SelfCollaboration,
  author = {Dong, Y. and Jiang, X. and Jin, Z. and Li, G.},
  title = {Self-collaboration code generation via {ChatGPT}},
  journal = {ACM Transactions on Software Engineering and Methodology},
  volume = {33},
  number = {7},
  pages = {1--38},
  year = {2024}
}

@article{wang2026memgovern,
  title={MemGovern: Enhancing Code Agents through Learning from Governed Human Experiences},
  author={Wang, Qihao and Cheng, Ziming and Zhang, Shuo and Liu, Fan and Xu, Rui and Lian, Heng and Wang, Kunyi and Yu, Xiaoming and Yin, Jianghao and Hu, Sen and others},
  journal={arXiv preprint arXiv:2601.06789},
  year={2026}
}

@article{Rasheed2024Codepori,
  author = {Rasheed, Z. and Sami, M. A. and Kemell, K.-K. and Waseem, M. and Saari, M. and Systa, K. and Abrahamsson, P.},
  title = {{Codepori}: Large-scale system for autonomous software development using multi-agent technology},
  journal = {arXiv preprint arXiv:2402.01411},
  year = {2024}
}

@inproceedings{Tao2024Magis,
  author = {Tao, W. and Zhou, Y. and Wang, Y. and Zhang, W. and Zhang, H. and Cheng, Y.},
  title = {{Magis}: {LLM}-based multi-agent framework for {GitHub} issue resolution},
  booktitle = {Advances in Neural Information Processing Systems (NeurIPS)},
  volume = {37},
  pages = {51963--51993},
  year = {2024}
}

@article{biswal2026agentsm,
  title={AgentSM: Semantic Memory for Agentic Text-to-SQL},
  author={Biswal, Asim and Lei, Chuan and Qin, Xiao and Li, Aodong and Narayanaswamy, Balakrishnan and Kraska, Tim},
  journal={arXiv preprint arXiv:2601.15709},
  year={2026}
}

@inproceedings{zhang2023repocoder,
  title={Repocoder: Repository-level code completion through iterative retrieval and generation},
  author={Zhang, Fengji and Chen, Bei and Zhang, Yue and Keung, Jacky and Liu, Jin and Zan, Daoguang and Mao, Yi and Lou, Jian-Guang and Chen, Weizhu},
  booktitle={Proceedings of the 2023 Conference on Empirical Methods in Natural Language Processing},
  pages={2471--2484},
  year={2023}
}

@article{wang2026memex,
  title={Memex (RL): Scaling Long-Horizon LLM Agents via Indexed Experience Memory},
  author={Wang, Zhenting and Chen, Huancheng and Wang, Jiayun and Wei, Wei},
  journal={arXiv preprint arXiv:2603.04257},
  year={2026}
}

@article{packer2023memgpt,
  title={MemGPT: Towards LLMs as Operating Systems},
  author={Packer, Charles and Wooders, Sarah and Lin, Kevin and Fang, Vivian and Patil, Shishir G and Stoica, Ion and Gonzalez, Joseph E},
  journal={arXiv preprint arXiv:2310.08560},
  year={2023}
}

@article{chu2024leveraging,
  title={Leveraging prior experience: An expandable auxiliary knowledge base for text-to-sql},
  author={Chu, Zhibo and Wang, Zichong and Qin, Qitao},
  journal={arXiv preprint arXiv:2411.13244},
  year={2024}
}

@article{Phan2024HyperAgent,
  author = {Phan, H. N. and Nguyen, T. N. and Nguyen, P. X. and Bui, N. D.},
  title = {{HyperAgent}: Generalist software engineering agents to solve coding tasks at scale},
  journal = {arXiv preprint arXiv:2409.16299},
  year = {2024}
}

@inproceedings{Zhang2024PairProgramming,
  author = {Zhang, H. and Cheng, W. and Wu, Y. and Hu, W.},
  title = {A pair programming framework for code generation via multi-plan exploration and feedback-driven refinement},
  booktitle = {Proceedings of the 39th IEEE/ACM International Conference on Automated Software Engineering (ASE)},
  pages = {1319--1331},
  year = {2024}
}

@inproceedings{Lin2025Soen101,
  author = {Lin, F. and Kim, D. J. and others},
  title = {{Soen-101}: Code generation by emulating software process models using large language model agents},
  booktitle = {Proceedings of the 47th International Conference on Software Engineering (ICSE)},
  pages = {1527--1539},
  year = {2025}
}

@article{Ishibashi2024SelfOrganized,
  author = {Ishibashi, Y. and Nishimura, Y.},
  title = {Self-organized agents: A {LLM} multi-agent framework toward ultra large-scale code generation and optimization},
  journal = {arXiv preprint arXiv:2404.02183},
  year = {2024}
}

@article{Zhao2024MAGE,
  author = {Zhao, Y. and Zhang, H. and Huang, H. and Yu, Z. and Zhao, J.},
  title = {{MAGE}: A multi-agent engine for automated {RTL} code generation},
  journal = {arXiv preprint arXiv:2412.07822},
  year = {2024}
}

@article{Nunez2024AutoSafeCoder,
  author = {Nunez, A. and Islam, N. T. and Jha, S. K. and Najafirad, P.},
  title = {{AutoSafeCoder}: A multi-agent framework for securing {LLM} code generation through static analysis and fuzz testing},
  journal = {arXiv preprint arXiv:2409.10737},
  year = {2024}
}

@article{Hu2025QualityFlow,
  author = {Hu, Y. and Zhou, Q. and Chen, Q. and Li, X. and Liu, L. and Zhang, D. and Kachroo, A. and Oz, T. and Tripp, O.},
  title = {{QualityFlow}: An agentic workflow for program synthesis controlled by {LLM} quality checks},
  journal = {arXiv preprint arXiv:2501.17167},
  year = {2025}
}

@article{Pan2025CodeCoR,
  author = {Pan, R. and Zhang, H. and Liu, C.},
  title = {{CodeCoR}: An {LLM}-based self-reflective multi-agent framework for code generation},
  journal = {arXiv preprint arXiv:2501.07811},
  year = {2025}
}

@misc{Rahman2025MACRO,
      title={MARCO: Multi-Agent Code Optimization with Real-Time Knowledge Integration for High-Performance Computing}, 
      author={Asif Rahman and Veljko Cvetkovic and Kathleen Reece and Aidan Walters and Yasir Hassan and Aneesh Tummeti and Bryan Torres and Denise Cooney and Margaret Ellis and Dimitrios S. Nikolopoulos},
      year={2025},
      eprint={2505.03906},
      archivePrefix={arXiv},
      primaryClass={cs.DC},
      url={https://arxiv.org/abs/2505.03906}
}

@article{Liu2025SEW,
  author = {Liu, S. and Fang, J. and Zhou, H. and Wang, Y. and Meng, Z.},
  title = {{SEW}: Self-evolving agentic workflows for automated code generation},
  journal = {arXiv preprint arXiv:2505.18646},
  year = {2025}
}

@inproceedings{Hu2025EvoMAC,
  author = {Hu, Y. and Cai, Y. and Du, Y. and Zhu, X. and Liu, X. and Yu, Z. and Hou, Y. and Tang, S. and Chen, S.},
  title = {Self-evolving multi-agent collaboration networks for software development},
  booktitle = {International Conference on Learning Representations (ICLR)},
  year = {2025}
}

@inproceedings{Qian2023ChatDev,
  author       = {Chen Qian and
                  Wei Liu and
                  Hongzhang Liu and
                  Nuo Chen and
                  Yufan Dang and
                  Jiahao Li and
                  Cheng Yang and
                  Weize Chen and
                  Yusheng Su and
                  Xin Cong and
                  Juyuan Xu and
                  Dahai Li and
                  Zhiyuan Liu and
                  Maosong Sun},
  editor       = {Lun{-}Wei Ku and
                  Andre Martins and
                  Vivek Srikumar},
  title        = {ChatDev: Communicative Agents for Software Development},
  booktitle    = {Proceedings of the 62nd Annual Meeting of the Association for Computational
                  Linguistics (Volume 1: Long Papers), {ACL} 2024, Bangkok, Thailand,
                  August 11-16, 2024},
  pages        = {15174--15186},
  publisher    = {Association for Computational Linguistics},
  year         = {2024},
  url          = {https://doi.org/10.18653/v1/2024.acl-long.810},
  doi          = {10.18653/V1/2024.ACL-LONG.810},
  bibsource    = {dblp computer science bibliography, https://dblp.org}
}

@inproceedings{Hong2023MetaGPT,
  author       = {Sirui Hong and
                  Mingchen Zhuge and
                  Jonathan Chen and
                  Xiawu Zheng and
                  Yuheng Cheng and
                  Jinlin Wang and
                  Ceyao Zhang and
                  Zili Wang and
                  Steven Ka Shing Yau and
                  Zijuan Lin and
                  Liyang Zhou and
                  Chenyu Ran and
                  Lingfeng Xiao and
                  Chenglin Wu and
                  J{\"{u}}rgen Schmidhuber},
  title        = {MetaGPT: Meta Programming for {A} Multi-Agent Collaborative Framework},
  booktitle    = {The Twelfth International Conference on Learning Representations,
                  {ICLR} 2024, Vienna, Austria, May 7-11, 2024},
  publisher    = {OpenReview.net},
  year         = {2024},
  url          = {https://openreview.net/forum?id=VtmBAGCN7o},
  bibsource    = {dblp computer science bibliography, https://dblp.org}
}

@article{Holt2023L2MAC,
  author = {Holt, S. and Luyten, M. R. and van der Schaar, M.},
  title = {{L2MAC}: Large language model automatic computer for extensive code generation},
  journal = {arXiv preprint arXiv:2310.02003},
  year = {2023}
}

@article{Li2025Cogito,
  author = {Li, Y. and Li, J. and Wang, Q. and Yang, M. and Kong, H. and Wang, S.},
  title = {Cogito, ergo sum: A neurobiologically-inspired cognition-memory-growth system for code generation},
  journal = {arXiv preprint arXiv:2501.18653},
  year = {2025}
}

@article{Qi2024CleanAgent,
  author = {Qi, D. and Miao, Z. and Wang, J.},
  title = {{CleanAgent}: Automating data standardization with {LLM}-based agents},
  journal = {arXiv preprint arXiv:2403.08291},
  year = {2024}
}

@article{Ma2024Lingma,
  author       = {Yingwei Ma and
                  Rongyu Cao and
                  Yongchang Cao and
                  Yue Zhang and
                  Jue Chen and
                  Yibo Liu and
                  Yuchen Liu and
                  Binhua Li and
                  Fei Huang and
                  Yongbin Li},
  title        = {Lingma {SWE-GPT:} An Open Development-Process-Centric Language Model
                  for Automated Software Improvement},
  journal      = {CoRR},
  volume       = {abs/2411.00622},
  year         = {2024},
  url          = {https://doi.org/10.48550/arXiv.2411.00622},
  doi          = {10.48550/ARXIV.2411.00622},
  eprinttype   = {arXiv},
  eprint       = {2411.00622},
  bibsource    = {dblp computer science bibliography, https://dblp.org}
}

@inproceedings{Guo2025SyncMind,
  author = {Guo, X. and Wang, X. and Chen, Y. and Li, S. and Han, C. and Li, M. and Ji, H.},
  title = {{SyncMind}: Measuring agent out-of-sync recovery in collaborative software engineering},
  booktitle = {International Conference on Machine Learning (ICML)},
  year = {2025}
}

@article{song2026envscaler,
  title={Envscaler: Scaling tool-interactive environments for llm agent via programmatic synthesis},
  author={Song, Xiaoshuai and Chang, Haofei and Dong, Guanting and Zhu, Yutao and Wen, Ji-Rong and Dou, Zhicheng},
  journal={arXiv preprint arXiv:2601.05808},
  year={2026}
}

@article{gao2026teaching,
  title={Teaching LLMs to Learn Tool Trialing and Execution through Environment Interaction},
  author={Gao, Xingjie and Huang, Pengcheng and Liu, Zhenghao and Yan, Yukun and Wang, Shuo and Chen, Zulong and Qian, Chen and Yu, Ge and Gu, Yu},
  journal={arXiv preprint arXiv:2601.12762},
  year={2026}
}

@inproceedings{xi2025agentgym,
  title={Agentgym: Evaluating and training large language model-based agents across diverse environments},
  author={Xi, Zhiheng and Ding, Yiwen and Chen, Wenxiang and Hong, Boyang and Guo, Honglin and Wang, Junzhe and Guo, Xin and Yang, Dingwen and Liao, Chenyang and He, Wei and others},
  booktitle={Proceedings of the 63rd Annual Meeting of the Association for Computational Linguistics (Volume 1: Long Papers)},
  pages={27914--27961},
  year={2025}
}

@article{liu2026dive,
  title={Dive into Claude Code: The Design Space of Today's and Future AI Agent Systems},
  author={Liu, Jiacheng and Zhao, Xiaohan and Shang, Xinyi and Shen, Zhiqiang},
  journal={arXiv preprint arXiv:2604.14228},
  year={2026}
}

@article{liang2026generalizable,
  title={Generalizable Self-Evolving Memory for Automatic Prompt Optimization},
  author={Liang, Guanbao and Bei, Yuanchen and Zhou, Sheng and Qin, Yuheng and Zhou, Huan and Jia, Bingxin and Li, Bin and Bu, Jiajun},
  journal={arXiv preprint arXiv:2603.21520},
  year={2026}
}

@inproceedings{shi2025longcodezip,
  title={LongCodeZip: Compress Long Context for Code Language Models},
  author={Shi, Yuling and Qian, Yichun and Zhang, Hongyu and Shen, Beijun and Gu, Xiaodong},
  booktitle={2025 40th IEEE/ACM International Conference on Automated Software Engineering (ASE)},
  pages={141--153},
  year={2025},
  organization={IEEE}
}

@article{wang2026swe,
  title={SWE-Pruner: Self-Adaptive Context Pruning for Coding Agents},
  author={Wang, Yuhang and Shi, Yuling and Yang, Mo and Zhang, Rongrui and He, Shilin and Lian, Heng and Chen, Yuting and Ye, Siyu and Cai, Kai and Gu, Xiaodong},
  journal={arXiv preprint arXiv:2601.16746},
  year={2026}
}

@article{driess2023palme,
  title={Palm-e: An embodied multimodal language model},
  author={Driess, Danny and Xia, Fei and Sajjadi, Mehdi SM and Lynch, Corey and Chowdhery, Aakanksha and Ichter, Brian and Wahid, Ayzaan and Tompson, Jonathan and Vuong, Quan and Yu, Tianhe and others},
  journal={arXiv preprint arXiv:2303.03378},
  year={2023}
}

@article{deepmind2025geminirobotics,
  title={Gemini robotics 1.5: Pushing the frontier of generalist robots with advanced embodied reasoning, thinking, and motion transfer},
  author={Team, Gemini Robotics and Abdolmaleki, Abbas and Abeyruwan, Saminda and Ainslie, Joshua and Alayrac, Jean-Baptiste and Arenas, Montserrat Gonzalez and Balakrishna, Ashwin and Batchelor, Nathan and Bewley, Alex and Bingham, Jeff and others},
  journal={arXiv preprint arXiv:2510.03342},
  year={2025}
}

@inproceedings{macenski2020nav2,
  title     = {The Marathon 2: A Navigation System},
  author    = {Macenski, Steve and Mart{\'i}n, Francisco and White, Ruffin and Gin{\'e}s Clavero, Jonatan},
  booktitle = {2020 IEEE/RSJ International Conference on Intelligent Robots and Systems},
  year      = {2020},
  organization = {IEEE},
  url       = {https://arxiv.org/abs/2003.00368}
}

@article{wang2025ui,
  title={Ui-tars-2 technical report: Advancing gui agent with multi-turn reinforcement learning},
  author={Wang, Haoming and Zou, Haoyang and Song, Huatong and Feng, Jiazhan and Fang, Junjie and Lu, Junting and Liu, Longxiang and Luo, Qinyu and Liang, Shihao and Huang, Shijue and others},
  journal={arXiv preprint arXiv:2509.02544},
  year={2025}
}

@inproceedings{liu2026llm,
  title={LLM-Assisted Circuit Verification: A Comprehensive Survey},
  author={Liu, Hongduo and Lu, Yuntao and Wang, Mingjun and Yao, Xufeng and Yu, Bei},
  booktitle={2026 31st Asia and South Pacific Design Automation Conference (ASP-DAC)},
  pages={439--446},
  year={2026},
  organization={IEEE}
}

@article{xiong2025self,
  title={Self-organizing agent network for llm-based workflow automation},
  author={Xiong, Yiming and Wang, Jian and Li, Bing and Zhu, Yuhan and Zhao, Yuqi},
  journal={arXiv preprint arXiv:2508.13732},
  year={2025}
}

@inproceedings{shi2025flowxpert,
  title={FlowXpert: Expertizing Troubleshooting Workflow Orchestration with Knowledge Base and Multi-Agent Coevolution},
  author={Shi, Binpeng and Luo, Yu and Wang, Jingya and Zhao, Yongxin and Zhang, Shenglin and Hao, Bowen and Zhao, Chenyu and Sun, Yongqian and Zhang, Zhi and Sun, Ronghua and others},
  booktitle={Proceedings of the 31st ACM SIGKDD Conference on Knowledge Discovery and Data Mining V. 2},
  pages={4839--4850},
  year={2025}
}

@article{jin2025reveal,
  title={ReVeal: Self-Evolving Code Agents via Reliable Self-Verification},
  author={Jin, Yiyang and Xu, Kunzhao and Li, Hang and Han, Xueting and Zhou, Yanmin and Li, Cheng and Bai, Jing},
  journal={arXiv preprint arXiv:2506.11442},
  year={2025}
}

@article{Xu2025Hallucination,
  author = {Xu, Q. and Wang, G. and Briand, L. and Liu, K.},
  title = {Hallucination to Consensus: Multi-Agent LLMs for End-to-End JUnit Test Generation},
  journal = {arXiv preprint arXiv:2506.02943},
  year = {2025}
}

@article{erman1980hearsay,
  author    = {Erman, Lee D. and Hayes-Roth, Frederick and Lesser, Victor R. and Reddy, D. Raj},
  title     = {The Hearsay-II Speech-Understanding System: Integrating Knowledge to Resolve Uncertainty},
  journal   = {ACM Computing Surveys (CSUR)},
  volume    = {12},
  number    = {2},
  pages     = {213--253},
  year      = {1980},
  publisher = {ACM New York, NY, USA},
  doi       = {10.1145/356810.356816}
}

@article{gao2025flowreasoner,
  title={FlowReasoner: Reinforcing Query-Level Meta-Agents},
  author={Gao, Hongcheng and Liu, Yue and He, Yufei and Dou, Longxu and Du, Chao and Deng, Zhijie and Hooi, Bryan and Lin, Min and Pang, Tianyu},
  journal={arXiv preprint arXiv:2504.15257},
  year={2025}
}

@article{gao2025traeagent,
  title={Trae Agent: An LLM-based Agent for Software Engineering with Test-time Scaling},
  author={Gao, Pengfei and Tian, Zhao and Meng, Xiangxin and Wang, Xinchen and Hu, Ruida and Xiao, Yuanan and Liu, Yizhou and Zhang, Zhao and Chen, Junjie and Gao, Cuiyun and Lin, Yun and Xiong, Yingfei and Peng, Chao and Liu, Xia},
  journal={arXiv preprint arXiv:2507.23370},
  year={2025}
}

@article{xu2025boad,
  title={BOAD: Discovering Hierarchical Software Engineering Agents via Bandit Optimization},
  author={Xu, Iris and Zeng, Guangtao and He, Zexue and Jin, Charles and Pareja, Aldo and Gutfreund, Dan and Gan, Chuang and Hong, Zhang-Wei},
  journal={arXiv preprint arXiv:2512.23631},
  year={2025}
}

@article{sapkota2025vibe,
  title={Vibe coding vs. agentic coding: Fundamentals and practical implications of agentic ai},
  author={Sapkota, Ranjan and Roumeliotis, Konstantinos I and Karkee, Manoj},
  journal={arXiv preprint arXiv:2505.19443},
  year={2025}
}

@article{watanabe2025use,
  title={On the use of agentic coding: An empirical study of pull requests on github},
  author={Watanabe, Miku and Li, Hao and Kashiwa, Yutaro and Reid, Brittany and Iida, Hajimu and Hassan, Ahmed E},
  journal={ACM Transactions on Software Engineering and Methodology},
  year={2025},
  publisher={ACM New York, NY}
}

@article{zhang2023toolcoder,
  title={Toolcoder: Teach code generation models to use api search tools},
  author={Zhang, Kechi and Zhang, Huangzhao and Li, Ge and Li, Jia and Li, Zhuo and Jin, Zhi},
  journal={arXiv preprint arXiv:2305.04032},
  year={2023}
}

@inproceedings{ahmed2024codeqa,
  title={CodeQA: Advanced programming question-answering using LLM agent and RAG},
  author={Ahmed, Mohamed and Dorrah, Mostafa and Ashraf, Ahmed and Adel, Yousef and Elatrozy, Abdelrahman and Mohamed, Bahaa Eldin and Gomaa, Wael},
  booktitle={2024 6th Novel Intelligent and Leading Emerging Sciences Conference (NILES)},
  pages={494--499},
  year={2024},
  organization={IEEE}
}

@article{zhao2025rag,
  title={RAG-Based AI Agents for Enterprise Software Development: Implementation Patterns and Production Deployment},
  author={Zhao, Xiuyuan and Sun, Tiejiang and Ren, Shaochen and Yang, Jingyun and Liu, Yang},
  journal={Frontiers in Artificial Intelligence Research},
  volume={2},
  number={3},
  pages={501--520},
  year={2025}
}

@inproceedings{zhou2023devil,
  title={The devil is in the tails: How long-tailed code distributions impact large language models},
  author={Zhou, Xin and Kim, Kisub and Xu, Bowen and Liu, Jiakun and Han, DongGyun and Lo, David},
  booktitle={2023 38th IEEE/ACM International Conference on Automated Software Engineering (ASE)},
  pages={40--52},
  year={2023},
  organization={IEEE}
}

@article{li2026environment,
  title={Environment-in-the-Loop: Rethinking Code Migration with LLM-based Agents},
  author={Li, Xiang and Fei, Zhiwei and Ma, Ying and Zhang, Jerry and Federica, Sarro and Ye, He},
  journal={arXiv preprint arXiv:2602.09944},
  year={2026}
}

@inproceedings{
chen2026grounded,
title={Grounded Test-Time Adaptation for {LLM} Agents},
author={Arthur Chen and Zuxin Liu and Jianguo Zhang and Akshara Prabhakar and Zhiwei Liu and Shelby Heinecke and Silvio Savarese and Victor Zhong and Caiming Xiong},
booktitle={The Fourteenth International Conference on Learning Representations},
year={2026}
}

@article{miculicich2025veriguard,
  title={Veriguard: Enhancing llm agent safety via verified code generation},
  author={Miculicich, Lesly and Parmar, Mihir and Palangi, Hamid and Dvijotham, Krishnamurthy Dj and Montanari, Mirko and Pfister, Tomas and Le, Long T},
  journal={arXiv preprint arXiv:2510.05156},
  year={2025}
}

@article{liu2024toolnet,
  title={Toolnet: Connecting large language models with massive tools via tool graph},
  author={Liu, Xukun and Peng, Zhiyuan and Yi, Xiaoyuan and Xie, Xing and Xiang, Lirong and Liu, Yuchen and Xu, Dongkuan},
  journal={arXiv preprint arXiv:2403.00839},
  year={2024}
}

@inproceedings{liu2024controlllm,
  title={Controlllm: Augment language models with tools by searching on graphs},
  author={Liu, Zhaoyang and Lai, Zeqiang and Gao, Zhangwei and Cui, Erfei and Li, Ziheng and Zhu, Xizhou and Lu, Lewei and Chen, Qifeng and Qiao, Yu and Dai, Jifeng and others},
  booktitle={European Conference on Computer Vision},
  pages={89--105},
  year={2024},
  organization={Springer}
}

@article{chen2021evaluating,
  title={Evaluating large language models trained on code},
  author={Chen, Mark and Tworek, Jerry and Jun, Heewoo and Yuan, Qiming and Pinto, Henrique Ponde De Oliveira and Kaplan, Jared and Edwards, Harri and Burda, Yuri and Joseph, Nicholas and Brockman, Greg and others},
  journal={arXiv preprint arXiv:2107.03374},
  year={2021}
}

@article{austin2021program,
  title={Program synthesis with large language models},
  author={Austin, Jacob and Odena, Augustus and Nye, Maxwell and Bosma, Maarten and Michalewski, Henryk and Dohan, David and Jiang, Ellen and Cai, Carrie and Terry, Michael and Le, Quoc and others},
  journal={arXiv preprint arXiv:2108.07732},
  year={2021}
}

@article{nijkamp2022codegen,
  title={Codegen: An open large language model for code with multi-turn program synthesis},
  author={Nijkamp, Erik and Pang, Bo and Hayashi, Hiroaki and Tu, Lifu and Wang, Huan and Zhou, Yingbo and Savarese, Silvio and Xiong, Caiming},
  journal={arXiv preprint arXiv:2203.13474},
  year={2022}
}

@article{li2022competition,
  title={Competition-level code generation with alphacode},
  author={Li, Yujia and Choi, David and Chung, Junyoung and Kushman, Nate and Schrittwieser, Julian and Leblond, R{\'e}mi and Eccles, Tom and Keeling, James and Gimeno, Felix and Dal Lago, Agustin and others},
  journal={Science},
  volume={378},
  number={6624},
  pages={1092--1097},
  year={2022},
  publisher={American Association for the Advancement of Science}
}

@inproceedings{bi2024program,
  title={When do program-of-thought works for reasoning?},
  author={Bi, Zhen and Zhang, Ningyu and Jiang, Yinuo and Deng, Shumin and Zheng, Guozhou and Chen, Huajun},
  booktitle={Proceedings of the AAAI conference on artificial intelligence},
  volume={38},
  number={16},
  pages={17691--17699},
  year={2024}
}

@article{jia2026compressing,
  title={Compressing Code Context for LLM-based Issue Resolution},
  author={Jia, Haoxiang and Barr, Earl T and Mechtaev, Sergey},
  journal={arXiv preprint arXiv:2603.28119},
  year={2026}
}

@article{sun2025scaling,
  title={Scaling long-horizon llm agent via context-folding},
  author={Sun, Weiwei and Lu, Miao and Ling, Zhan and Liu, Kang and Yao, Xuesong and Yang, Yiming and Chen, Jiecao},
  journal={arXiv preprint arXiv:2510.11967},
  year={2025}
}

@article{zhang2025code,
  title={Code-enabled language models can outperform reasoning models on diverse tasks},
  author={Zhang, Cedegao E and Colas, C{\'e}dric and Poesia, Gabriel and Tenenbaum, Joshua B and Andreas, Jacob},
  journal={arXiv preprint arXiv:2510.20909},
  year={2025}
}

@article{yu2024reasoning,
  title={Reasoning through execution: Unifying process and outcome rewards for code generation},
  author={Yu, Zhuohao and Gu, Weizheng and Wang, Yidong and Jiang, Xingru and Zeng, Zhengran and Wang, Jindong and Ye, Wei and Zhang, Shikun},
  journal={arXiv preprint arXiv:2412.15118},
  year={2024}
}

@inproceedings{li2025codeprm,
  title={Codeprm: Execution feedback-enhanced process reward model for code generation},
  author={Li, Qingyao and Dai, Xinyi and Li, Xiangyang and Zhang, Weinan and Wang, Yasheng and Tang, Ruiming and Yu, Yong},
  booktitle={Findings of the Association for Computational Linguistics: ACL 2025},
  pages={8169--8182},
  year={2025}
}

@article{le2022coderl,
  title={Coderl: Mastering code generation through pretrained models and deep reinforcement learning},
  author={Le, Hung and Wang, Yue and Gotmare, Akhilesh Deepak and Savarese, Silvio and Hoi, Steven Chu Hong},
  journal={Advances in Neural Information Processing Systems},
  volume={35},
  pages={21314--21328},
  year={2022}
}

@article{yu2026reinforcement,
  title={Reinforcement World Model Learning for LLM-based Agents},
  author={Yu, Xiao and Peng, Baolin and Xu, Ruize and Shen, Yelong and He, Pengcheng and Nath, Suman and Singh, Nikhil and Gao, Jiangfeng and Yu, Zhou},
  journal={arXiv preprint arXiv:2602.05842},
  year={2026}
}

@article{nye2021show,
  title={Show your work: Scratchpads for intermediate computation with language models},
  author={Nye, Maxwell and Andreassen, Anders Johan and Gur-Ari, Guy and Michalewski, Henryk and Austin, Jacob and Bieber, David and Dohan, David and Lewkowycz, Aitor and Bosma, Maarten and Luan, David and others},
  year={2021}
}

@inproceedings{pi2022reasoning,
  title={Reasoning like program executors},
  author={Pi, Xinyu and Liu, Qian and Chen, Bei and Ziyadi, Morteza and Lin, Zeqi and Fu, Qiang and Gao, Yan and Lou, Jian-Guang and Chen, Weizhu},
  booktitle={Proceedings of the 2022 conference on empirical methods in natural language processing},
  pages={761--779},
  year={2022}
}

@inproceedings{payoungkhamdee2025towards,
  title={Towards better understanding of program-of-thought reasoning in cross-lingual and multilingual environments},
  author={Payoungkhamdee, Patomporn and Tuchinda, Pume and Baek, Jinheon and Cahyawijaya, Samuel and Udomcharoenchaikit, Can and Manakul, Potsawee and Limkonchotiwat, Peerat and Chuangsuwanich, Ekapol and Nutanong, Sarana},
  booktitle={Findings of the Association for Computational Linguistics: ACL 2025},
  pages={15810--15828},
  year={2025}
}

@article{su2025method,
  title={Method-based reasoning for large language models: Extraction, reuse, and continuous improvement},
  author={Su, Hong},
  journal={arXiv preprint arXiv:2508.04289},
  year={2025}
}

@inproceedings{besta2024graph,
  title={Graph of thoughts: Solving elaborate problems with large language models},
  author={Besta, Maciej and Blach, Nils and Kubicek, Ales and Gerstenberger, Robert and Podstawski, Michal and Gianinazzi, Lukas and Gajda, Joanna and Lehmann, Tomasz and Niewiadomski, Hubert and Nyczyk, Piotr and others},
  booktitle={Proceedings of the AAAI conference on artificial intelligence},
  volume={38},
  number={16},
  pages={17682--17690},
  year={2024}
}

@article{shi2025ssr,
  title={SSR: Socratic Self-Refine for Large Language Model Reasoning},
  author={Shi, Haizhou and Liu, Ye and Pang, Bo and Liu, Zeyu Leo and Wang, Hao and Savarese, Silvio and Xiong, Caiming and Zhou, Yingbo and Yavuz, Semih},
  journal={arXiv preprint arXiv:2511.10621},
  year={2025}
}

@article{yu2025self,
  title={Self-Verifying Reflection Helps Transformers with CoT Reasoning},
  author={Yu, Zhongwei and Xia, Wannian and Yan, Xue and Xu, Bo and Zhang, Haifeng and Du, Yali and Wang, Jun},
  journal={arXiv preprint arXiv:2510.12157},
  year={2025}
}

@article{chen2025codesteer,
  title={CodeSteer: Symbolic-Augmented Language Models via Code/Text Guidance},
  author={Chen, Yongchao and Hao, Yilun and Liu, Yueying and Zhang, Yang and Fan, Chuchu},
  journal={arXiv preprint arXiv:2502.04350},
  year={2025}
}

@inproceedings{chen2025code,
  title={Code-as-symbolic-planner: Foundation model-based robot planning via symbolic code generation},
  author={Chen, Yongchao and Hao, Yilun and Zhang, Yang and Fan, Chuchu},
  booktitle={2025 IEEE/RSJ International Conference on Intelligent Robots and Systems (IROS)},
  pages={19248--19254},
  year={2025},
  organization={IEEE}
}

@article{ni2024next,
  title={Next: Teaching large language models to reason about code execution},
  author={Ni, Ansong and Allamanis, Miltiadis and Cohan, Arman and Deng, Yinlin and Shi, Kensen and Sutton, Charles and Yin, Pengcheng},
  journal={arXiv preprint arXiv:2404.14662},
  year={2024}
}

@article{ding2024cycle,
  title={Cycle: Learning to self-refine the code generation},
  author={Ding, Yangruibo and Min, Marcus J and Kaiser, Gail and Ray, Baishakhi},
  journal={Proceedings of the ACM on Programming Languages},
  volume={8},
  number={OOPSLA1},
  pages={392--418},
  year={2024},
  publisher={ACM New York, NY, USA}
}

@inproceedings{zhang2023self,
  title={Self-edit: Fault-aware code editor for code generation},
  author={Zhang, Kechi and Li, Zhuo and Li, Jia and Li, Ge and Jin, Zhi},
  booktitle={Proceedings of the 61st Annual Meeting of the Association for Computational Linguistics (Volume 1: Long Papers)},
  pages={769--787},
  year={2023}
}

@article{jiang2025coderl+,
  title={CodeRL+: Improving Code Generation via Reinforcement with Execution Semantics Alignment},
  author={Jiang, Xue and Dong, Yihong and Liu, Mengyang and Deng, Hongyi and Wang, Tian and Tao, Yongding and Cao, Rongyu and Li, Binhua and Jin, Zhi and Jiao, Wenpin and others},
  journal={arXiv preprint arXiv:2510.18471},
  year={2025}
}

@article{liu2023rltf,
  title={Rltf: Reinforcement learning from unit test feedback},
  author={Liu, Jiate and Zhu, Yiqin and Xiao, Kaiwen and Fu, Qiang and Han, Xiao and Yang, Wei and Ye, Deheng},
  journal={arXiv preprint arXiv:2307.04349},
  year={2023}
}

@inproceedings{dou2024stepcoder,
  title={Stepcoder: improving code generation with reinforcement learning from compiler feedback},
  author={Dou, Shihan and Liu, Yan and Jia, Haoxiang and Zhou, Enyu and Xiong, Limao and Shan, Junjie and Huang, Caishuang and Wang, Xiao and Fan, Xiaoran and Xi, Zhiheng and others},
  booktitle={Proceedings of the 62nd Annual Meeting of the Association for Computational Linguistics (Volume 1: Long Papers)},
  pages={4571--4585},
  year={2024}
}

@article{gehring2024rlef,
  title={Rlef: Grounding code llms in execution feedback with reinforcement learning},
  author={Gehring, Jonas and Zheng, Kunhao and Copet, Jade and Mella, Vegard and Carbonneaux, Quentin and Cohen, Taco and Synnaeve, Gabriel},
  journal={arXiv preprint arXiv:2410.02089},
  year={2024}
}

@article{lavon2025execution,
  title={Execution guided line-by-line code generation},
  author={Lavon, Boaz and Katz, Shahar and Wolf, Lior},
  journal={arXiv preprint arXiv:2506.10948},
  year={2025}
}

@article{chen2025r1,
  title={R1-code-interpreter: Training llms to reason with code via supervised and reinforcement learning},
  author={Chen, Yongchao and Liu, Yueying and Zhou, Junwei and Hao, Yilun and Wang, Jingquan and Zhang, Yang and Fan, Chuchu},
  journal={arXiv e-prints},
  pages={arXiv--2505},
  year={2025}
}

@article{ren2023robots,
  title={Robots that ask for help: Uncertainty alignment for large language model planners},
  author={Ren, Allen Z and Dixit, Anushri and Bodrova, Alexandra and Singh, Sumeet and Tu, Stephen and Brown, Noah and Xu, Peng and Takayama, Leila and Xia, Fei and Varley, Jake and others},
  journal={arXiv preprint arXiv:2307.01928},
  year={2023}
}

@article{zhang2023bootstrap,
  title={Bootstrap your own skills: Learning to solve new tasks with large language model guidance},
  author={Zhang, Jesse and Zhang, Jiahui and Pertsch, Karl and Liu, Ziyi and Ren, Xiang and Chang, Minsuk and Sun, Shao-Hua and Lim, Joseph J},
  journal={arXiv preprint arXiv:2310.10021},
  year={2023}
}

@inproceedings{ha2023scaling,
  title={Scaling up and distilling down: Language-guided robot skill acquisition},
  author={Ha, Huy and Florence, Pete and Song, Shuran},
  booktitle={Conference on Robot Learning},
  pages={3766--3777},
  year={2023},
  organization={PMLR}
}

@inproceedings{tziafas2024lifelong,
  title={Lifelong robot library learning: Bootstrapping composable and generalizable skills for embodied control with language models},
  author={Tziafas, Georgios and Kasaei, Hamidreza},
  booktitle={2024 IEEE International Conference on Robotics and Automation (ICRA)},
  pages={515--522},
  year={2024},
  organization={IEEE}
}

@article{zhai2026skillvla,
  title={SkillVLA: Tackling Combinatorial Diversity in Dual-Arm Manipulation via Skill Reuse},
  author={Zhai, Xuanran and Huang, Zekai and Wu, Longyan and Zhao, Qianyou and Yu, Qiaojun and Ren, Jieji and Hao, Ce and Soh, Harold},
  journal={arXiv preprint arXiv:2603.03836},
  year={2026}
}

@article{mu2024robocodex,
  title={Robocodex: Multimodal code generation for robotic behavior synthesis},
  author={Mu, Yao and Chen, Junting and Zhang, Qinglong and Chen, Shoufa and Yu, Qiaojun and Ge, Chongjian and Chen, Runjian and Liang, Zhixuan and Hu, Mengkang and Tao, Chaofan and others},
  journal={arXiv preprint arXiv:2402.16117},
  year={2024}
}

@inproceedings{zhang2025codebt,
  title={Code-BT: A Code-Driven Approach to Behavior Tree Generation for Robot Tasks Planning with Large Language Models},
  author={Zhang, Siyang and Li, Bin and Qi, Jingtao and Wang, Xueying and Li, Fu and Wang, Jianan and Zhu, En and Sun, Jinjing},
  booktitle={Proceedings of the Thirty-Fourth International Joint Conference on Artificial Intelligence},
  pages={8814--8822},
  year={2025}
}

@article{szeider2025cp,
  title={Cp-agent: Agentic constraint programming},
  author={Szeider, Stefan},
  journal={arXiv preprint arXiv:2508.07468},
  year={2025}
}

@article{wang2025llm,
  title={LLM-Driven Corrective Robot Operation Code Generation with Static Text-Based Simulation},
  author={Wang, Wenhao and Rong, Yi and Li, Yanyan and Jiao, Long and Yuan, Jiawei},
  journal={arXiv preprint arXiv:2512.02002},
  year={2025}
}

@article{ji2026genswarm,
  title={Genswarm: Scalable multi-robot code-policy generation and deployment via language models},
  author={Ji, Wenkang and Chen, Huaben and Chen, Mingyang and Zhu, Guobin and Xu, Lufeng and Gro{\ss}, Roderich and Zhou, Rui and Cao, Ming and Zhao, Shiyu},
  journal={npj Robotics},
  volume={4},
  number={1},
  pages={5},
  year={2026},
  publisher={Nature Publishing Group UK London}
}

@article{guan2025normcode,
  title={NormCode: A Semi-Formal Language for Auditable AI Planning},
  author={Guan, Xin and Li, Yunshan and Wu, Zekun and Zhang, Ruibo},
  journal={arXiv preprint arXiv:2512.10563},
  year={2025}
}

@article{santos2026alrm,
  title={ALRM: Agentic LLM for Robotic Manipulation},
  author={Santos, Vitor Gaboardi dos and Khadraoui, Ibrahim and Farhat, Ibrahim and Yous, Hamza and Teffahi, Samy and Hacid, Hakim},
  journal={arXiv preprint arXiv:2601.19510},
  year={2026}
}

@article{ashley2026racas,
  title={RACAS: Controlling Diverse Robots With a Single Agentic System},
  author={Ashley, Dylan R and Przepi{\'o}ra, Jan and Chen, Yimeng and Abualsaud, Ali and Yesmagambet, Nurzhan and Park, Shinkyu and Feron, Eric and Schmidhuber, J{\"u}rgen},
  journal={arXiv preprint arXiv:2603.05621},
  year={2026}
}

@article{meng2025growing,
  title={Growing with your embodied agent: A human-in-the-loop lifelong code generation framework for long-horizon manipulation skills},
  author={Meng, Yuan and Sun, Zhenguo and Fest, Max and Li, Xukun and Bing, Zhenshan and Knoll, Alois},
  journal={arXiv preprint arXiv:2509.18597},
  year={2025}
}

@article{kagaya2025vireskill,
  title={Vireskill: Vision-grounded replanning with skill memory for llm-based planning in lifelong robot learning},
  author={Kagaya, Tomoyuki and Lakshmi, Subramanian and Ye, Anbang and Yuan, Thong Jing and Karlekar, Jayashree and Pranata, Sugiri and Murakami, Natsuki and Kinose, Akira and You, Yang},
  journal={arXiv preprint arXiv:2509.24219},
  year={2025}
}

@inproceedings{wang2026lifelong,
  title={Lifelong Language-Conditioned Robotic Manipulation Learning},
  author={Wang, Xudong and Han, Zebin and Liu, Zhiyu and Li, Gan and Dong, Jiahua and Liu, Baichen and Liu, Lianqing and Han, Zhi},
  booktitle={Proceedings of the AAAI Conference on Artificial Intelligence},
  volume={40},
  number={22},
  pages={18629--18637},
  year={2026}
}

@article{lin2026ui,
  title={UI-Voyager: A Self-Evolving GUI Agent Learning via Failed Experience},
  author={Lin, Zichuan and Liu, Feiyu and Yang, Yijun and Lyu, Jiafei and Gao, Yiming and Liu, Yicheng and Lu, Zhicong and Yu, Yangbin and Yang, Mingyu and Li, Junyou and others},
  journal={arXiv preprint arXiv:2603.24533},
  year={2026}
}

@article{hsu2025programs,
  title={From Programs to Poses: Factored Real-World Scene Generation via Learned Program Libraries},
  author={Hsu, Joy and Jin, Emily and Wu, Jiajun and Mitra, Niloy J},
  journal={arXiv preprint arXiv:2510.10292},
  year={2025}
}

@article{piriyakulkij2025poe,
  title={Poe-world: Compositional world modeling with products of programmatic experts},
  author={Piriyakulkij, Wasu Top and Liang, Yichao and Tang, Hao and Weller, Adrian and Kryven, Marta and Ellis, Kevin},
  journal={arXiv preprint arXiv:2505.10819},
  year={2025}
}

@article{zheng2026code2world,
  title={Code2world: A gui world model via renderable code generation},
  author={Zheng, Yuhao and Zhong, Li'an and Wang, Yi and Dai, Rui and Liu, Kaikui and Chu, Xiangxiang and Lv, Linyuan and Torr, Philip and Lin, Kevin Qinghong},
  journal={arXiv preprint arXiv:2602.09856},
  year={2026}
}

@article{zhang2026code2worlds,
  title={Code2Worlds: Empowering Coding LLMs for 4D World Generation},
  author={Zhang, Yi and Wang, Yunshuang and Zhang, Zeyu and Tang, Hao},
  journal={arXiv preprint arXiv:2602.11757},
  year={2026}
}

@article{wang2026agent,
  title={Agent world model: Infinity synthetic environments for agentic reinforcement learning},
  author={Wang, Zhaoyang and Xu, Canwen and Liu, Boyi and Wang, Yite and Han, Siwei and Yao, Zhewei and Yao, Huaxiu and He, Yuxiong},
  journal={arXiv preprint arXiv:2602.10090},
  year={2026}
}

@article{copet2025cwm,
  title={Cwm: An open-weights llm for research on code generation with world models},
  author={Copet, Jade and Carbonneaux, Quentin and Cohen, Gal and Gehring, Jonas and Kahn, Jacob and Kossen, Jannik and Kreuk, Felix and McMilin, Emily and Meyer, Michel and Wei, Yuxiang and others},
  journal={arXiv preprint arXiv:2510.02387},
  year={2025}
}

@article{ren2026aligning,
  title={Aligning Agentic World Models via Knowledgeable Experience Learning},
  author={Ren, Baochang and Yao, Yunzhi and Sun, Rui and Qiao, Shuofei and Zhang, Ningyu and Chen, Huajun},
  journal={arXiv preprint arXiv:2601.13247},
  year={2026}
}

@article{gu2024cruxeval,
  title={Cruxeval: A benchmark for code reasoning, understanding and execution},
  author={Gu, Alex and Rozi{\`e}re, Baptiste and Leather, Hugh and Solar-Lezama, Armando and Synnaeve, Gabriel and Wang, Sida I},
  journal={arXiv preprint arXiv:2401.03065},
  year={2024}
}

@article{jain2024livecodebench,
  title={Livecodebench: Holistic and contamination free evaluation of large language models for code},
  author={Jain, Naman and Han, King and Gu, Alex and Li, Wen-Ding and Yan, Fanjia and Zhang, Tianjun and Wang, Sida and Solar-Lezama, Armando and Sen, Koushik and Stoica, Ion},
  journal={arXiv preprint arXiv:2403.07974},
  year={2024}
}

@inproceedings{xu2025cruxeval,
  title={Cruxeval-x: A benchmark for multilingual code reasoning, understanding and execution},
  author={Xu, Ruiyang and Cao, Jialun and Lu, Yaojie and Wen, Ming and Lin, Hongyu and Han, Xianpei and He, Ben and Cheung, Shing-Chi and Sun, Le},
  booktitle={Proceedings of the 63rd Annual Meeting of the Association for Computational Linguistics (Volume 1: Long Papers)},
  pages={23762--23779},
  year={2025}
}

@article{xie2025core,
  title={Core: Benchmarking llms code reasoning capabilities through static analysis tasks},
  author={Xie, Danning and Zheng, Mingwei and Liu, Xuwei and Wang, Jiannan and Wang, Chengpeng and Tan, Lin and Zhang, Xiangyu},
  journal={arXiv preprint arXiv:2507.05269},
  year={2025}
}

@article{luo2025geogrambench,
  title={Geogrambench: Benchmarking the geometric program reasoning in modern llms},
  author={Luo, Shixian and Zhu, Zezhou and Yuan, Yu and Yang, Yuncheng and Shan, Lianlei and Wu, Yong},
  journal={arXiv preprint arXiv:2505.17653},
  year={2025}
}

@article{wang2026codeglance,
  title={CodeGlance: Understanding Code Reasoning Challenges in LLMs through Multi-Dimensional Feature Analysis},
  author={Wang, Yunkun and Zhang, Xuanhe and Han, Junxiao and Zhi, Chen and Deng, Shuiguang},
  journal={arXiv preprint arXiv:2602.13962},
  year={2026}
}

@article{gandhi2026endless,
  title={Endless Terminals: Scaling RL Environments for Terminal Agents},
  author={Gandhi, Kanishk and Garg, Shivam and Goodman, Noah D and Papailiopoulos, Dimitris},
  journal={arXiv preprint arXiv:2601.16443},
  year={2026}
}

@misc{yao2023reactsynergizingreasoningacting,
      title={ReAct: Synergizing Reasoning and Acting in Language Models}, 
      author={Shunyu Yao and Jeffrey Zhao and Dian Yu and Nan Du and Izhak Shafran and Karthik Narasimhan and Yuan Cao},
      year={2023},
      eprint={2210.03629},
      archivePrefix={arXiv},
      primaryClass={cs.CL},
      url={https://arxiv.org/abs/2210.03629}, 
}

@misc{lu2024aiscientistfullyautomated,
      title={The AI Scientist: Towards Fully Automated Open-Ended Scientific Discovery}, 
      author={Chris Lu and Cong Lu and Robert Tjarko Lange and Jakob Foerster and Jeff Clune and David Ha},
      year={2024},
      eprint={2408.06292},
      archivePrefix={arXiv},
      primaryClass={cs.AI},
      url={https://arxiv.org/abs/2408.06292}, 
}

@misc{yamada2025aiscientistv2workshoplevelautomated,
      title={The AI Scientist-v2: Workshop-Level Automated Scientific Discovery via Agentic Tree Search}, 
      author={Yutaro Yamada and Robert Tjarko Lange and Cong Lu and Shengran Hu and Chris Lu and Jakob Foerster and Jeff Clune and David Ha},
      year={2025},
      eprint={2504.08066},
      archivePrefix={arXiv},
      primaryClass={cs.AI},
      url={https://arxiv.org/abs/2504.08066}, 
}

@misc{gottweis2025aicoscientist,
      title={Towards an AI co-scientist}, 
      author={Juraj Gottweis and Wei-Hung Weng and Alexander Daryin and Tao Tu and Anil Palepu and Petar Sirkovic and Artiom Myaskovsky and Felix Weissenberger and Keran Rong and Ryutaro Tanno and Khaled Saab and Dan Popovici and Jacob Blum and Fan Zhang and Katherine Chou and Avinatan Hassidim and Burak Gokturk and Amin Vahdat and Pushmeet Kohli and Yossi Matias and Andrew Carroll and Kavita Kulkarni and Nenad Tomasev and Yuan Guan and Vikram Dhillon and Eeshit Dhaval Vaishnav and Byron Lee and Tiago R D Costa and José R Penadés and Gary Peltz and Yunhan Xu and Annalisa Pawlosky and Alan Karthikesalingam and Vivek Natarajan},
      year={2025},
      eprint={2502.18864},
      archivePrefix={arXiv},
      primaryClass={cs.AI},
      url={https://arxiv.org/abs/2502.18864}, 
}

@article{swanson2025virtual,
  title={The Virtual Lab of AI agents designs new SARS-CoV-2 nanobodies},
  author={Swanson, Kyle and Wu, Wesley and Bulaong, Nash L and Pak, John E and Zou, James},
  journal={Nature},
  volume={646},
  number={8085},
  pages={716--723},
  year={2025},
  publisher={Nature Publishing Group UK London}
}

@article{huang2025biomni,
  title={Biomni: A general-purpose biomedical ai agent},
  author={Huang, Kexin and Zhang, Serena and Wang, Hanchen and Qu, Yuanhao and Lu, Yingzhou and Roohani, Yusuf and Li, Ryan and Qiu, Lin and Li, Gavin and Zhang, Junze and others},
  journal={biorxiv},
  year={2025}
}

@article{boiko2023autonomous,
  title={Autonomous chemical research with large language models},
  author={Boiko, Daniil A and MacKnight, Robert and Kline, Ben and Gomes, Gabe},
  journal={Nature},
  volume={624},
  number={7992},
  pages={570--578},
  year={2023},
  publisher={Nature Publishing Group UK London}
}

@misc{bran2023chemcrowaugmentinglargelanguagemodels,
      title={ChemCrow: Augmenting large-language models with chemistry tools}, 
      author={Andres M Bran and Sam Cox and Oliver Schilter and Carlo Baldassari and Andrew D White and Philippe Schwaller},
      year={2023},
      eprint={2304.05376},
      archivePrefix={arXiv},
      primaryClass={physics.chem-ph},
      url={https://arxiv.org/abs/2304.05376}, 
}

@article{Zou_2025,
   title={El Agente: An autonomous agent for quantum chemistry},
   volume={8},
   ISSN={2590-2385},
   url={http://dx.doi.org/10.1016/j.matt.2025.102263},
   DOI={10.1016/j.matt.2025.102263},
   number={7},
   journal={Matter},
   publisher={Elsevier BV},
   author={Zou, Yunheng and Cheng, Austin H. and Aldossary, Abdulrahman and Bai, Jiaru and Leong, Shi Xuan and Campos-Gonzalez-Angulo, Jorge Arturo and Choi, Changhyeok and Ser, Cher Tian and Tom, Gary and Wang, Andrew and Zhang, Zijian and Yakavets, Ilya and Hao, Han and Crebolder, Chris and Bernales, Varinia and Aspuru-Guzik, Alán},
   year={2025},
   pages={102263} }

@misc{schmidgall2025agentlaboratoryusingllm,
      title={Agent Laboratory: Using LLM Agents as Research Assistants}, 
      author={Samuel Schmidgall and Yusheng Su and Ze Wang and Ximeng Sun and Jialian Wu and Xiaodong Yu and Jiang Liu and Michael Moor and Zicheng Liu and Emad Barsoum},
      year={2025},
      eprint={2501.04227},
      archivePrefix={arXiv},
      primaryClass={cs.HC},
      url={https://arxiv.org/abs/2501.04227}, 
}

@misc{schmidgall2025agentrxivcollaborativeautonomousresearch,
      title={AgentRxiv: Towards Collaborative Autonomous Research}, 
      author={Samuel Schmidgall and Michael Moor},
      year={2025},
      eprint={2503.18102},
      archivePrefix={arXiv},
      primaryClass={cs.AI},
      url={https://arxiv.org/abs/2503.18102}, 
}

@misc{baek2025researchagentiterativeresearchidea,
      title={ResearchAgent: Iterative Research Idea Generation over Scientific Literature with Large Language Models}, 
      author={Jinheon Baek and Sujay Kumar Jauhar and Silviu Cucerzan and Sung Ju Hwang},
      year={2025},
      eprint={2404.07738},
      archivePrefix={arXiv},
      primaryClass={cs.CL},
      url={https://arxiv.org/abs/2404.07738}, 
}

@misc{huang2024mlagentbenchevaluatinglanguageagents,
      title={MLAgentBench: Evaluating Language Agents on Machine Learning Experimentation}, 
      author={Qian Huang and Jian Vora and Percy Liang and Jure Leskovec},
      year={2024},
      eprint={2310.03302},
      archivePrefix={arXiv},
      primaryClass={cs.LG},
      url={https://arxiv.org/abs/2310.03302}, 
}

@misc{chan2025mlebenchevaluatingmachinelearning,
      title={MLE-bench: Evaluating Machine Learning Agents on Machine Learning Engineering}, 
      author={Jun Shern Chan and Neil Chowdhury and Oliver Jaffe and James Aung and Dane Sherburn and Evan Mays and Giulio Starace and Kevin Liu and Leon Maksin and Tejal Patwardhan and Lilian Weng and Aleksander Mądry},
      year={2025},
      eprint={2410.07095},
      archivePrefix={arXiv},
      primaryClass={cs.CL},
      url={https://arxiv.org/abs/2410.07095}, 
}

@misc{chen2025scienceagentbenchrigorousassessmentlanguage,
      title={ScienceAgentBench: Toward Rigorous Assessment of Language Agents for Data-Driven Scientific Discovery}, 
      author={Ziru Chen and Shijie Chen and Yuting Ning and Qianheng Zhang and Boshi Wang and Botao Yu and Yifei Li and Zeyi Liao and Chen Wei and Zitong Lu and Vishal Dey and Mingyi Xue and Frazier N. Baker and Benjamin Burns and Daniel Adu-Ampratwum and Xuhui Huang and Xia Ning and Song Gao and Yu Su and Huan Sun},
      year={2025},
      eprint={2410.05080},
      archivePrefix={arXiv},
      primaryClass={cs.CL},
      url={https://arxiv.org/abs/2410.05080}, 
}

@misc{majumder2024discoverybenchdatadrivendiscoverylarge,
      title={DiscoveryBench: Towards Data-Driven Discovery with Large Language Models}, 
      author={Bodhisattwa Prasad Majumder and Harshit Surana and Dhruv Agarwal and Bhavana Dalvi Mishra and Abhijeetsingh Meena and Aryan Prakhar and Tirth Vora and Tushar Khot and Ashish Sabharwal and Peter Clark},
      year={2024},
      eprint={2407.01725},
      archivePrefix={arXiv},
      primaryClass={cs.CL},
      url={https://arxiv.org/abs/2407.01725}, 
}

@misc{ni2024matpilotllmenabledaimaterials,
      title={MatPilot: an LLM-enabled AI Materials Scientist under the Framework of Human-Machine Collaboration}, 
      author={Ziqi Ni and Yahao Li and Kaijia Hu and Kunyuan Han and Ming Xu and Xingyu Chen and Fengqi Liu and Yicong Ye and Shuxin Bai},
      year={2024},
      eprint={2411.08063},
      archivePrefix={arXiv},
      primaryClass={physics.soc-ph},
      url={https://arxiv.org/abs/2411.08063}, 
}

@article{szymanski2023autonomous,
  title={An autonomous laboratory for the accelerated synthesis of inorganic materials},
  author={Szymanski, Nathan J and Rendy, Bernardus and Fei, Yuxing and Kumar, Rishi E and He, Tanjin and Milsted, David and McDermott, Matthew J and Gallant, Max and Cubuk, Ekin Dogus and Merchant, Amil and others},
  journal={Nature},
  volume={624},
  number={7990},
  pages={86},
  year={2023}
}

@misc{macleod2020selfdrivinglaboratoryaccelerateddiscovery,
      title={Self-driving laboratory for accelerated discovery of thin-film materials}, 
      author={Benjamin P. MacLeod and Fraser G. L. Parlane and Thomas D. Morrissey and Florian Häse and Loïc M. Roch and Kevan E. Dettelbach and Raphaell Moreira and Lars P. E. Yunker and Michael B. Rooney and Joseph R. Deeth and Veronica Lai and Gordon J. Ng and Henry Situ and Ray H. Zhang and Michael S. Elliott and Ted H. Haley and David J. Dvorak and Alán Aspuru-Guzik and Jason E. Hein and Curtis P. Berlinguette},
      year={2020},
      eprint={1906.05398},
      archivePrefix={arXiv},
      primaryClass={physics.app-ph},
      url={https://arxiv.org/abs/1906.05398}, 
}

@article{mehr2020universal,
  title={A universal system for digitization and automatic execution of the chemical synthesis literature},
  author={Mehr, S Hessam M and Craven, Matthew and Leonov, Artem I and Keenan, Graham and Cronin, Leroy},
  journal={Science},
  volume={370},
  number={6512},
  pages={101--108},
  year={2020},
  publisher={American Association for the Advancement of Science}
}

@article{hubert2025olympiad,
  title={Olympiad-level formal mathematical reasoning with reinforcement learning},
  author={Hubert, Thomas and Mehta, Rishi and Sartran, Laurent and Horv{\'a}th, Mikl{\'o}s Z and {\v{Z}}u{\v{z}}i{\'c}, Goran and Wieser, Eric and Huang, Aja and Schrittwieser, Julian and Schroecker, Yannick and Masoom, Hussain and others},
  journal={Nature},
  pages={1--3},
  year={2025},
  publisher={Nature Publishing Group UK London}
}

@misc{novikov2025alphaevolvecodingagentscientific,
      title={AlphaEvolve: A coding agent for scientific and algorithmic discovery}, 
      author={Alexander Novikov and Ngân Vũ and Marvin Eisenberger and Emilien Dupont and Po-Sen Huang and Adam Zsolt Wagner and Sergey Shirobokov and Borislav Kozlovskii and Francisco J. R. Ruiz and Abbas Mehrabian and M. Pawan Kumar and Abigail See and Swarat Chaudhuri and George Holland and Alex Davies and Sebastian Nowozin and Pushmeet Kohli and Matej Balog},
      year={2025},
      eprint={2506.13131},
      archivePrefix={arXiv},
      primaryClass={cs.AI},
      url={https://arxiv.org/abs/2506.13131}, 
}

@inproceedings{li2025metal,
  title={Metal: A multi-agent framework for chart generation with test-time scaling},
  author={Li, Bingxuan and Wang, Yiwei and Gu, Jiuxiang and Chang, Kai-Wei and Peng, Nanyun},
  booktitle={Proceedings of the 63rd Annual Meeting of the Association for Computational Linguistics (Volume 1: Long Papers)},
  pages={30054--30069},
  year={2025}
}

@misc{odonoghue2023bioplannerautomaticevaluationllms,
      title={BioPlanner: Automatic Evaluation of LLMs on Protocol Planning in Biology}, 
      author={Odhran O'Donoghue and Aleksandar Shtedritski and John Ginger and Ralph Abboud and Ali Essa Ghareeb and Justin Booth and Samuel G Rodriques},
      year={2023},
      eprint={2310.10632},
      archivePrefix={arXiv},
      primaryClass={cs.CL},
      url={https://arxiv.org/abs/2310.10632}, 
}

@misc{lala2023paperqaretrievalaugmentedgenerativeagent,
      title={PaperQA: Retrieval-Augmented Generative Agent for Scientific Research}, 
      author={Jakub Lála and Odhran O'Donoghue and Aleksandar Shtedritski and Sam Cox and Samuel G. Rodriques and Andrew D. White},
      year={2023},
      eprint={2312.07559},
      archivePrefix={arXiv},
      primaryClass={cs.CL},
      url={https://arxiv.org/abs/2312.07559}, 
}

@misc{zheng2024gpt4visiongeneralistwebagent,
      title={GPT-4V(ision) is a Generalist Web Agent, if Grounded}, 
      author={Boyuan Zheng and Boyu Gou and Jihyung Kil and Huan Sun and Yu Su},
      year={2024},
      eprint={2401.01614},
      archivePrefix={arXiv},
      primaryClass={cs.IR},
      url={https://arxiv.org/abs/2401.01614}, 
}

@misc{qin2025uitarspioneeringautomatedgui,
      title={UI-TARS: Pioneering Automated GUI Interaction with Native Agents}, 
      author={Yujia Qin and Yining Ye and Junjie Fang and Haoming Wang and Shihao Liang and Shizuo Tian and Junda Zhang and Jiahao Li and Yunxin Li and Shijue Huang and Wanjun Zhong and Kuanye Li and Jiale Yang and Yu Miao and Woyu Lin and Longxiang Liu and Xu Jiang and Qianli Ma and Jingyu Li and Xiaojun Xiao and Kai Cai and Chuang Li and Yaowei Zheng and Chaolin Jin and Chen Li and Xiao Zhou and Minchao Wang and Haoli Chen and Zhaojian Li and Haihua Yang and Haifeng Liu and Feng Lin and Tao Peng and Xin Liu and Guang Shi},
      year={2025},
      eprint={2501.12326},
      archivePrefix={arXiv},
      primaryClass={cs.AI},
      url={https://arxiv.org/abs/2501.12326}, 
}

@misc{zhang2023appagentmultimodalagentssmartphone,
      title={AppAgent: Multimodal Agents as Smartphone Users}, 
      author={Chi Zhang and Zhao Yang and Jiaxuan Liu and Yucheng Han and Xin Chen and Zebiao Huang and Bin Fu and Gang Yu},
      year={2023},
      eprint={2312.13771},
      archivePrefix={arXiv},
      primaryClass={cs.CV},
      url={https://arxiv.org/abs/2312.13771}, 
}

@misc{wang2024mobileagentv2mobiledeviceoperation,
      title={Mobile-Agent-v2: Mobile Device Operation Assistant with Effective Navigation via Multi-Agent Collaboration}, 
      author={Junyang Wang and Haiyang Xu and Haitao Jia and Xi Zhang and Ming Yan and Weizhou Shen and Ji Zhang and Fei Huang and Jitao Sang},
      year={2024},
      eprint={2406.01014},
      archivePrefix={arXiv},
      primaryClass={cs.CL},
      url={https://arxiv.org/abs/2406.01014}, 
}

@misc{liu2024autoglmautonomousfoundationagents,
      title={AutoGLM: Autonomous Foundation Agents for GUIs}, 
      author={Xiao Liu and Bo Qin and Dongzhu Liang and Guang Dong and Hanyu Lai and Hanchen Zhang and Hanlin Zhao and Iat Long Iong and Jiadai Sun and Jiaqi Wang and Junjie Gao and Junjun Shan and Kangning Liu and Shudan Zhang and Shuntian Yao and Siyi Cheng and Wentao Yao and Wenyi Zhao and Xinghan Liu and Xinyi Liu and Xinying Chen and Xinyue Yang and Yang Yang and Yifan Xu and Yu Yang and Yujia Wang and Yulin Xu and Zehan Qi and Yuxiao Dong and Jie Tang},
      year={2024},
      eprint={2411.00820},
      archivePrefix={arXiv},
      primaryClass={cs.HC},
      url={https://arxiv.org/abs/2411.00820}, 
}

@misc{lin2024showuivisionlanguageactionmodelgui,
      title={ShowUI: One Vision-Language-Action Model for GUI Visual Agent}, 
      author={Kevin Qinghong Lin and Linjie Li and Difei Gao and Zhengyuan Yang and Shiwei Wu and Zechen Bai and Weixian Lei and Lijuan Wang and Mike Zheng Shou},
      year={2024},
      eprint={2411.17465},
      archivePrefix={arXiv},
      primaryClass={cs.CV},
      url={https://arxiv.org/abs/2411.17465}, 
}

@misc{hong2024cogagentvisuallanguagemodel,
      title={CogAgent: A Visual Language Model for GUI Agents}, 
      author={Wenyi Hong and Weihan Wang and Qingsong Lv and Jiazheng Xu and Wenmeng Yu and Junhui Ji and Yan Wang and Zihan Wang and Yuxuan Zhang and Juanzi Li and Bin Xu and Yuxiao Dong and Ming Ding and Jie Tang},
      year={2024},
      eprint={2312.08914},
      archivePrefix={arXiv},
      primaryClass={cs.CV},
      url={https://arxiv.org/abs/2312.08914}, 
}

@misc{cheng2024seeclickharnessingguigrounding,
      title={SeeClick: Harnessing GUI Grounding for Advanced Visual GUI Agents}, 
      author={Kanzhi Cheng and Qiushi Sun and Yougang Chu and Fangzhi Xu and Yantao Li and Jianbing Zhang and Zhiyong Wu},
      year={2024},
      eprint={2401.10935},
      archivePrefix={arXiv},
      primaryClass={cs.HC},
      url={https://arxiv.org/abs/2401.10935}, 
}

@misc{you2024ferretuigroundedmobileui,
      title={Ferret-UI: Grounded Mobile UI Understanding with Multimodal LLMs}, 
      author={Keen You and Haotian Zhang and Eldon Schoop and Floris Weers and Amanda Swearngin and Jeffrey Nichols and Yinfei Yang and Zhe Gan},
      year={2024},
      eprint={2404.05719},
      archivePrefix={arXiv},
      primaryClass={cs.CV},
      url={https://arxiv.org/abs/2404.05719}, 
}

@misc{wu2024osatlasfoundationactionmodel,
      title={OS-ATLAS: A Foundation Action Model for Generalist GUI Agents}, 
      author={Zhiyong Wu and Zhenyu Wu and Fangzhi Xu and Yian Wang and Qiushi Sun and Chengyou Jia and Kanzhi Cheng and Zichen Ding and Liheng Chen and Paul Pu Liang and Yu Qiao},
      year={2024},
      eprint={2410.23218},
      archivePrefix={arXiv},
      primaryClass={cs.CL},
      url={https://arxiv.org/abs/2410.23218}, 
}

@misc{yang2025ariauivisualgroundinggui,
      title={Aria-UI: Visual Grounding for GUI Instructions}, 
      author={Yuhao Yang and Yue Wang and Dongxu Li and Ziyang Luo and Bei Chen and Chao Huang and Junnan Li},
      year={2025},
      eprint={2412.16256},
      archivePrefix={arXiv},
      primaryClass={cs.HC},
      url={https://arxiv.org/abs/2412.16256}, 
}

@misc{sun2025osgenesisautomatingguiagent,
      title={OS-Genesis: Automating GUI Agent Trajectory Construction via Reverse Task Synthesis}, 
      author={Qiushi Sun and Kanzhi Cheng and Zichen Ding and Chuanyang Jin and Yian Wang and Fangzhi Xu and Zhenyu Wu and Chengyou Jia and Liheng Chen and Zhoumianze Liu and Ben Kao and Guohao Li and Junxian He and Yu Qiao and Zhiyong Wu},
      year={2025},
      eprint={2412.19723},
      archivePrefix={arXiv},
      primaryClass={cs.AI},
      url={https://arxiv.org/abs/2412.19723}, 
}

@misc{gou2025navigatingdigitalworldhumans,
      title={Navigating the Digital World as Humans Do: Universal Visual Grounding for GUI Agents}, 
      author={Boyu Gou and Ruohan Wang and Boyuan Zheng and Yanan Xie and Cheng Chang and Yiheng Shu and Huan Sun and Yu Su},
      year={2025},
      eprint={2410.05243},
      archivePrefix={arXiv},
      primaryClass={cs.AI},
      url={https://arxiv.org/abs/2410.05243}, 
}

@misc{zheng2026code2worldguiworldmodel,
      title={Code2World: A GUI World Model via Renderable Code Generation}, 
      author={Yuhao Zheng and Li'an Zhong and Yi Wang and Rui Dai and Kaikui Liu and Xiangxiang Chu and Linyuan Lv and Philip Torr and Kevin Qinghong Lin},
      year={2026},
      eprint={2602.09856},
      archivePrefix={arXiv},
      primaryClass={cs.CV},
      url={https://arxiv.org/abs/2602.09856}, 
}

\clearpage

\vfill

\end{document}